\documentclass[11pt]{article}

\usepackage{lineno,hyperref}
\usepackage{amsmath}
\usepackage{amssymb}
\usepackage{epstopdf}
\usepackage{adjustbox}
\usepackage{booktabs}
\usepackage{cancel}
\usepackage{xcolor}
\modulolinenumbers[5]
\usepackage{color}
\usepackage{dsfont}

\newcommand\MB{\mathrm{MB}}
\newcommand\layer{\ell}

\begin{document}


\title{Physics-aware deep neural networks for surrogate modeling of turbulent natural convection}

\author{Didier Lucor$^{(a,\star)}$, Atul Agrawal$^{(b,a)}$ and Anne Sergent$^{(a,c)}$ \medskip \\
$^{(a)}$\small{ Universit\'e Paris-Saclay, CNRS,} \\ \small{Laboratoire Interdisciplinaire des Sciences du Num\'erique (LISN),} \\ \small{Orsay, France} \smallskip\\
$^{(b)}$\small{Department of Mechanical Engineering, Technical University of Munich,}\\ \small{Garching b. M\"unchen, Germany} \smallskip\\
$^{(c)}$\small{Sorbonne Universit\'e, Facult\'e des Sciences et Ing\'enierie,}\\ \small{UFR Ing\'enierie, Paris, France}
}

%
%
%
%

\maketitle

\begin{abstract}
Recent works have explored the potential of machine learning as data-driven turbulence closures for RANS and LES  techniques. Beyond these advances, the high expressivity and agility of physics-informed neural networks (PINNs) make them promising candidates for full fluid flow PDE modeling. An important question is whether this new paradigm, exempt from the traditional notion of discretization of the underlying operators very much connected to the flow scales resolution, is capable of sustaining high levels of turbulence characterized by multi-scale features? 
We investigate the use of PINNs surrogate modeling for turbulent Rayleigh-B\'enard (RB) convection flows in rough and smooth rectangular cavities, mainly relying on DNS temperature data from the fluid bulk. We carefully quantify the computational requirements under which the formulation is capable of accurately recovering the flow hidden quantities.
We then propose a new padding technique to distribute some of the scattered coordinates - at which PDE residuals are minimized - around the region of labeled data acquisition. 
We show how it comes to play as a regularization close to the training boundaries which are zones of poor accuracy for standard PINNs and results in a noticeable global accuracy improvement at iso-budget.
Finally, we propose for the first time to relax the incompressibility condition in such a way that it drastically benefits the optimization search and results in a much improved convergence of the composite loss function. 
The RB results obtained at high Rayleigh number $Ra=2\cdot 10^9$ are particularly impressive: the predictive accuracy of the surrogate over the entire half a billion DNS coordinates yields errors for all flow variables ranging between $[0.3\%-4\%]$  in the relative $L_2$ norm, with a training relying  only on $1.6\%$ of the DNS data points.    
\end{abstract}


\section{Introduction}

Deep learning (DL) is investigated among data-driven methods as a surrogate for physics-driven computational fluid dynamics (CFD) methods solving expensive nonlinear coupled PDEs, such as the ones describing turbulent numerical simulations or experiments. It seems to be somehow capable of producing 
realistic instantaneous flow fields with reasonable physically accurate spatio-temporal coherence, without solving the actual partial differential equations (PDEs) governing the system \cite{kutz2017deep,brunton2020machine}. 
DL is also promising because of its proficiency in extracting low-dimensional information from large amount of high-dimensional turbulent data.
This new paradigm is interesting for applications involving flow optimization and control, uncertainty quantification, gappy data reconstruction or multi-scale flow analysis, for which the turbulent prediction may be queried in real-time or many times. 
In practice, DL models are solved with streams of data as ``black boxes" \cite{goodfellow2016deep}. 
In their standard form, they lack knowledge of the underlying physics and when they achieve low prediction errors their efficiency  remains hard to interpret. In fact, they do not necessarily satisfy the physics of the systems they model. It is therefore crucial to inject some known physics and principles into their framework, not only to get more physically meaningful results but also in order to better guide the learning process \cite{ling2016machine}. Besides, the physical invariants may help in recovering hidden system quantities for which no data were available, which is very common in experiments. While machine learning methods make sense as data-driven closure models for Reynolds-Averaged Navier Stokes (RANS)  \cite{xiao2019quantification} and Large Eddy Simulations (LES) techniques \cite{duraisamy2019turbulence}, they may also be used for full PDE modeling \cite{RAISSI2019686}. An interrogation remains in terms of their potential as actual replacement of {\em costly} traditional PDE solution methods {\em at all scales}, such as direct numerical simulations (DNS). In this paper, we will test the efficiency of physics-aware DL for metamodeling turbulent natural convection.\\
Turbulent natural convection is a spontaneous physical process present in many natural systems (oceans, atmospheres or mantles) as well as in engineering applications, such as passive cooling of braking systems, nuclear power plants or electronic devices or natural ventilation of buildings. 
A canonical system of such turbulent heat transport mechanisms is the 
Rayleigh-B\'enard (RB) cell, where the temperature and velocity fields interact through the buoyancy force \cite{ChillaNewperspectives}.
Efficiently modelling of this phenomenon is the first step towards heat transfer control for more sustainable energy systems, but it requires to track the heat carriers, namely the small-scale plumes. This remains a challenge due to the double kinetic and thermal nature of plumes, and the nonlinear interactions between various spatial and time scales from the large scale circulation to the small vortices and plumes \cite{Castillo_JFM_2019}. The continued increase of supercomputing
power in recent years has enabled the DNS of highly turbulent flows, resolving the entire array of scales at very high Rayleigh number. But it involves such a computational effort in terms of degree of parallelism, CPU resources and storage capacity that it will eventually entail a restriction on the spatial DNS resolution/storage and will therefore hamper its analysis. 
Few recent studies have been pioneering the use of deep learning in the framework of turbulent heat transfer with various aims. For instance, Kim et al. \cite{kim2020prediction} used DL, in the form of a convolutional neural network (CNN), to predict the turbulent heat transfer -- reconstructing the wall-normal heat flux at the wall -- on the basis of other wall information (such as wall shear stress) obtained by DNS of channel flow with a passive temperature field.
Fonda et al. \cite{fonda2019deep} have tracked turbulent superstructures in RB convection in horizontally extended systems at $Ra =10^{5,6,7}$ with U-shaped network (encoder-decoder CNN) allowing to reduce the dimensionality of the structures to slowly-evolving temporal 2D planar network of ridges. The idea here is to propose an automated tool exploiting a large DNS simulation in order to explore heat transfer properties more easily. Pandey et al. were more interested by turbulent statistical prediction of 2D large scale structures. They relied on reservoir computing modeling, which may be seen as an hybridization between a proper orthogonal decomposition (POD) of DNS data and a recurrent neural network (RNN), to tackle RB cavity flow at $Ra =10^{7}$ \cite{pandey2020reservoir}.
At the foundation of all of these works is the use of large DNS database from which partial information, in the form of wall data or time-windowed averaging, or more global information, in the form of POD, are extracted. We propose to leverage deep neural networks to {\em replace} the DNS {\em three-dimensional} solver by a much more agile data-driven and physics-aware surrogate. Most importantly, this surrogate will be trained with {\em partial} DNS data, but will incorporate known physics (such as symmetries, constraints and conservation laws in the training process), allowing the inference of hidden fluid quantities of interest \cite{raissi2020hidden,jin2020nsfnets}.\\
Raissi et al. \cite{RAISSI2019686} first introduced the concept of physics-informed neural networks (PINNs) to solve forward and inverse problems involving several different types of PDEs. This approach may be apprehended as a combination of data-driven supervised and physically-constrained unsupervised learning. For instance,  they propose a way of approximating Navier-Stokes (NS) solutions that do not require mesh generation. Most of the flows considered in the aforementioned works are laminar at relatively low Reynolds numbers. A fundamental question related to whether PINNs could simulate three-dimensional turbulence directly, similarly to high-order DNS, was answered in \cite{jin2020nsfnets}. In this study, the authors test different NS formulations for simulating turbulence. For their velocity-pressure formulation, that they found most effective, they provide their PINNs with velocity DNS data collected from the initial condition, boundaries and inside of a subdomain of their turbulent channel flow system. Closer to our application, they propose in \cite{raissi2020hidden} a similar approach for inferring pressure and fluid velocity from monitoring of {\em passively} advected scalar data with applications to laminar flows. In this paper, we wish to investigate the relevance of the approach to three-dimensional turbulent heat transfer in the form of natural convection and modeled as Navier-Stokes equations under Boussinesq approximation, for which the temperature acts as an active quantity with a feedback on the momentum equations. We will put  particular emphasis on the efficiency of the surrogate modeling with respect to data and residual sampling strategy.

The paper is organized as follows. We first recall the basics of the standard PINNs formulation in section \ref{sec:PINNs}. We then introduce the turbulent  RB cavity class of problems that we consider. We present various PINNs results and propose some methodological improvements in section \ref{sec:results}. Finally, we discuss our findings, introduce and test a new idea on a more turbulent setup and end up with some perspectives in section \ref{sec:discussion}.

\section{Physics-informed DNN as a viable data-driven method to solve nonlinear PDEs}
\label{sec:PINNs}
Deep learning tools have recently seemed to start providing a different approach to computational mechanics.
In particular, deep neural networks (DNN) are now considered as an alternative way of approximating the solution of various {\em deterministic} PDEs types. Since some earlier studies 
\cite{MEADE19941,lagaris2000neural,lagaris1998artificial,McFall2009}, and thanks to significant computational advances in automatic differentiation, DNN used as surrogates of PDEs solutions have generated a broad interest from the community \cite{KUMAR20113796,MALL2016347,berg2018unified,Yang2016}. Nevertheless, in the small data regime, their efficiency remains often limited and their prediction lacks robustness and interpretability, motivating the idea of ``adding" any form of prior knowledge to the numerical surrogate, in order to provide some kind of ``training guidance". \\
One approach is to design a specialized network architecture embedding the prior knowledge relevant to the task at hand. This is for instance the case of the convolutional neural networks (CNN) which have revolutionized the field of computer vision thanks to their translation invariant characteristics with impressive applications in image classification, medical image analysis, natural language processing, etc. Another approach relies on a softer enforcing of this knowledge.
Raissi et al. \cite{RAISSI2019686} have proposed to rely on physics-inspired neural networks (PINNs) for approximating solutions to general nonlinear PDEs and validated it with a series of benchmark test cases. The main feature is the inclusion of some form of prior knowledge about the physics of the problem in the learning algorithm, in addition to the data used to train the network. This is done through an enlarged/enhanced loss/cost function. This way, the outputs of the neural network are constrained to approximately satisfy a system of PDEs by using a regularization functional $\mathcal{L}_{\text{PDE}}$ that typically corresponds to the residual (or the variational energy) of the set of PDEs under the neural network representation.
The algorithm imposes a penalty for non-physical solutions and hopefully redirects it quicker towards the correct solution. As a result, the algorithm has good generalization property even in the small data set regime. This approach recently drew a lot of attention and is the subject of several numerical investigations including recent development of dedicated computational packages \cite{HAGHIGHAT2021113552}.\\
Nevertheless, the plain version of PINNs numerically suffers from several drawbacks. Indeed they are for instance notoriously hard to train for multi-scale and/or high-frequency problems. In fact, a first difficulty resides in the discrepancy of convergence rate between the different terms of the loss function depending on the change in the learning rate. This comes from an imbalanced magnitude of the back-propagated gradients during model training. It is therefore possible in practice to assign some weight coefficients within the loss function than can effectively assign a different learning rate to each individual loss term. These weights may be user-specified or tuned automatically during network training \cite{wang2020understanding}. Moreover, the required depth of the network increases with increasing order of the set of PDEs, leading to slow learning-rate due to the issue of vanishing gradients. It was also noticed that PINNs are not always robust in representing sharp local gradients \cite{dwivedi2019distributed}.\\
Another source of discredit of PINNs as described is its dependence to the data. 
Other teams have developed physics-constrained, {\em data-free} DNN for surrogate modeling of
incompressible flows. The idea is to {\em enforce} the initial/boundary conditions instead of being penalized together during training, which is solely driven by minimizing the residuals of the governing PDEs. Some two-dimensional vascular laminar flows with idealized geometries are tested with this approach in \cite{SUN2020112732}.

\subsection{Notations and formulation}

\begin{figure}[!h]
\centering
\includegraphics[trim={7.5cm 10cm 10cm 6cm},clip,width=11cm]{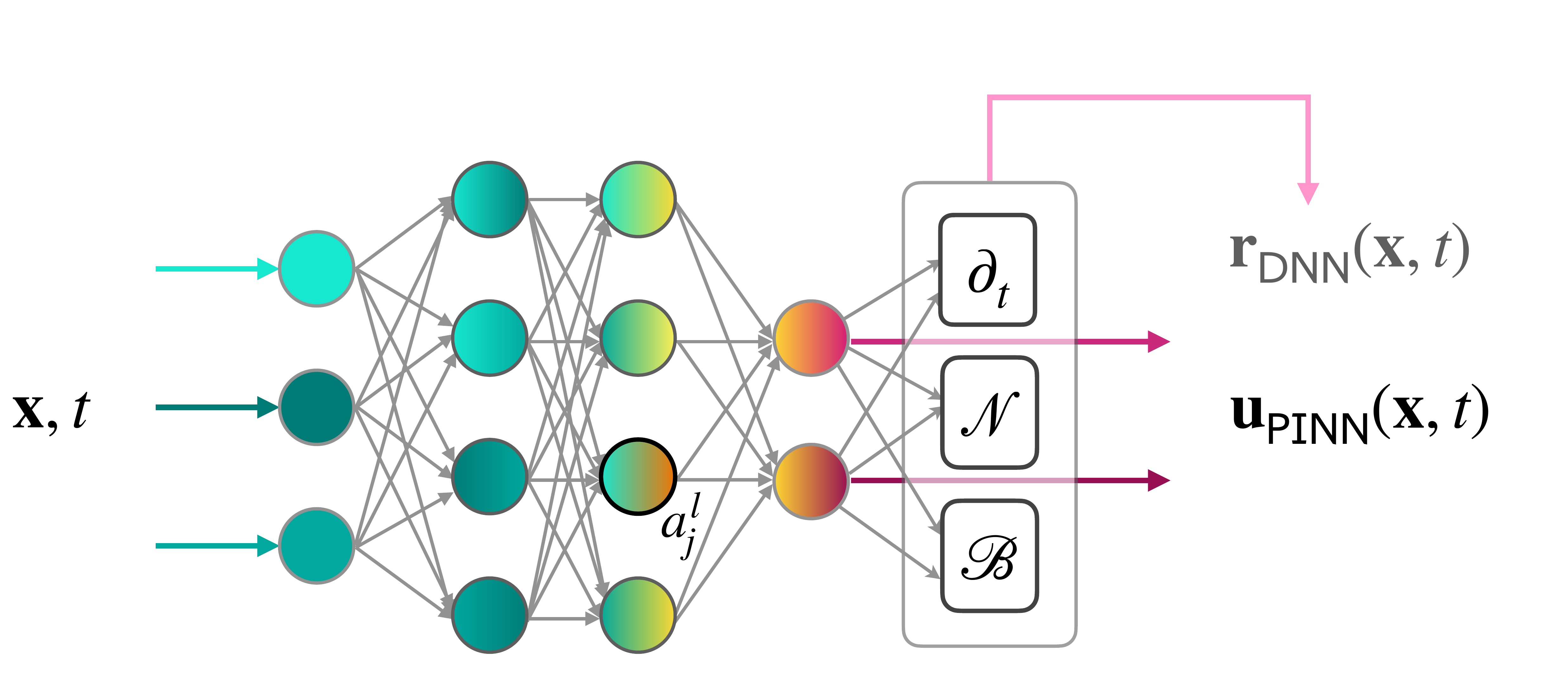}
\caption{Schematic of a physics-informed neural network (PINN). 
Thanks to automatic differentiation, differential operators applied to the $\mathbf{u}_{\text{DNN}}$ outputs are available to evaluate the residuals $\mathbf{r}_{\text{DNN}}$ (that are wished small) of the set of PDEs . These residuals are an additional byproduct information which is used to regularize the loss function and help in the overall training process.  }
\label{fig:PINN}
\end{figure}

Here we introduce our notations and briefly describe the general DNN framework and how it is coupled with a second network to form the PINN approach.\\
The goal is to {\em approximate} the exact solution of a model noted $\mathcal{M}$, i.e. a set of unsteady PDEs, written in a generic form inside and at the boundary of a physical domain $\Omega$ evolving over a time interval $\mathcal{D}=[0,T_f]$ as:
\begin{eqnarray}
    \mathbf{u}(\mathbf{x},t)_t + \mathcal{N}\left (\mathbf{u}(\mathbf{x},t);\lambda \right ) & = & R(\mathbf{x},t), \quad (\mathbf{x},t)\in \Omega\in \mathbb{R}^d \times \mathcal{D}, \nonumber \\
    \mathcal{B}\left (\mathbf{u}(\mathbf{x},t) \right ) & = & B(\mathbf{x},t), \quad (\mathbf{x},t)\in \partial\Omega \in \mathbb{R}^{d-1} \times \mathcal{D}.
    \label{eq:PDEs_generic}
\end{eqnarray}
whose exact solution representing the system unknown variables is defined as: 
\begin{equation}
    \mathbf{u}=\hat{f}(\mathbf{x},t), \quad \text{and satisfying} \quad \mathcal{M}(\mathbf{u})=0,
\end{equation}
through the response of a neural network:
\begin{equation}
\mathbf{u} \approx \mathbf{u}_{\text{DNN}} = f_{\boldsymbol{\theta}}(\mathbf{x},t) , \quad \text{with} \quad \mathcal{M}\big ( f_{\boldsymbol{\theta}}(\mathbf{x},t)\big )=\mathbf{r}(\mathbf{x},t),
\end{equation}
with $\mathbf{r}$ representing the residual fields of the set of equations.
In Eq. (\ref{eq:PDEs_generic}), $\mathcal{N}$ is a general spatial differential operator (which may include a parameter $\lambda$) in the domain $\Omega$, while $\mathcal{B}$ is the boundary operator on $\partial\Omega$; $R$ and $B$ are potential source fields. \\
More specifically, in our case, the model describing our physical system of interest encompasses the three-dimensional incompressible Navier-Stokes equations under the Boussinesq approximation, which may be written in non-dimensional form as:
\begin{eqnarray}
    \mathbf{v}_t + \left (\mathbf{v} \cdot \nabla \right ) \mathbf{v} & = & - \nabla p + \frac{\text{Pr}}{\text{Ra}^{1/2}} \Delta \mathbf{v} + \text{Pr}\, T\, \mathbf{e}_z,  \nonumber \\
         T_t + \mathbf{v} \cdot \nabla T & = &  \frac{1}{\text{Ra}^{1/2}} \Delta T, \nonumber \\
         \nabla \cdot \mathbf{v} & = & 0,
     \label{eq:PDEs_NS}
\end{eqnarray}
with the Rayleigh number $\text{Ra}=g\beta\, \Delta T\, H^3/\nu\kappa$, $\nu$ the kinematic viscosity,
$\kappa$ the thermal diffusivity, $\beta$ the thermal expansion coefficient, $\Delta T$ the reference temperature difference and $H$ the domain reference length, and the Prandtl number $\text{Pr}=\nu/\kappa$.
The reference velocity is taken equal to the convective velocity $V_{ref}=\dfrac{\kappa}{H}\sqrt{Ra}$. The time $t$ is therefore expressed in convective time units.
So, in our case, we denote the solution as: $\mathbf{u}\equiv \left (\mathbf{v},p,T \right)$, where $\mathbf{v}\equiv(\mathrm{v}_x,\mathrm{v}_y,\mathrm{v}_z)^{\mathrm{t}}$, $p$ and $T$ are the {dimensionless} fluid velocity, pressure and temperature, respectively.  
Following the idea proposed in \cite{raissi2020hidden}, an auxiliary variable $\overline{T}=1-T$ is added, that  satisfies a similar transport equation as the temperature field. Later on, this complementary equation will be useful to the training of the networks. It will act as an additional  constraint helping the algorithm to better converge as shown in \cite{raissi2020hidden}. The DNN will therefore have to learn the nonlinear continuous mapping relating inputs and outputs of the system.\\
The main portion of this network is a multi-layer perceptron (MLP) made of interconnected neurons assembled in $\layer \in \mathbb{N}$ hidden layers. The network dimensions are as follows: 
$n^{0}=n_{\mathbf{x}}+1 \in \mathbb{N}$ the input dimension, $n^{\layer+1}=n_{\mathbf{u}} \in \mathbb{N}$ the output dimension and $n^l$ the dimension of each hidden layer. The network architecture sequence $\mathcal{A}$ may be summarized as $\mathcal{A}=\big ( n_{\mathbf{x}}+1,n^{1},\ldots,n^{l},\ldots,n^{\layer},n_{\mathbf{u}} \big )$. 
Looking at computational mechanisms of the DNN in more details, we define the following affine linear maps between adjacent layers:
\begin{eqnarray}
    g_{\boldsymbol{\theta}^{l}}^{l} & : & \mathbb{R}^{n^{l-1}} \rightarrow \mathbb{R}^{n^{l}} \nonumber \\
    & : & \boldsymbol{a}^{l-1} \mapsto \boldsymbol{W}^{l}\boldsymbol{a}^{l-1} + \boldsymbol{b}^{l} , \quad \text{for} \quad l =1,\ldots,\layer,
\end{eqnarray}
where $\boldsymbol{a}^{l-1} \in \mathbb{R}^{n^{l-1}}$ is an array containing all the values taken by the neurons belonging to the $(l-1)-$layer. 
The quantities $\boldsymbol{\theta}^{(\cdot)} \equiv \left (\boldsymbol{W}^{(\cdot)}\in \mathbb{R}^{n_{l-1}\times n_l},\boldsymbol{b}^{(\cdot)} \in \mathbb{R}^{ n_l} \right )$ represent the parameters containing the weights and biases to be calibrated. \\
Summarizing the telescoping approximation form of the DNN output, we may write it as follows:
\begin{equation}
    \mathbf{u} \approx \mathbf{u}_{\text{DNN}}=f_{\boldsymbol{\theta}}(\mathbf{x,t})=g^{\layer}_{\boldsymbol{\theta}^{\layer}} \Bigg ( \rho \bigg ( g^{\layer-1}_{\boldsymbol{\theta}^{\layer-1}} \Big (\rho \big (\ldots \rho  (g^{1}_{\boldsymbol{\theta}^{1}}(\mathbf{x},t) ) \big ) \Big ) \bigg ) \Bigg ),
\end{equation}
where $\rho: \mathbb{R} \rightarrow \mathbb{R}$ is a nonlinear activation function, kept the same in the entire network in this study.\\
Classically, for a chosen architecture, the neural network may be trained with a large, but potentially noisy and scattered, training set of data  $ \{ ((\mathbf{x},t)^{(\text{train})},\mathbf{u}_{\star}^{(\text{train})}) \}$  by optimizing its parameters  $\boldsymbol{\theta}$.
In order to minimize the error associated with the prediction of the DNN, an objective function is required by the optimization. It is referred as the loss (or cost) function and maps the set of parameter values for the network onto a scalar value. For regression problems, mean-squared error (MSE) loss functions also named $L_2$-based loss functions are usually preferred:
\begin{equation}
    \mathcal{L}_{\text{Label}}\left ( \boldsymbol{\theta},\{ (\mathbf{x},t)^{(i))} \}_{i \in\mathcal{I}}\right )=\frac{1}{n_L}\sum_{i=1}^{n_L} \| \mathbf{u}_{\text{DNN}}^{(i)}-\mathbf{u}_{\star}^{(i)} \|,
    \label{eq:standard_loss}
\end{equation}
where $\mathcal{I}=\{1,\ldots,{n_L}\}$ is a defined set with $n_L$ the size of the data sample.
Finding the optimal value of $\boldsymbol{\theta}$ under this norm is equivalent to maximizing the conditional log-likelihood distribution $\sum_{i=1}^{N_L}\log \pi \left ( \mathbf{u}^{(i)} | (\mathbf{x},t)^{(i)},\boldsymbol{\theta}\right )$ \cite{goodfellow2016deep}.\\
Once the parameters have been tuned, thanks to the graph-based implementation of DNNs, it is in fact straightforward to compute exactly derivatives of the surrogate network of output $\mathbf{u}$ with respect to its inputs, i.e. spatial/temporal derivatives,  by applying the chain rule for differentiating compositions of functions using the automatic differentiation, which is conveniently integrated in many machine learning packages such as Tensorflow \cite{tensorflow2015-whitepaper}.\\
The PINN approach takes advantage of this functionality. 
Figure (\ref{fig:PINN}) graphically describes the structure of the PINN approach, for which the loss
function contains a mismatch in the given {\em partial} data on {\em some} state variables combined with the residual of the PDEs computed on a set of random points in the time-space domain. 
This time the loss function may be written as a combination of a loss term $\mathcal{L}_{\text{Label}}$ based on the data, and another one  $\mathcal{L}_{\text{PDE}}$ based on the residuals of the PDEs:
\begin{eqnarray}
    \mathcal{L}\left ( \boldsymbol{\theta},\{ (\mathbf{x},t)^{(k))}\}_{k\in \mathcal{I}\cup \mathcal{J}}\right ) & = & \mathcal{L}_{\text{Label}}
    \left (\{ (\mathbf{x},t)^{(i))}\}_{i\in\mathcal{I}}\right )
    +\mathcal{L}_{\text{PDE}}\left (\{ (\mathbf{x},t)^{(j))}\}_{j\in\mathcal{J}}\right ) \nonumber \\
    & = & \frac{1}{n_L}\sum_{i=1}^{n_L} \| \underline{\mathbf{u}}_{\text{PINN}}^{(i)}-\mathbf{\underline{u}}_{\star}^{(i)} \|+\frac{1}{n_R}\sum_{j=1}^{n_R} \| \mathbf{r}_{\text{PINN}}^{(j)} \|,
    \label{eq:loss_function}
\end{eqnarray}
where $n_L$ is the size of a given set of input-output data sample {$\{ (\mathbf{x},t)^{(i)},\mathbf{\underline{u}}_{\star}^{(i)}\}_{i\in \mathcal{I}}$} (collected inside and/or at the boundaries of the training domain), $n_R$ is the size of the sample at which PDEs residuals are computed, and $\mathbf{\underline{u}}_{\cdot}$ may be a sub-component of the full output (e.g. $\mathbf{\underline{u}}_{\star}^{(i)} \equiv T_{\star}^{(i)}$ or $\mathbf{\underline{u}}_{\star}^{(i)} \equiv  (T_{\star} ^{(i)},\mathbf{v}_{\star,\partial \Omega \times \mathcal{D}}^{(i)} )$ if flow velocity information is also considered in the form of data sampled from the domain boundaries). For a given sample (e.g. fixed spatial and temporal coordinates), the residual is evaluated as the sum of an array of squared residuals of size equals to the {\em full} number of PDEs in the model. In our approach, the same weight is assigned to each residual of each equation of  the system (\ref{eq:PDEs_NS}). Note that the label and residuals sample sets are not necessarily the same, as their size and location may differ. A very recent work proposed to decompose $\mathcal{L}_{\text{Label}}$ in various terms corresponding for instance to the contribution of various data sources: e.g. the initial condition, the boundaries or the inside of the domain. This allows to dynamically assign some weights to each term in order to get a better error balance \cite{wang2020understanding}. 
The standard PINN model is therefore a grid-free approach as no mesh is needed. All the complexities of solving the model    are transferred into the optimization/training
stage of the neural network.
Updating the parameters requires the knowledge of the loss gradient $\partial \mathcal{L}( \boldsymbol{\theta}, (\mathbf{x},t)) /\partial \boldsymbol{\theta}$ that is computed thanks to the back-propagation algorithm \cite{rumelhart1986learning}. A particular algorithm from the stochastic gradient descent (SGD) class  with mini-batch updates based on an average of the gradients inside each block of $\MB$ examples:
\begin{equation}
    \boldsymbol{\theta}^{k} \leftarrow \boldsymbol{\theta}^{k-1} - \epsilon_k \frac{1}{\MB}\sum_{k^{'}=k\MB+1}^{(k+1)\MB} \partial \mathcal{L}( \boldsymbol{\theta}, (\mathbf{x},t)^{(k')}) /\partial \boldsymbol{\theta},
\end{equation}
is considered.
The great advantage of SGD update methods is that their convergence does not depend on the size of the training set, only on the number of updates and the richness of the training distribution \cite{bengio2012practical}. To be more specific, an Adam (for Adaptive moment estimation) optimizer \cite{kingma2014adam} is used,
which combines the best properties of the AdaGrad and RMSProp algorithms. Moreover, the parameters of the neural networks are randomly initialized using the Xavier scheme \cite{glorot2010understanding}. \\
Regarding the choice of the hyper-parameters for the approximate approximation, we have followed the literature advices \cite{bengio2012practical}. The learning rate of the Adam algorithm is $\epsilon$ (and will take different values depending on the epochs cycle) and the beta values are $\beta_1= 0.9$, and $\beta_2 = 0.999$.\\
In the following section we will investigate the use of standard PINNs for turbulent flows and then propose a new methodological development to improve its capability.

\section{Numerical improvements and experiments}
\label{sec:results}
\subsection{DNS database of turbulent Rayleigh-B\'enard convection with heated blocks}

\begin{figure}[!h]
\centering
\includegraphics[trim={1.0cm 0.1cm 0.5cm 0.0cm},clip,width=5.5cm]{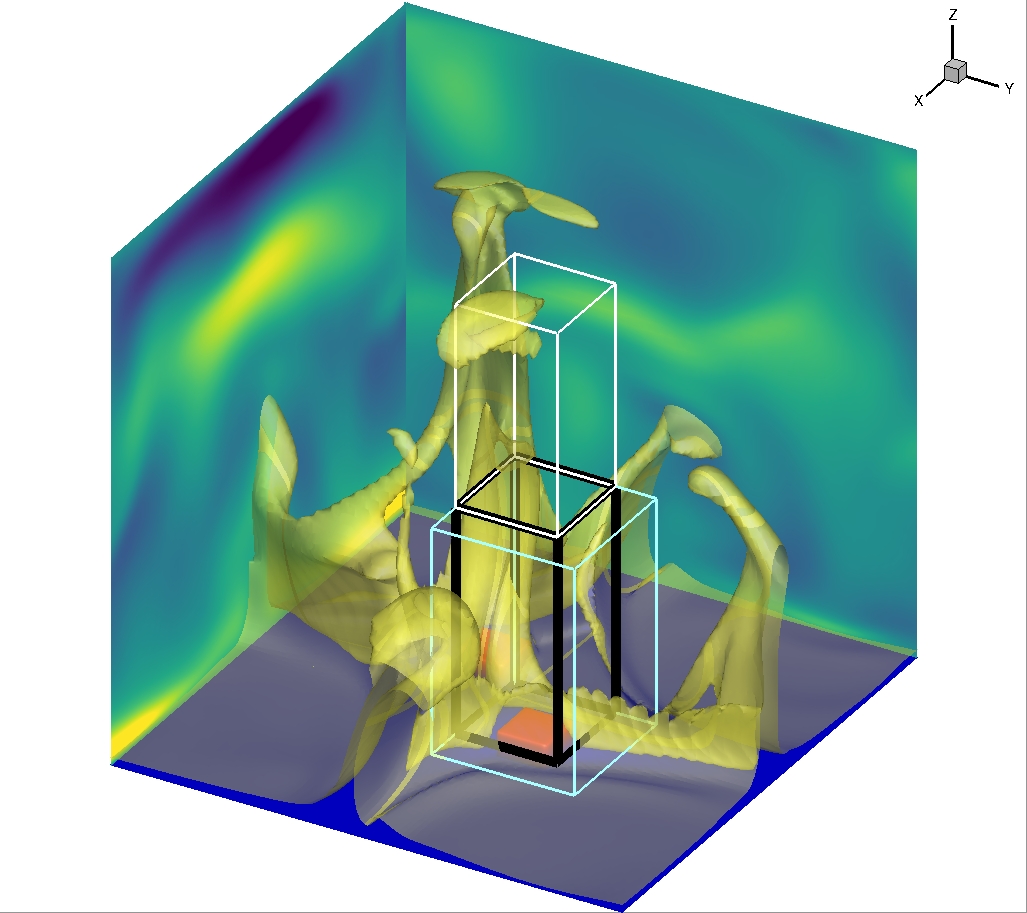}(a)\hspace{0.5cm}
\includegraphics[trim={2.5cm 2.5cm 2.5cm 2.5cm},clip,width=5.95cm]{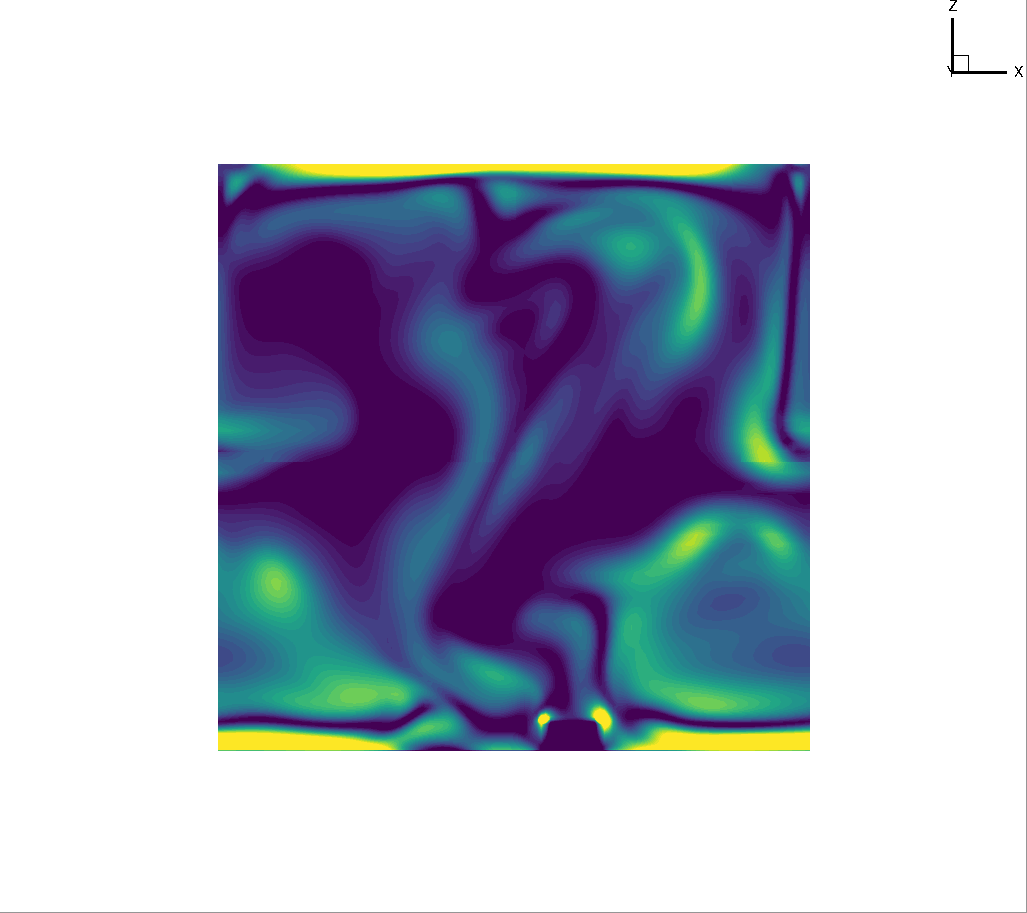}(b)
\caption{Rayleigh-B\'enard cavity flow with two square-based roughness elements (red cubes) attached to the heated bottom plate at $t=80.9$. Temperature isocontours  and normal velocities on the vertical planes (a) ; vorticity component $\omega_y$ in plane $(x,y=0.6H,z)$ (b). The  smaller domain $\Omega_{\text{PINN}}=[0.5H,0.7H]\times [0.5H,0.7H] \times [0.055H,0.5H]  $ over which the PINN model is trained, is depicted as a transparent box with thick black borders, located just above one of the roughness elements, so as to maximize the chance to contain some plumes (a). Other boxes with lighter borders indicate the positioning of padding regions, used later in the study.}
\label{fig:domain}
\end{figure}

{We consider a Rayleigh-B\'enard-like configuration made of a bi-periodic water layer heated from below ($Pr = 4.3$). The two horizontal plates are isothermal ($T_{\text{bottom}}=1$; $T_{\text{top}}=0$). {Previous studies had shown that surrogate modeling of PINN type performs better when the training domain encompasses lively flow structures with non-zero gradients. Based on this experience, we wish to propose a configuration producing a more organized natural convection in order to easily position our training domain. To this end,} two heated square-based blocks at $T_{\text{bottom}}$ are closely placed on the bottom plate in order to better localize the flow. They are aligned along one of the main diagonals. The resulting flow will be dominated by two main plumes developing over the blocks. They interact with each others in a complex pattern and swirl before impacting the ceiling (as seen on Figure \ref{fig:domain}). }

The computational domain is a cube of width equal to $H=1$, so the computational domain is defined as $\Omega = [0,H]\times[0,H]\times[0,H]$, cf. Fig. \ref{fig:domain}. {The height of the square-based blocks is equal to $h = 0.05H$.} Their base spans $(0.1H \times 0.1H)$, and their centers are located at $(x=0.4,y=0.4)$ and $(x=0.6,y=0.6)$, respectively. 

\begin{figure}[!h]
\centering
\includegraphics[width=1.025\linewidth]{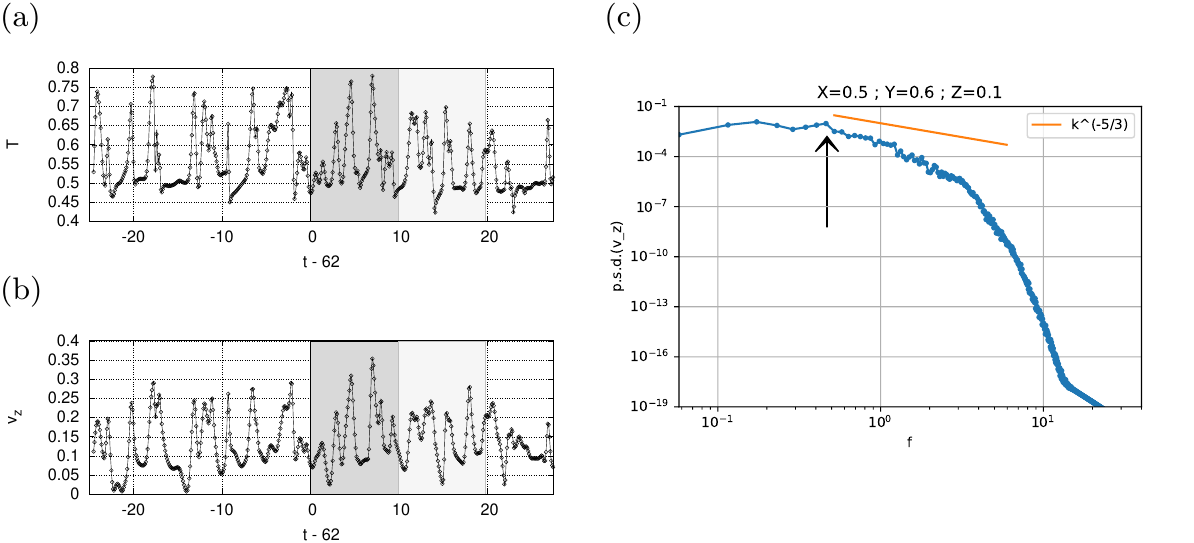}
\caption{DNS-predicted temperature (a) and vertical velocity (b) at location $(x=0.5H,y=0.6H,z=0.1H)$. {Temporal power spectrum of temperature at location $(x=0.5H,y=0.6H,z=0.1H)$. The frequency of energy maximum is marked by an arrow  at $f_{max}=0.462$} (c). The grey zones delimit the time windows over which the PINN models are trained: dark grey ($\Delta T_s$) and dark/light grey ($\Delta T_l$).}
\label{fig:timeseries}
\end{figure}

The Rayleigh number of the studied test case is  equal to $Ra=2\cdot 10^7$ leading to a turbulent flow regime (Figure \ref{fig:timeseries}). The DNS database is obtained using the in-house numerical solver  SUNFLUIDH. It is based on a finite volume approach on staggered grids. A semi-implicit scheme and a pressure-correction algorithm for the velocity–pressure coupling \cite{GUERMOND20066011} are combined together to achieve a second-order time accuracy. The resulting Poisson's equation is solved by  a multi-grid method.
The solid blocks are modeled through a loop truncation technique. A domain decomposition method using MPI is applied for parallel computation. The code has been validated in the context of turbulent Rayleigh-Bénard convection with roughness \cite{belkadi_jfm_2021}.
DNS calculations are done using $(2\times 2 \times 2)$ subdomains of $64^3$ cells each with a constant convective time step equal to 2.5e-3. 

{We retain a training domain about $55$ times smaller than the DNS computational domain (figure \ref{fig:domain}). It is a box-shaped volume placed  over one of the two roughness elements, of dimension: \{$\Omega_{\text{PINN}}=[0.5H,0.7H]\times [0.5H,0.7H] \times [0.055H,0.5H]$, 
 containing $(n_x=26\times n_y=26 \times n_z=38)$ grid points in the $(x,y,z)$ frame of reference, cf. Table (\ref{tab:testcases})}. The domain over which the PINN model is going to be trained, therefore spans 20\% along each $x-$ and $y-$direction and 45\% along the vertical $z-$direction. The training domain is large enough to cover the entirety of the plume formed at the block in its spatial ascent over half the height of the cavity, and the fluid it entrains in its near field.

The DNS databases are made of a collection of fields spanning a maximum time length of 19.8 convective time units. Typical temperature and vertical velocity time series are displayed on Figure (\ref{fig:timeseries}). In addition to the turbulent character of the flow, the power spectrum of the vertical velocity shows the dominant shedding frequency  $f_{max}$ of the plume emission. This indicates that  the training domain time length typically includes about 10 plume rise times through the studied domain.

\begin{table}[hbt]
\centering
\small
	\centering
	\begin{tabular}{ l c c c c c c c}
\toprule
Database & $Ra$ & size & $\Delta t$ & snapshots & resolution & time interval &  domain \\
\midrule
1Db1  & $2\cdot 10^7$ & $2.5688\mathrm{e}6$ & 0.1 & 100	& $26\times 26 \times 38$ & $\Delta T_s$ & $\Omega_{\text{PINN}}$ \\
1Db2 & $2\cdot 10^7$ &  $1.2844\mathrm{e}6$ &   0.2 & 50	& $26\times 26 \times 38$ & $\Delta T_s$ & $\Omega_{\text{PINN}}$ \\
1Db3  & $2\cdot 10^7$ &  $6.422\mathrm{e}5$ &   0.4 & 25	& $26\times 26 \times 38$ & $\Delta T_s$ & $\Omega_{\text{PINN}}$ \\
\midrule
2Db1 & $2\cdot 10^7$ & $5.111912\mathrm{e}6$ & 0.1 & 200	& $26\times 26 \times 38$ & $\Delta T_l$ & $\Omega_{\text{PINN}}$ \\
2Db2 & $2\cdot 10^7$ &  $2.5688\mathrm{e}6$ &   0.2 & 100	& $26\times 26 \times 38$ & $\Delta T_l$ & $\Omega_{\text{PINN}}$ \\
\midrule
3Db1 & $2\cdot 10^9$ &  $2.0673972\mathrm{e}7$ &   0.054 & 83	& $66\times 34 \times 111$ & $\Delta T_m$ & $\Omega_{\text{PINN}^{\text{r}}}$ \\
\bottomrule
\end{tabular}
\caption{Specifics of DNS-extracted databases. The short, long and medium time intervals are defined as: $\Delta T_s=[62,71.9]$, $\Delta T_l=[62,81.8]$ and $\Delta T_m=[287.572,292]$ while $\Delta t$ refers to the time between two successive snapshots taken in the time intervals.}
\label{tab:testcases}
\end{table}

\subsection{PINNs hyperparameters}

\begin{table}[!ht]\small
	\centering
	\begin{adjustbox}{center}
	\begin{tabular}{c c l l }
    \toprule
\multicolumn{1}{c}{Training cycles} &\multicolumn{1}{l}{MB}& \multicolumn{1}{l}{epochs } & \multicolumn{1}{l}{LR}\\
1 & 2000 & 50 & 1e-03\\
2 & 2000 & 62 & 6.683e-4 \\
3 & 2000 & 138 & 2.992e-4 \\
4 & 2000 & 309 & 1.337e-4 \\
5 & 2000 & 309 & 5.98e-5 \\
6 & 2000 & 309 & 1e-5 \\
7 & 2000 & 160 & 1e-6 \\
\bottomrule
    \end{tabular}
    \end{adjustbox}
    \caption{PINNs training hyper-parameters. For each PINN model, the training is made of seven subsequent cycles. }
	\label{tab:hyperparameter}
\end{table}

In this section, the goal is to compare PINNs prediction with the DNS reference and understand how the PINN model can be made accurate, while as data-frugal as possible. The results are presented both in terms of training/validation for the chosen turbulent RB cavity with partial data information.

The PINN architecture contains $\layer=10$ hidden layers (unless mentioned otherwise) of size $n^{l=1,\ldots ,\layer}=300$ neurons each, so we have $\mathcal{A}=(n^0=4,n^1=300,\ldots,n^{\layer}=300,n_{\mathbf{u}}=6)$, for a total of $\| \boldsymbol{W}\|=813\mathrm{e}3$ weights to be calibrated\footnote{Here is the computation of the weights goes according to: $\| \boldsymbol{W}\|=n^0\times n^1+\sum_{i=2}^{\layer} n^i*n^{i-1}+n^{\layer}\times n_{\mathbf{u}}$, computation including the biases quantities is: $\|  \boldsymbol{\theta}\|=(n^0+1)\times n^1+\sum_{i=2}^{\layer} n^i*(n^{i-1}+1)+(n^{\layer}+1)\times n_{\mathbf{u}}$.\\ If we compare the number of degrees of freedom (dof) between the DNS (in the full domain, i.e. $5\times 128^3$ because of the 5 unknown fields in the NS under Boussinesq system to be solved) and the PINN in the training domain, we get a ratio of 18.5, if we consider the dof of the DNS in the training domain, the ratio drops to 0.23 in favor of the DNS. Considering now the storage: - storing the PINN parameters, so that an approximation of the DNS fields may be recovered very efficiently, amounts to 4.3Mb of data}.

As for the training procedure, the results reported in the following are obtained after seven cycles, each of them being made of a certain number of consecutive epochs of the stochastic gradient Adam optimizer with various learning rates, cf. Table (\ref{tab:hyperparameter}), 
each epoch corresponding to one pass through an entire dataset. 
The total number of iterations of the Adam optimizer is therefore given by the total number of epochs (i.e. 1500) times the size of the training data used, divided by the mini-batch size. The mini-batch size we have used is $\MB=2\cdot 10^3$ and the number of data points are clearly specified in the following on a case by case basis. 
The training is performed on our laboratory Lab-IA cluster with a single NVIDIA Tesla V100 32GB GPU.\\ 
The details of the various databases that have been extracted from DNS data and used to train different PINN models are written down in Tables (\ref{tab:testcases},\ref{tab:training}). Table (\ref{tab:testcases}) describes the spatial and temporal domains and their resolutions as well as the size of each database. The spatial resolution is equal to (or half of)  the full DNS while temporal resolution is much coarser than the one of the full DNS.
Here the size refers to the number of discrete points $(\mathbf{x},t)$ at which the solution of the system of equations (\ref{eq:PDEs_NS}) is known.\\ Table (\ref{tab:training}) present most of the proposed PINN models in terms of the choice of their training and testing databases. {In the paper, the} data labels of {total} size $N_L$ is the following set $ \{ ((\mathbf{x},t)^{(i)},T^{(i)},\overline{T}^{(i)},\mathbf{v}_{\partial \Omega^{\dag} \times \mathcal{D}}^{(i)} ) \}_{i=1}^{N_L}$, where $ \mathbf{v}_{\partial \Omega^{\dag}}^{(\cdot)}$ are fluid velocity components collected at the boundaries of the rectangular blocks (excluding the top face). { $N_T$ refers to the total size of the testing database that is always disjoint from the training database}. \\ 
{Unless mentioned otherwise,} for these models the training database is common for the $N_L$ data labels (and $N_R$ data points) at which $\mathcal{L}_{\text{Label}}$ (and $\mathcal{L}_{\text{PDE}}$) loss is evaluated, respectively. For instance, if we refer to $\mathcal{K}_{\mathrm{1Db1}}=\left \{ k_l \right \}_{l=1}^{\left | \mathcal{K}_{\mathrm{1Db1}} \right |}$ as the set of data points indices from the 1Db1 database of cardinality $\left | \mathcal{K}_{\mathrm{1Db1}} \right | = 2.5688\mathrm{e}6$, then for the case 1C6 we consider $\mathcal{I} \subset \mathcal{K}_{\mathrm{1Db1}}$, the subset with cardinality $\left | \mathcal{I} \right | = 5\mathrm{e}5$ and containing {\em any} elements $k_i \in \mathcal{K}_{\mathrm{1Db1}}$. For the residuals, the subset is then taken the same, i.e.  $\mathcal{J}\equiv \mathcal{I}$. Nevertheless, during training, mini-batches of data and points are {\em independently} randomly selected among these subsets $\mathcal{I}$ and $\mathcal{J}$, meaning that the points at which labeled data are used and the points at which residuals are computed are not necessarily collocated within the training domain. This is illustrated in the grey region of Figure (\ref{fig:sampling}), showing an example of mini-batch sampling.

\begin{table}[!ht]\small
	\centering
	\begin{adjustbox}{center}
	\begin{tabular}{l c c c c c}
    \toprule
\multicolumn{1}{c}{PINN model} & $\layer$ & \multicolumn{2}{c}{Training}&
\multicolumn{2}{c}{Testing} \\
\midrule
\multicolumn{2}{c}{} & \multicolumn{1}{c}{size $(N_L,N_R)$} &\multicolumn{1}{c}{database} &  \multicolumn{1}{c}{size $(N_T)$}& \multicolumn{1}{c}{database} \\

{1C3} & 6 & {$(2\mathrm{e}6,2\mathrm{e}6)$} &
{1Db1} &
{$5.688\mathrm{e}5$}& {1Db1} \\

{1C4} & 8 & {$(2\mathrm{e}6,2\mathrm{e}6)$} &
{1Db1} &
{$5.688\mathrm{e}5$}& {1Db1}  \\

{\bf 1Ref} & 10 &  $\boldsymbol{(2\mathrm{e}6,2\mathrm{e}6)}$ &
{\bf 1Db1} &
{ $\boldsymbol{5.688\mathrm{e}5}$}& {\bf 1Db1} \\

{1C5} & 10 & {$(1\mathrm{e}6,1\mathrm{e}6)$} &
{1Db1} &
{$5.688\mathrm{e}5$}& {1Db1} \\

{1C6} & 10 &{$(5\mathrm{e}5,5\mathrm{e}5)$} &
{1Db1} &
{$5.688\mathrm{e}5$}& {1Db1}  \\

{1C7} & 10 & {$(1\mathrm{e}6,1\mathrm{e}6)$} &
{1Db2} &
{$2.844\mathrm{e}5$}& {1Db2}  \\

{1C8} & 10 &{$(5\mathrm{e}5,5\mathrm{e}5)$} &
{1Db3} &
{$1.422\mathrm{e}5$}& {1Db3} \\

\midrule

{\bf 2Ref} & 10 & $\boldsymbol{(2\mathrm{e}6,2\mathrm{e}6)}$ &
{\bf 2Db2} &
{ $\boldsymbol{1.131912 \mathrm{e}6}$}& {\bf 2Db1} \\

{2C2} & 10 &{$(1\mathrm{e}6,1\mathrm{e}6)$} &
{2Db2} &
{$1.131912 \mathrm{e}6$}& {2Db1} \\

{2C3} & 10 & {$(5\mathrm{e}5,5\mathrm{e}5)$} &
{2Db2} &
{$1.131912 \mathrm{e}6$}& {2Db1}  \\

\bottomrule
    \end{tabular}
    \end{adjustbox}
    \caption{Training and testing details of the $\layer$-layer PINN models considered in this study. Each number of points spans space/time domain. 
    When a single database is mentioned, it means that this database is used for the labeled data {\em and} for the residual points (i.e. residuals are evaluated at some DNS grid points).
    DNS databases used for training and testing are detailed in Table \ref{tab:testcases}. Reference models are highlighted in bold.}
	\label{tab:training}
\end{table}

\subsection{PINNs results} 
\begin{figure}[!h]
\centering
\includegraphics[width=11cm]{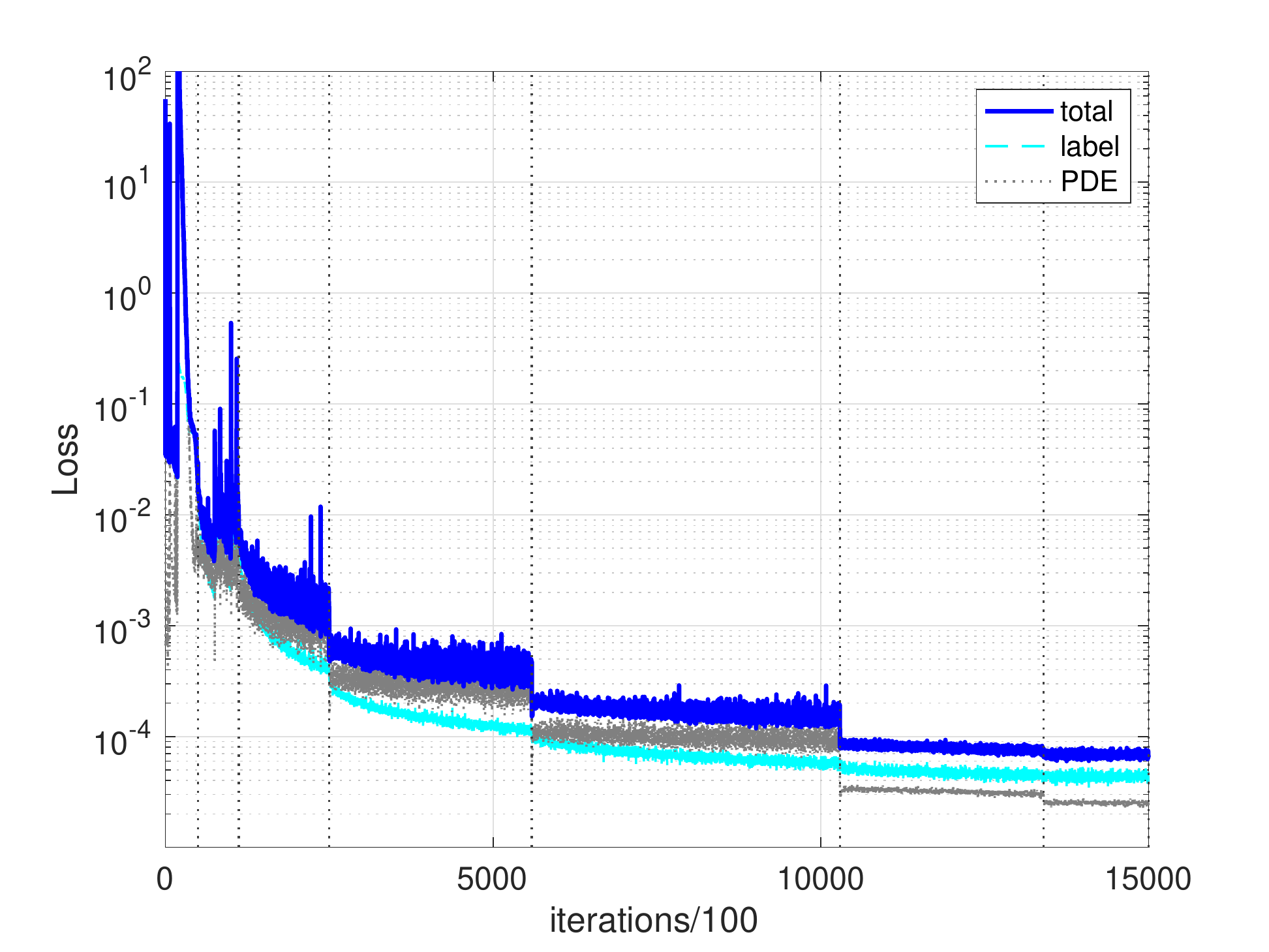}
\caption{Evolution of the loss function $\mathcal{L}$ -- during the training of 1Ref model -- against the stochastic gradient descent optimization algorithm iterations. The total loss is decomposed into its label and PDE contributions. }
\label{fig:loss}
\end{figure}

We first present the best results obtained, for the 1Ref case trained with a large data sample over a short time period, i.e. 2e6 data points spanning the  space/time domain (i.e. 2e4 points per DNS snapshot). Figure \ref{fig:loss} shows the convergence of the loss function (cf. Eq. \ref{eq:loss_function}) against the iterations of the optimization algorithm for the 1Ref model. Vertical dashed lines separate the different training cycles. The loss is decomposed into its label and residual contributions. It is interesting to notice that the two contributions behave quite differently both in terms of decay and magnitude. While the residual errors dominate over the label errors at early stages, they become lower around the million iteration. Moreover, the convergence of the label errors is much more regular and progressive than the residual ones, which are very much impacted by the changes in the learning rate but much less by the algorithm iterations. Overall, the convergence is satisfactory and the total loss evaluated in this case over $2\mathrm{e}6$ points is less than $10^{-4}$.

\begin{figure}[!h]
\centering
\includegraphics[trim={1cm 5cm 1cm 5cm},clip,width=4.5cm]{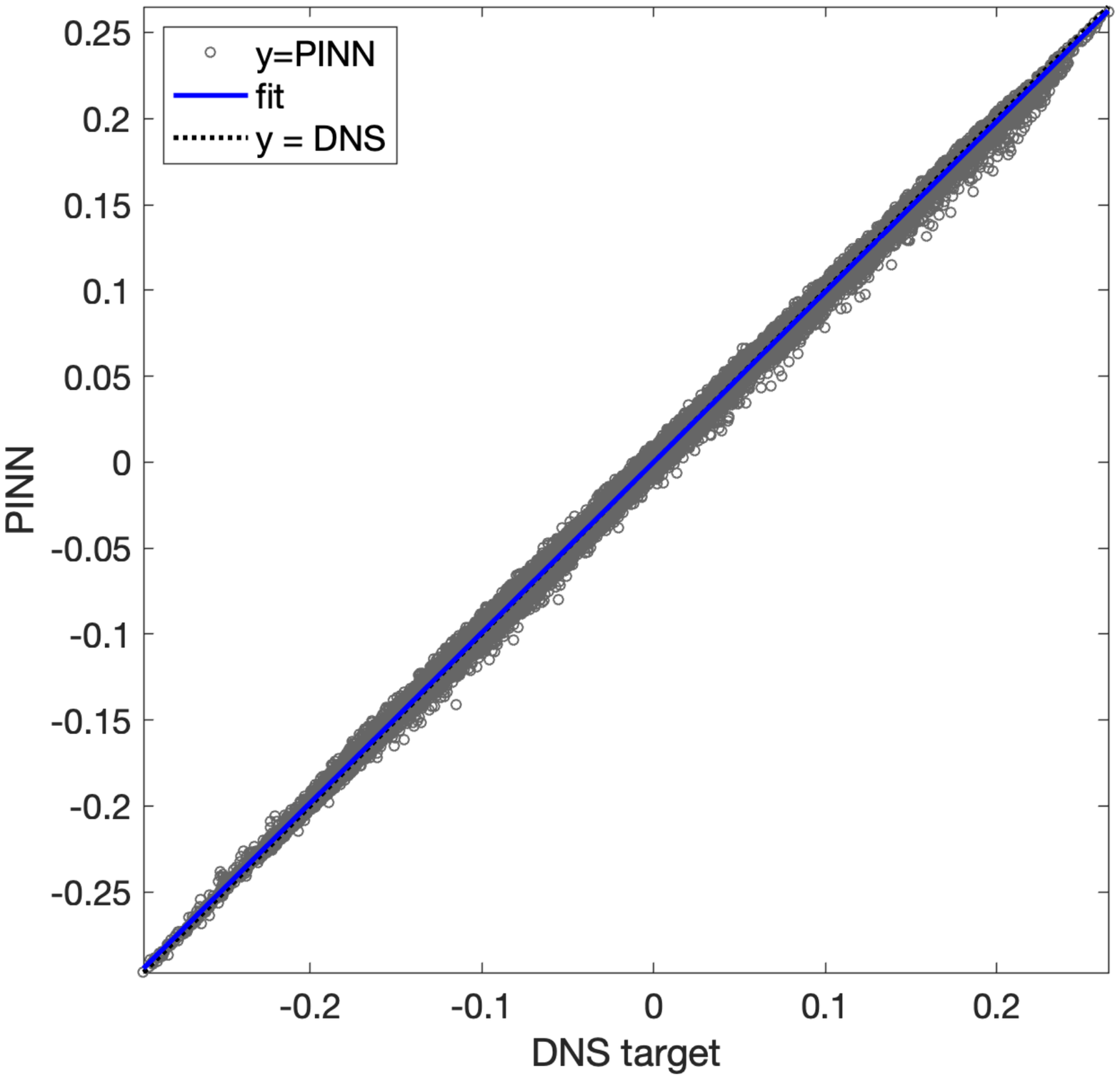}(a)
\includegraphics[trim={1cm 5cm 1cm 5cm},clip,width=4.5cm]{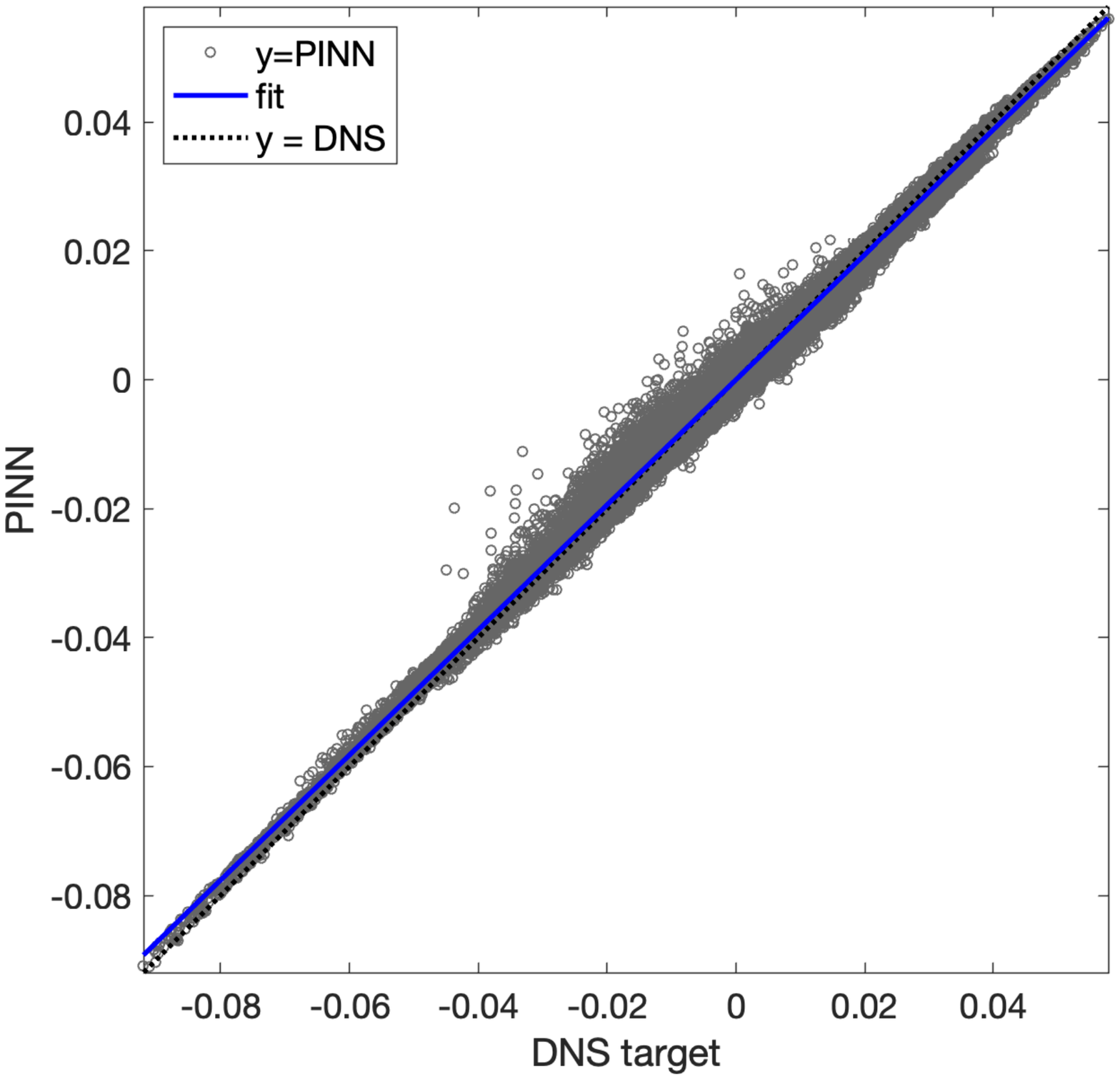}(b)\\
\includegraphics[trim={1cm 5cm 1cm 5cm},clip,width=4.5cm]{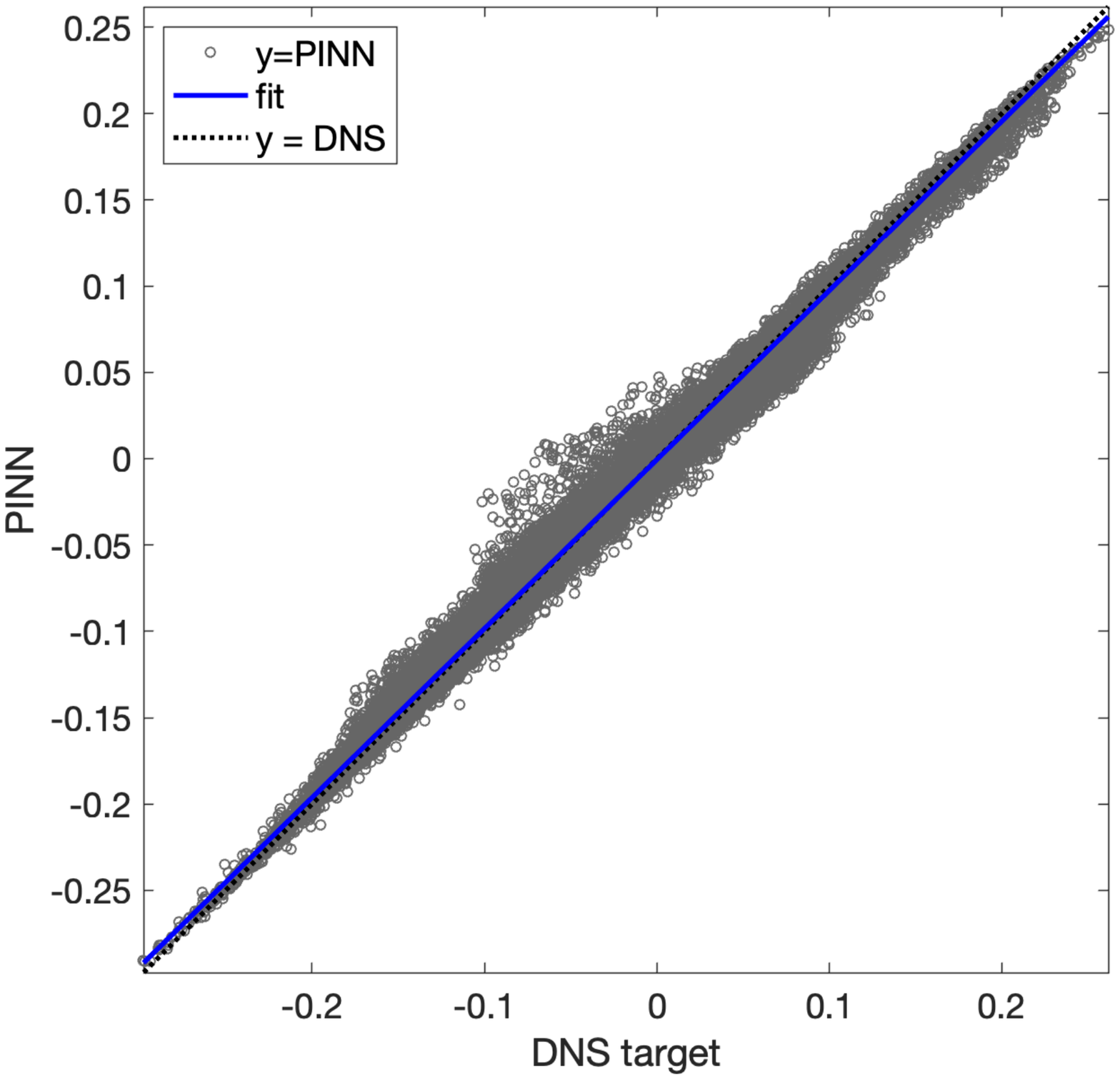}(c)
\includegraphics[trim={1cm 5cm 1cm 5cm},clip,width=4.5cm]{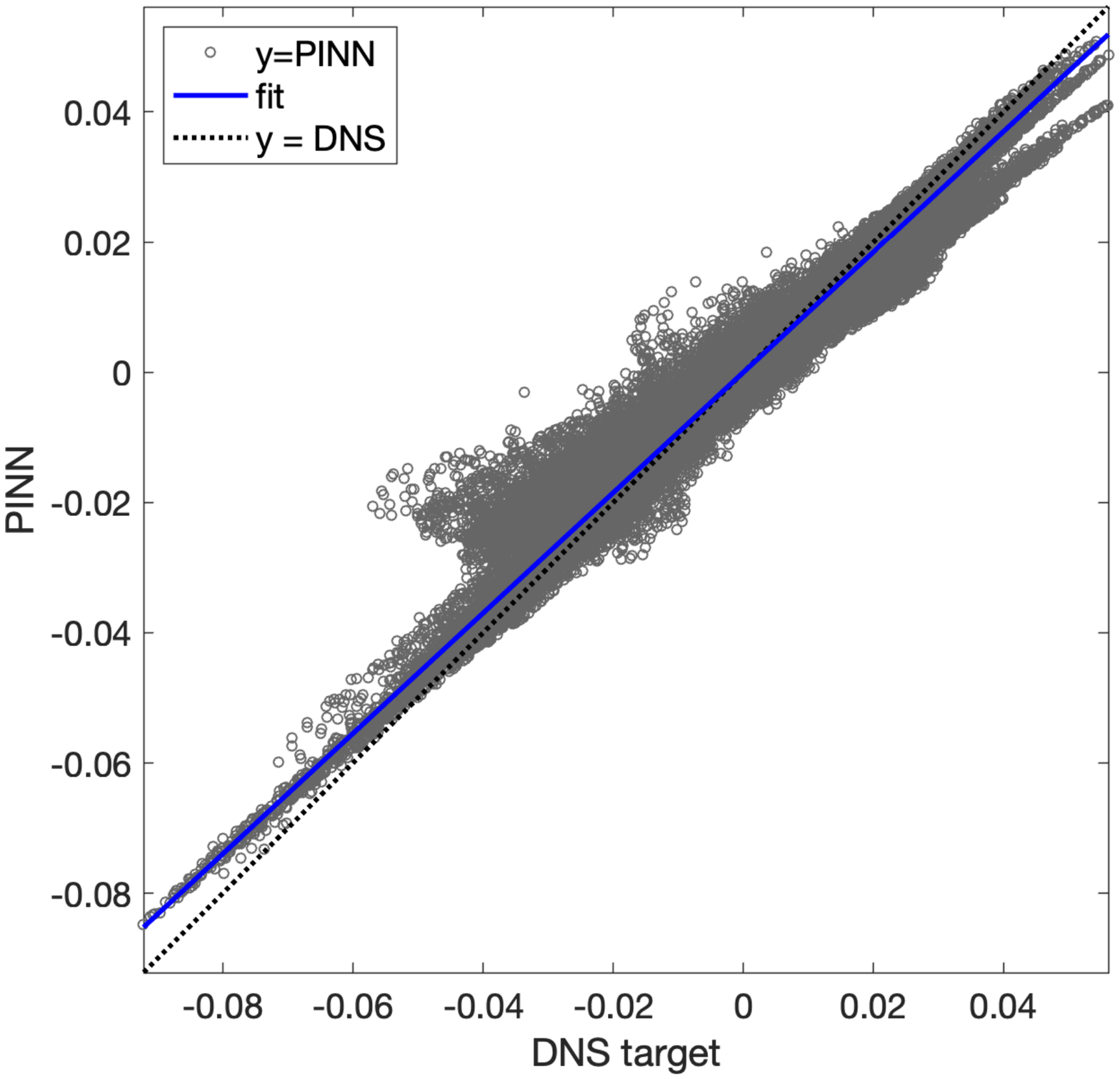}(d)\\
\includegraphics[trim={1cm 5cm 1cm 5cm},clip,width=4.5cm]{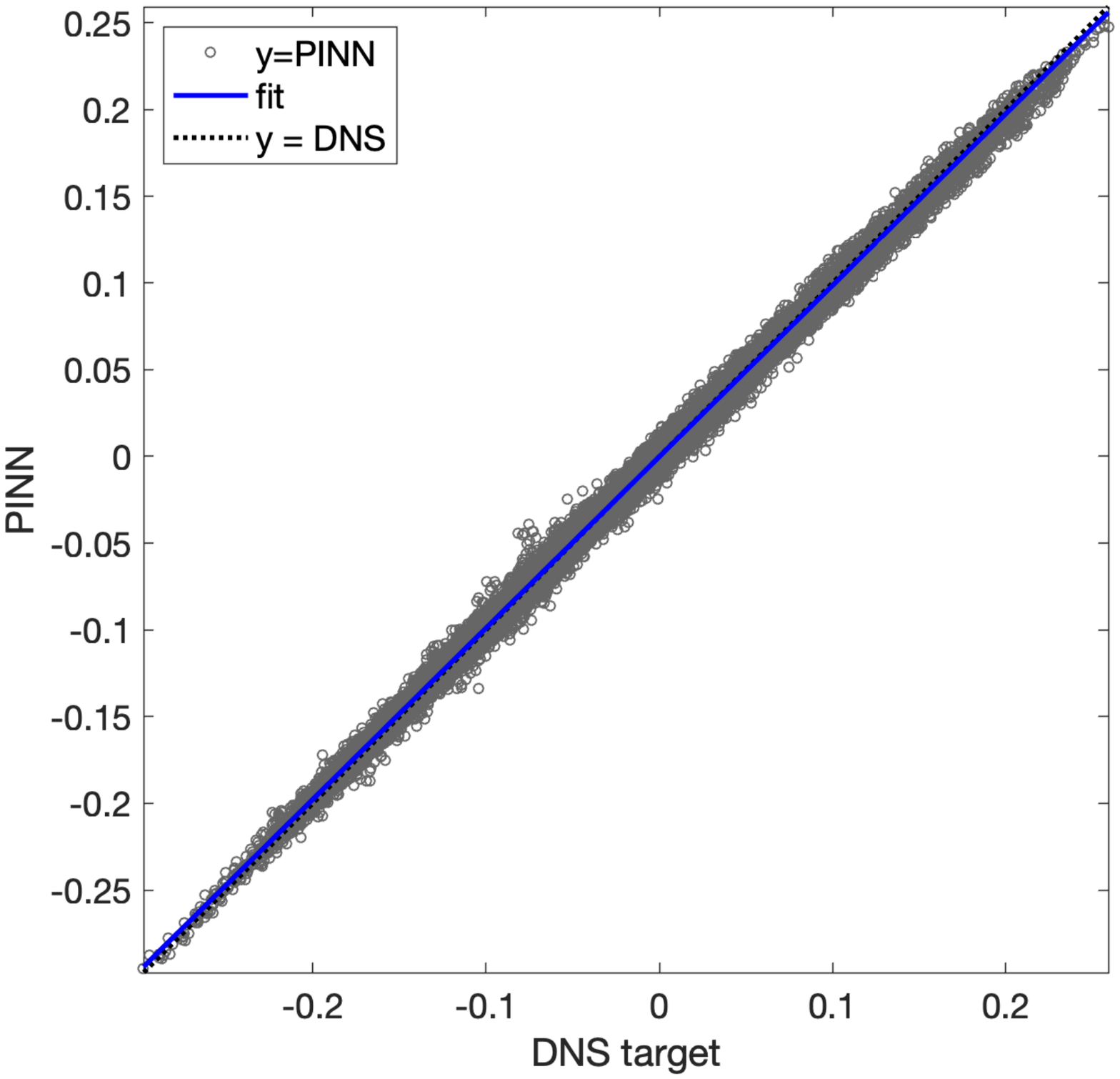}(e)
\includegraphics[trim={1cm 5cm 1cm 5cm},clip,width=4.5cm]{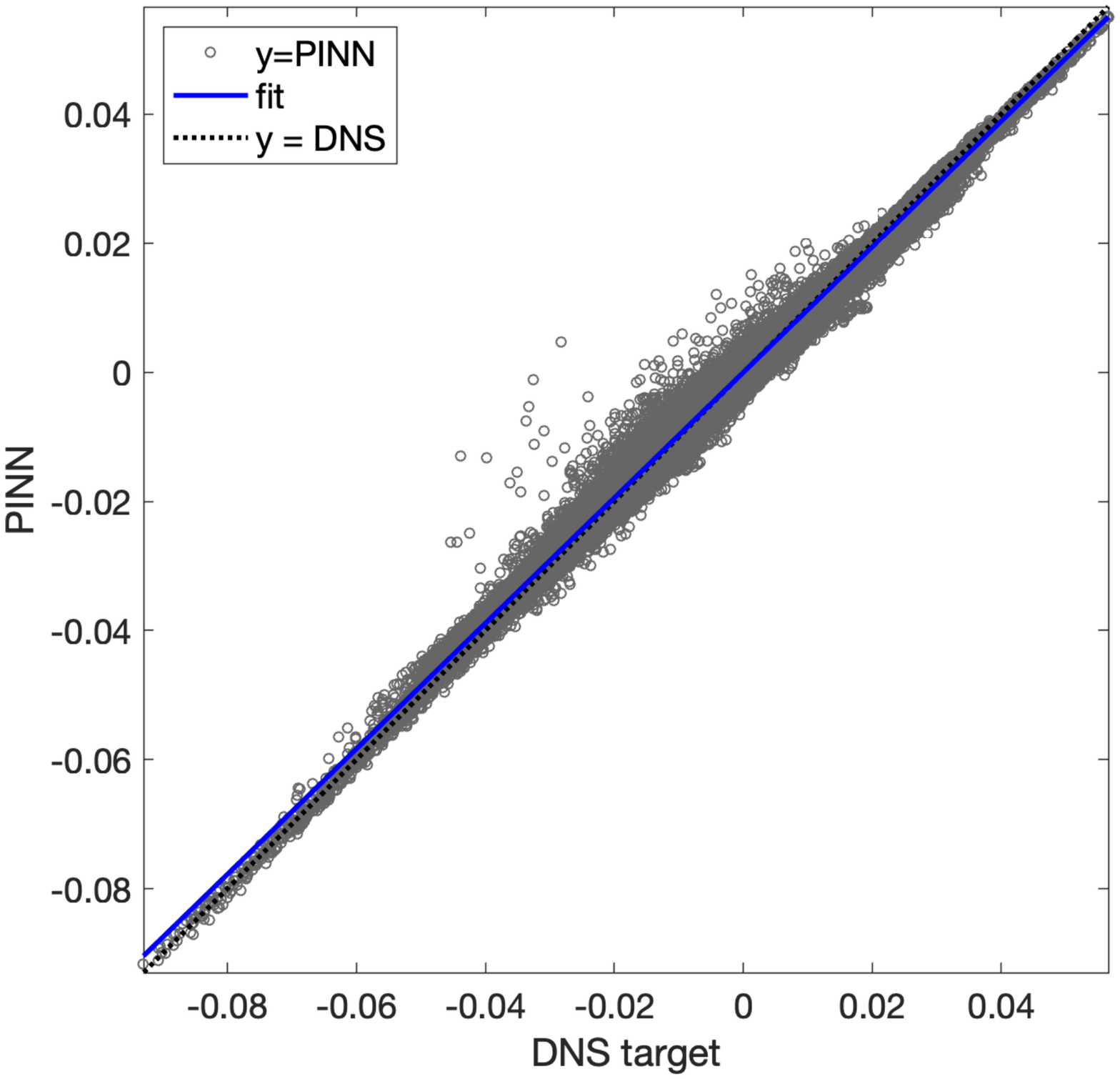}(f)
\caption{Examples of scatter plots comparing DNS and predicted values from most accurate 1Ref (a-b), least accurate 1C7 (c-d) and padded 2C6 (e-f) PINN models, with left column: in-plane $x-$axis flow velocity $\mathrm{v}_x$, and right column: fluid pressure $p$.}
\label{fig:scatters}
\end{figure}

The specifics of the accuracy of the PINN 1Ref model for each flow field are summarized in Table (\ref{tab:accuracy2}). We see that the results are excellent with very small errors and high correlations between the PINN model and the DNS for each field. We note that the accuracy is a bit lower for the pressure field, and this finding will be consistent across all of our numerical experiments. We also note that the accuracy on the temperature prediction is slightly lower than the one obtained from a plain DNN (noted 1Ref DNN in the table) with 10 layers and a loss function given by Eq. (\ref{eq:standard_loss}). A multi-output regression providing linear velocity and pressure predictions based on spatial/temporal coordinates and temperature field regressors show poor agreement especially for the first two components of velocity, which are known to be less correlated to the temperature field than the vertical one.  \\
\begin{table}[!ht]\small
	\centering
	\begin{adjustbox}{center}
	\begin{tabular}{l c c c c c c }
    \toprule
\multicolumn{1}{c}{Model} &  \multicolumn{5}{c}{Accuracy}\\
\midrule
\multicolumn{1}{c}{1Ref PINN}  &\multicolumn{1}{c}{RMSE} & \multicolumn{1}{c}{MAE}& \multicolumn{1}{c}{$\mu$ error } &\multicolumn{1}{c}{$\sigma$ error}& \multicolumn{1}{c}{$R_{corr}$ } & \multicolumn{1}{c}{$R^2$}\\

{$T$} & 3.336e-03 & 2.05e-03 & 0.01 & 0.4 & 9.993e-01 & 9.986e-01\\

{$\mathrm{v}_x$} & 1.778e-03 & 1.260e-03 & 0.8 & 0.8 & 9.997e-01 & 9.993e-01 \\

{$\mathrm{v}_y$} & 1.953e-03 & 1.386e-03 & 0.7 & 1.0 & 9.996e-01 & 9.991e-01 \\

{$\mathrm{v}_z$} & 3.316e-03 & 2.064e-03 & 0.4 & 0.5 & 9.998e-01 & 9.996e-01 \\

{$p$} & 8.904e-04 & 6.341e-04 & -- & 2.9 & 9.989e-01 & 9.971e-01 \\

\midrule
\multicolumn{1}{c}{1Ref DNN}  &\multicolumn{1}{c}{RMSE} & \multicolumn{1}{c}{MAE}& \multicolumn{1}{c}{$\mu$ error } &\multicolumn{1}{c}{$\sigma$ error}& \multicolumn{1}{c}{$R_{corr}$ } & \multicolumn{1}{c}{$R^2$}\\

 {$T$} & 9.035e-04  & 6.362e-04 & 0.004 & 0.03 & 9.999e-01 & 9.999e-01\\

\midrule
\multicolumn{1}{c}{1Ref MOR}  &\multicolumn{1}{c}{RMSE} & \multicolumn{1}{c}{MAE}& \multicolumn{1}{c}{$\mu$ error } &\multicolumn{1}{c}{$\sigma$ error}& \multicolumn{1}{c}{$R_{corr}$ } & \multicolumn{1}{c}{$R^2$}\\

{$\mathrm{v}_x$} & 5.95e-02  & 4.73e-02 & 0.16 & 51.7 & 4.83e-01 & 2.33e-01 \\

{$\mathrm{v}_y$} & 5.25e-02  & 4.11e-02 & 0.11 & 40.5 & 5.96e-01 & 3.55e-01 \\

{$\mathrm{v}_z$} & 1.121-e01 & 9.02e-02 & 0.06 & 24 & 7.61e-01 & 5.8e-01 \\

{$p$} & 1.08-e02 & 8.2-e03 & 0.03 & 22.15 & 7.78e-01 & 6.05e-01 \\

\bottomrule
    \end{tabular}
    \end{adjustbox}
    \caption{Accuracy details (root mean squared error: RMSE, mean absolute error: MAE, the correlation coefficient: $R_{corr}$ and the coefficient of determination : $R^2$) of the 1Ref PINN model for each of the flow fields. Mean ($\mu$) and standard deviation ($\sigma$) errors are expressed as percentage. Relative error of the mean pressure is not computable as the pressure signals are centered (zero-mean) prior to be compared. The 1Ref DNN temperature prediction and a multi-output linear regression (MOR) are also provided for comparison. }
	\label{tab:accuracy2}
\end{table}
Inspired by these promising results, more models are trained to understand the effect of - the architecture complexity, - the size and sampling frequency of the training dataset and - the change in the time acquisition range. 
Table (\ref{tab:accuracy}) provides a summary of the accuracy of the different PINN models considered. The model names -- starting with a 1$\cdot$, refer to the cases with training and testing over a short time window $\Delta T_s$, while the names -- starting with a 2$\cdot$ refer to the cases with a longer time window $\Delta T_l$. Another major difference resides in the way models are tested. The first models are always tested on an independent sample of points coming from the {\em same}  database as the one used for training. For instance, model 1C7 relies on the 1Db2 database, which contains $1.2844\mathrm{e}6$ data points collected over space/time according to the specifics of Table (\ref{tab:testcases}), and from which $1\mathrm{e}6$ points are randomly selected from the database for training and the remaining $2.844\mathrm{e}5$ points are kept for testing. The second models are always tested on an independent sample of points coming from a {\em different}  database (e.g. denser in time) than the one used for training.\\
We see how the 10-layer PINN architecture with large amount of data and residual evaluations provide the best overall results for the first models. With less data sampled in space (i.e. conserving the same temporal resolution), cases 1C5-6 show some reasonable decline of their accuracy. With less data sampled in time (i.e. conserving the same averaged spatial sampling frequency), cases 1C7-8 exhibit a worse decay of their predictive capability. Best and worse predictions are illustrated in Figure (\ref{fig:scatters}), where reference and predicted $\mathrm{v}_x$ flow velocity and fluid pressure are represented together with their regression fit.

\begin{table}[!ht]\small
	\centering
	\begin{adjustbox}{center}
	\begin{tabular}{l c c c c c c }
    \toprule
\multicolumn{1}{c}{PINN model} & \multicolumn{5}{c}{Accuracy}\\
\midrule
\multicolumn{1}{c}{} &\multicolumn{1}{c}{aRMSE} & \multicolumn{1}{c}{aMAE}& \multicolumn{1}{c}{$\mu$ error } &\multicolumn{1}{c}{$\sigma$ error}& \multicolumn{1}{c}{$aR_{\mathrm{corr}}$ } & \multicolumn{1}{c}{$aR^2$}\\

{1C3} &  3.701e-03 & 2.418e-03 & 0.4 & 1.6 & 9.986e-01 & 9.969e-01 \\

{1C4} &  2.468e-03 & 1.612e-03 & 0.5 & 1.2 & 9.994e-01 & 9.985e-01 \\

{\bf 1Ref} &   {\bf 2.255e-03} & {\bf 1.479e-03} & {\bf 0.5} & {\bf 1.1} & {\bf 9.995e-01} & {\bf 9.988e-01}\\

{1C5} &  3.156e-03 & 2.125e-02 & 0.6 & 1.6 & 9.990e-01 & 9.976e-01 \\

{1C6} &  7.134e-03 & 4.802e-03 & 0.5 & 3.3 & 9.944e-01 & 9.880e-01 \\

{1C7} & 4.952e-03 & 2.841e-03 & 0.5 & 2.5 & 9.968e-01 & 9.928e-01 \\

{1C8} &  4.485e-03 & 2.999e-03 & 0.3 & 1.7 & 9.976e-01 & 9.950e-01 \\

\midrule

{\bf 2Ref} &  {\bf 4.163e-03} & {\bf 2.469e-03} & {\bf 0.5} & {\bf 1.7} & {\bf 9.982e-01} & {\bf 9.961e-01}\\

{2C2}  & 6.313e-03  & 4.271e-03 & 0.5 & 2.4 & 9.959e-01 & 9.914e-01 \\

{2C3} & 7.571e-03 & 5.156e-03 & 0.5 & 2.8 & 9.943e-01 & 9.881e-01 \\

\bottomrule
    \end{tabular}
    \end{adjustbox}
    \caption{Accuracy of PINN models compared to the DNS simulation. Statistics are computed from the available testing dataset and collected for each component of the flow fields $\mathbf{u}= \left (\mathbf{v},p,T \right)$. They are then averaged for aRMSE, e.g. $\text{aRMSE} = \frac{1}{n_{\mathbf{u}}}\sum_{j=1}^{n_{\mathbf{u}}} \big ( \frac{1}{N_T}\sum_{i=1}^{N_T} ( {\mathbf{u}}_{\text{PINN},j}^{(i)}-\mathbf{{u}}_{\text{DNS},j}^{(i)} )^2\big )^{1/2}$, aMAE, $aR_{\mathrm{corr}}$ and $aR^2$.}
	\label{tab:accuracy}
\end{table}
\begin{figure}[!h]
\centering
\includegraphics[trim={5.5cm 7cm 5.5cm 7cm},clip,width=3cm]{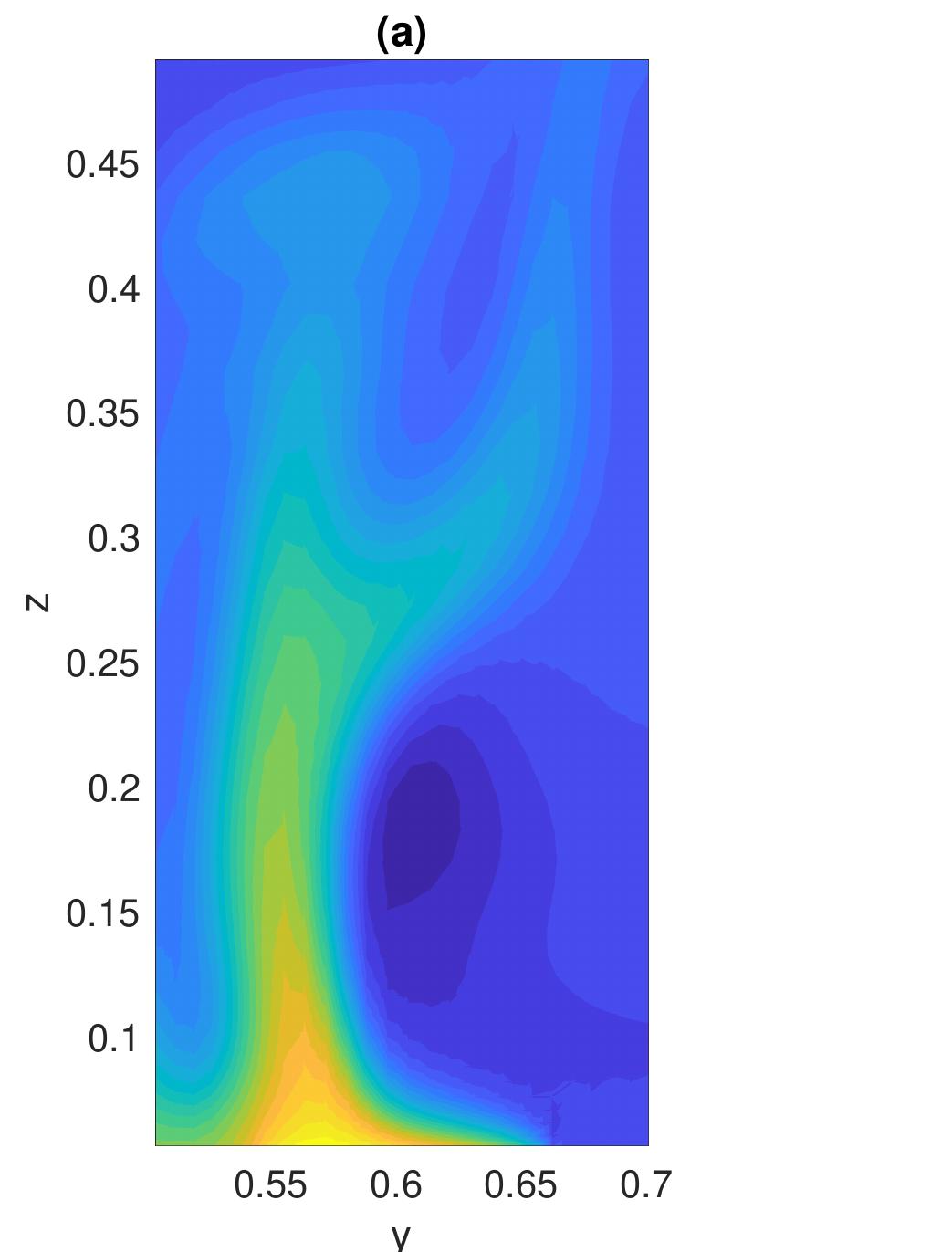}\includegraphics[trim={5.5cm 7cm 2.5cm 7cm},clip,width=3.86cm]{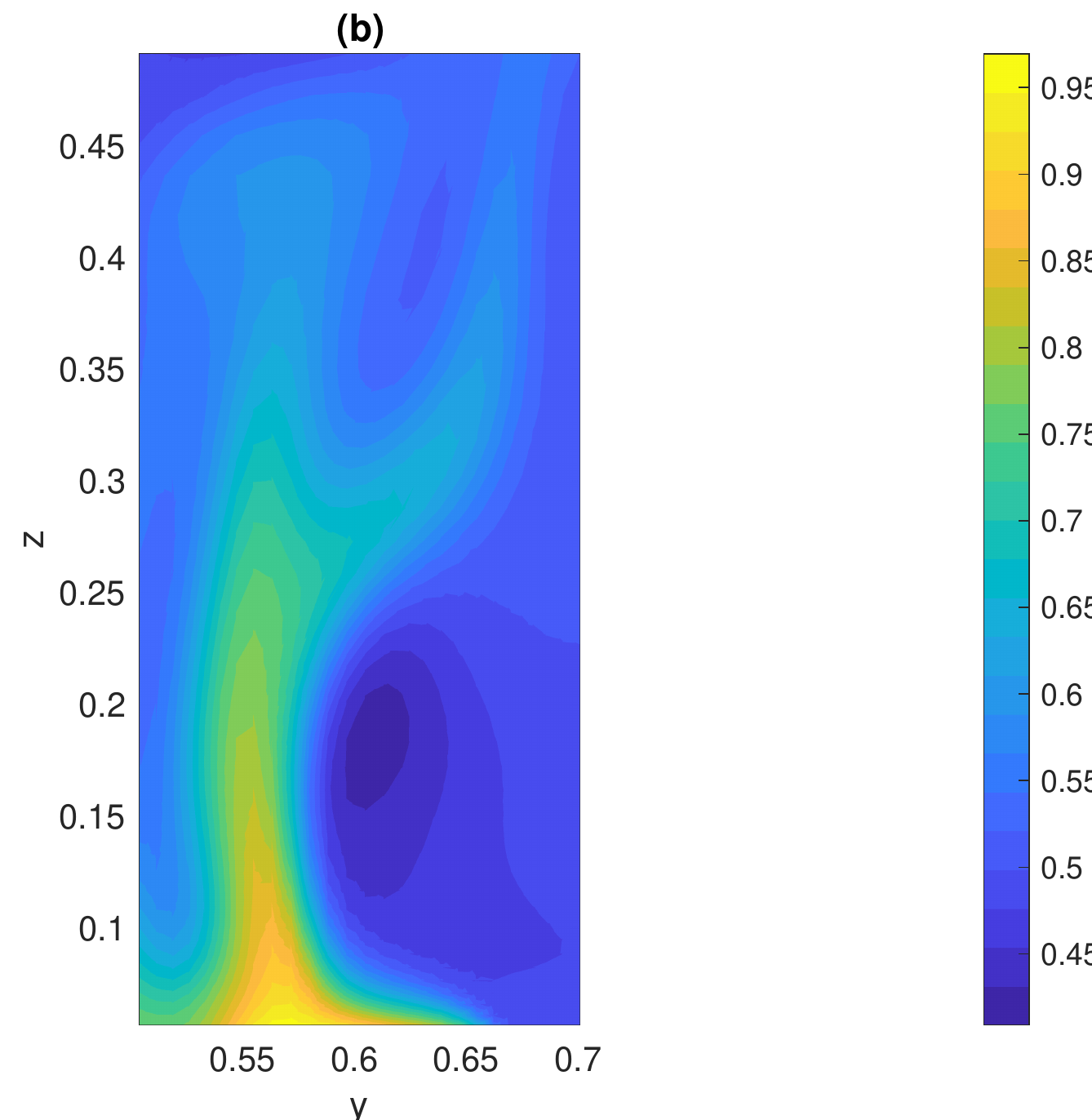}\includegraphics[trim={0cm 7cm 4cm 0cm},clip,width=4.6cm]{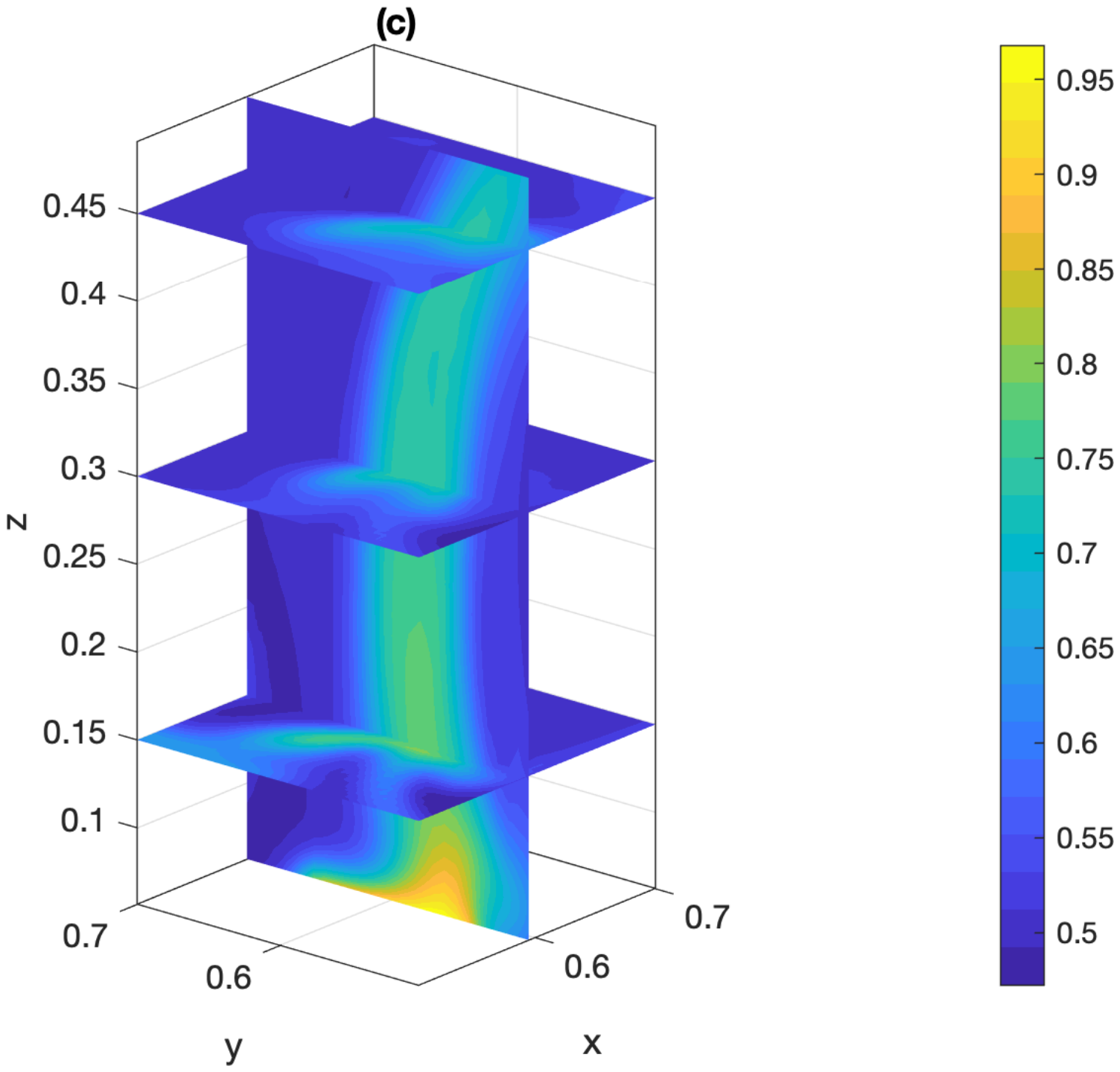}\includegraphics[trim={4cm 7cm 0cm 0cm},clip,width=4.6cm]{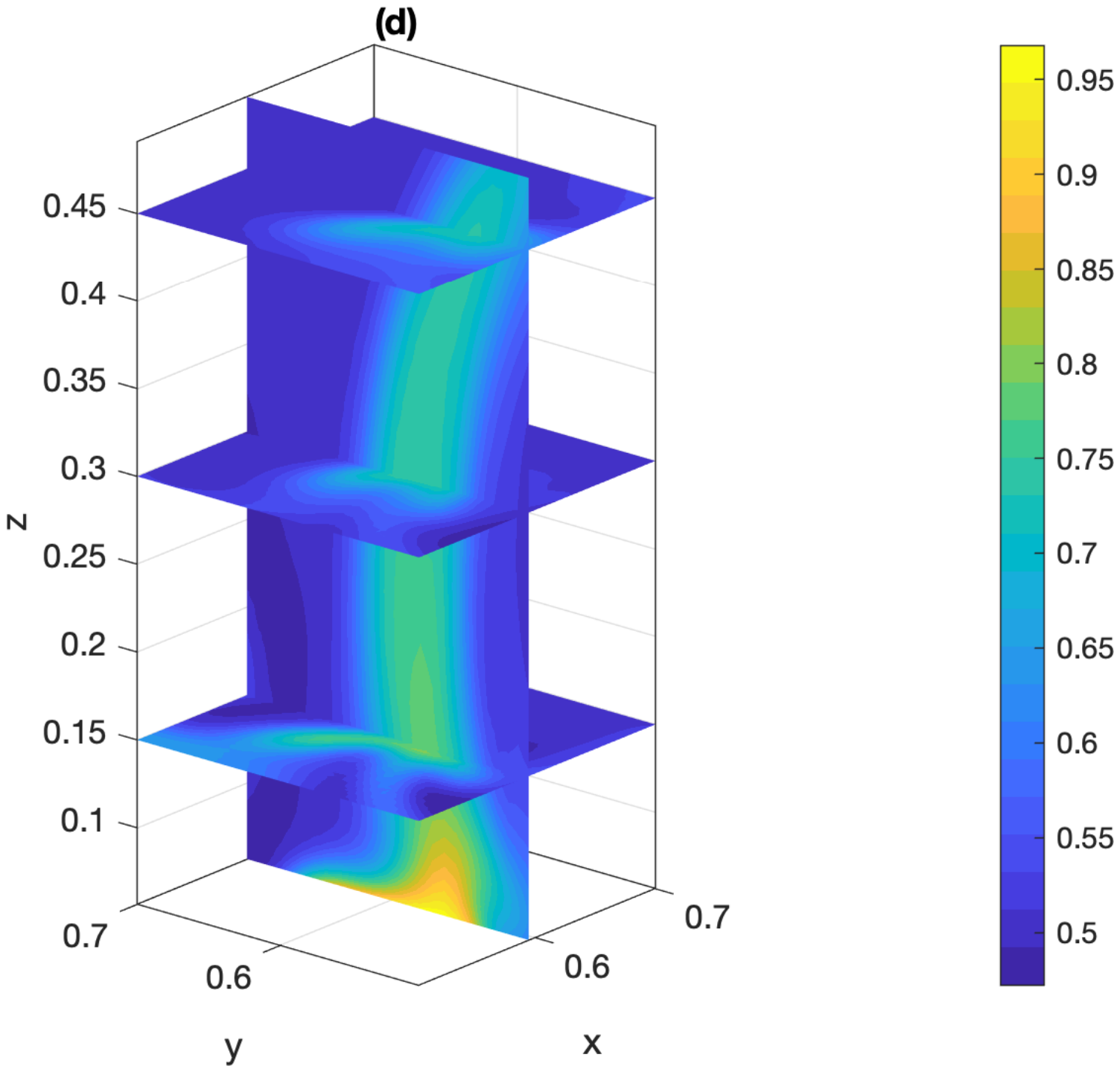}\\
\includegraphics[trim={5.5cm 7cm 5.5cm 7cm},clip,width=3cm]{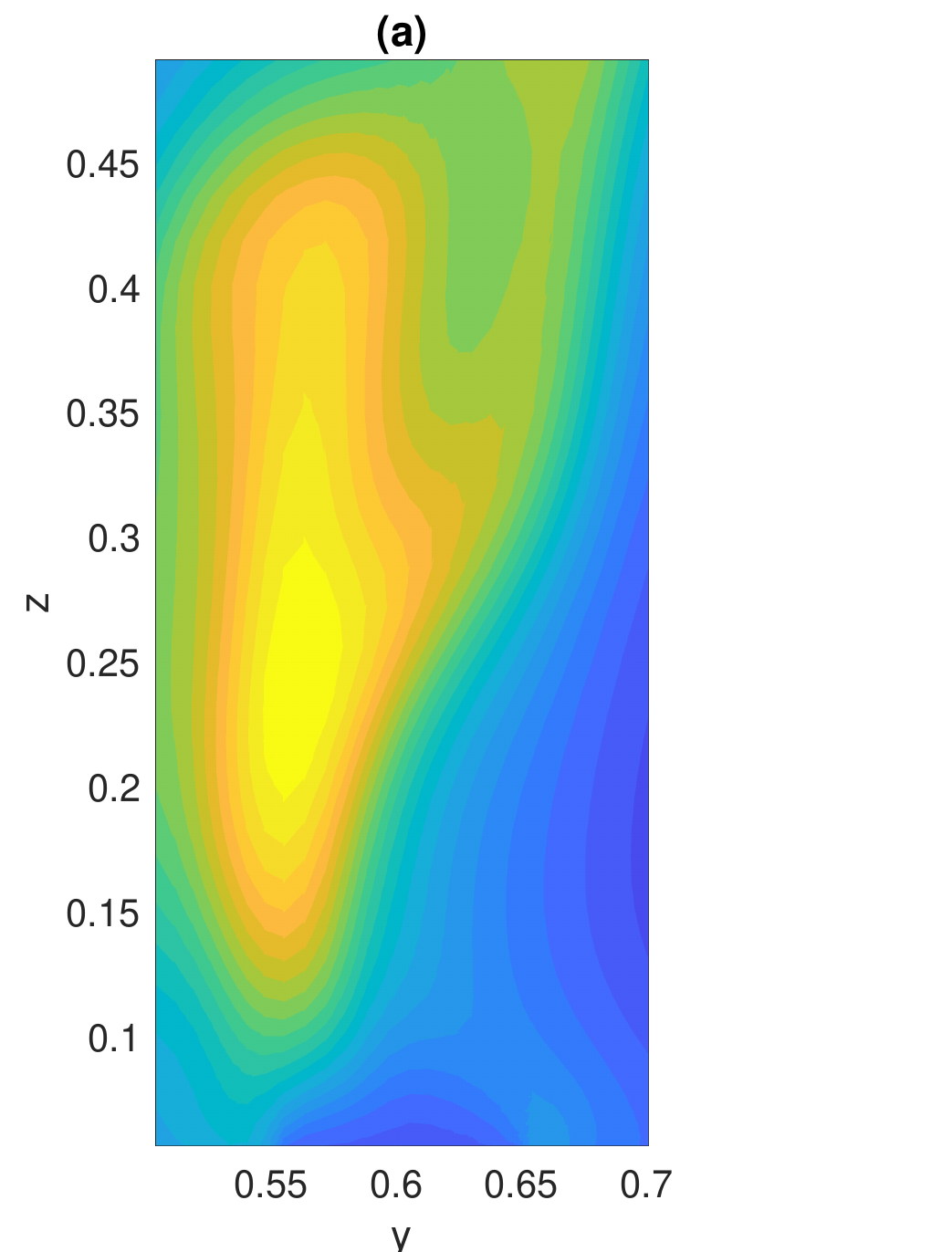}\includegraphics[trim={5.5cm 7cm 2.5cm 7cm},clip,width=3.86cm]{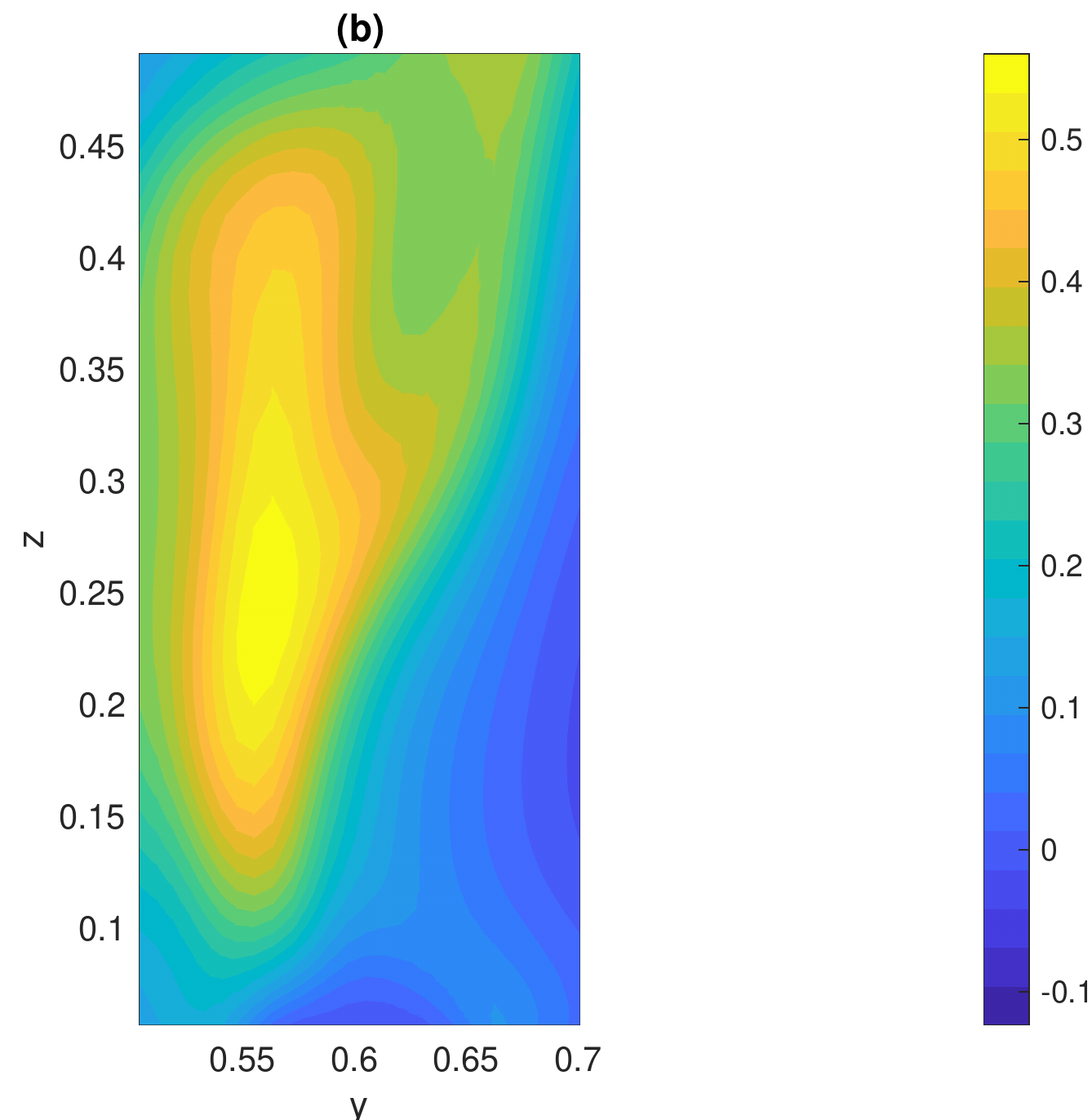}\includegraphics[trim={0cm 7cm 4cm 7cm},clip,width=4.6cm]{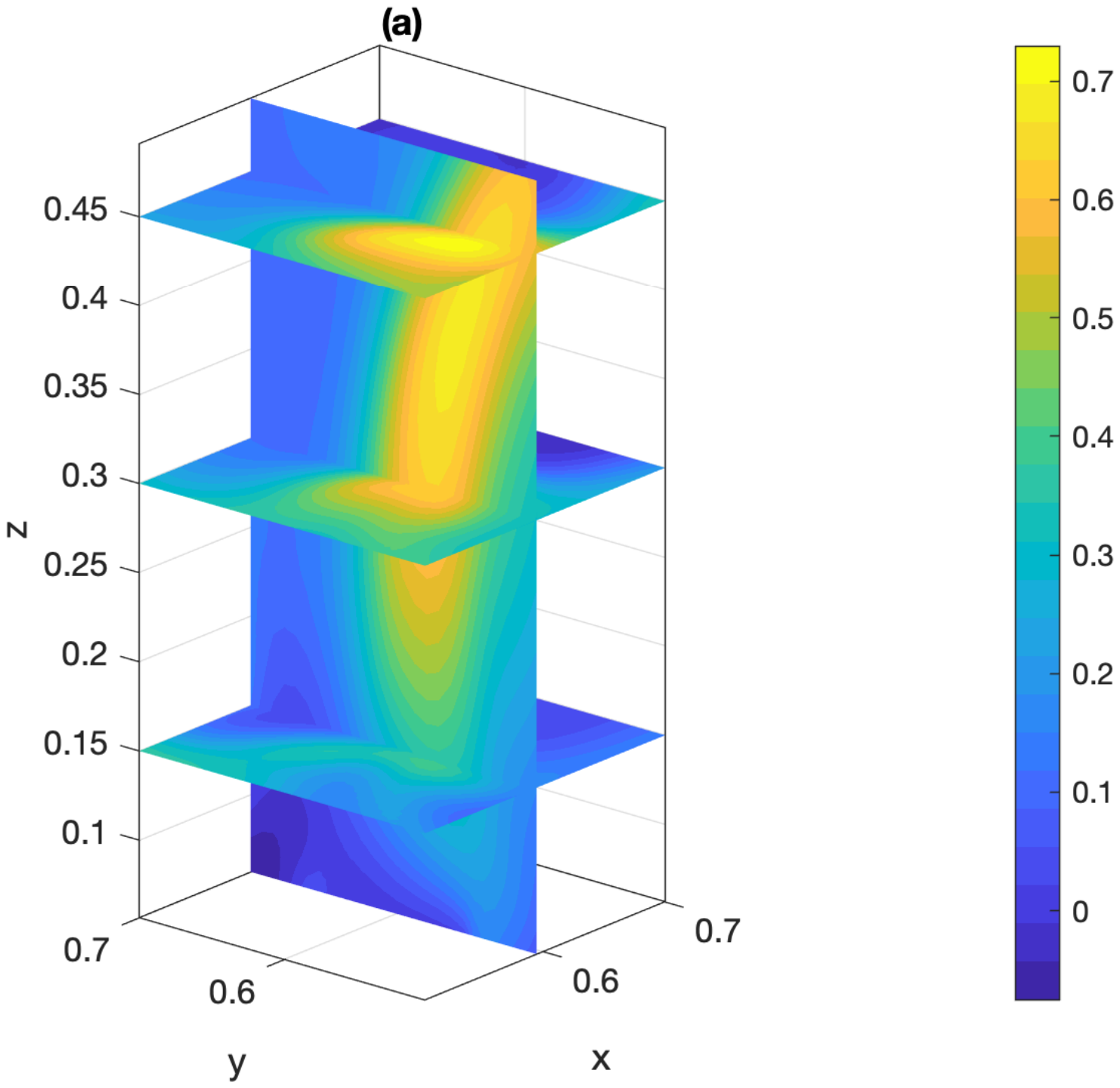}\includegraphics[trim={4cm 7cm 0cm 7cm},clip,width=4.6cm]{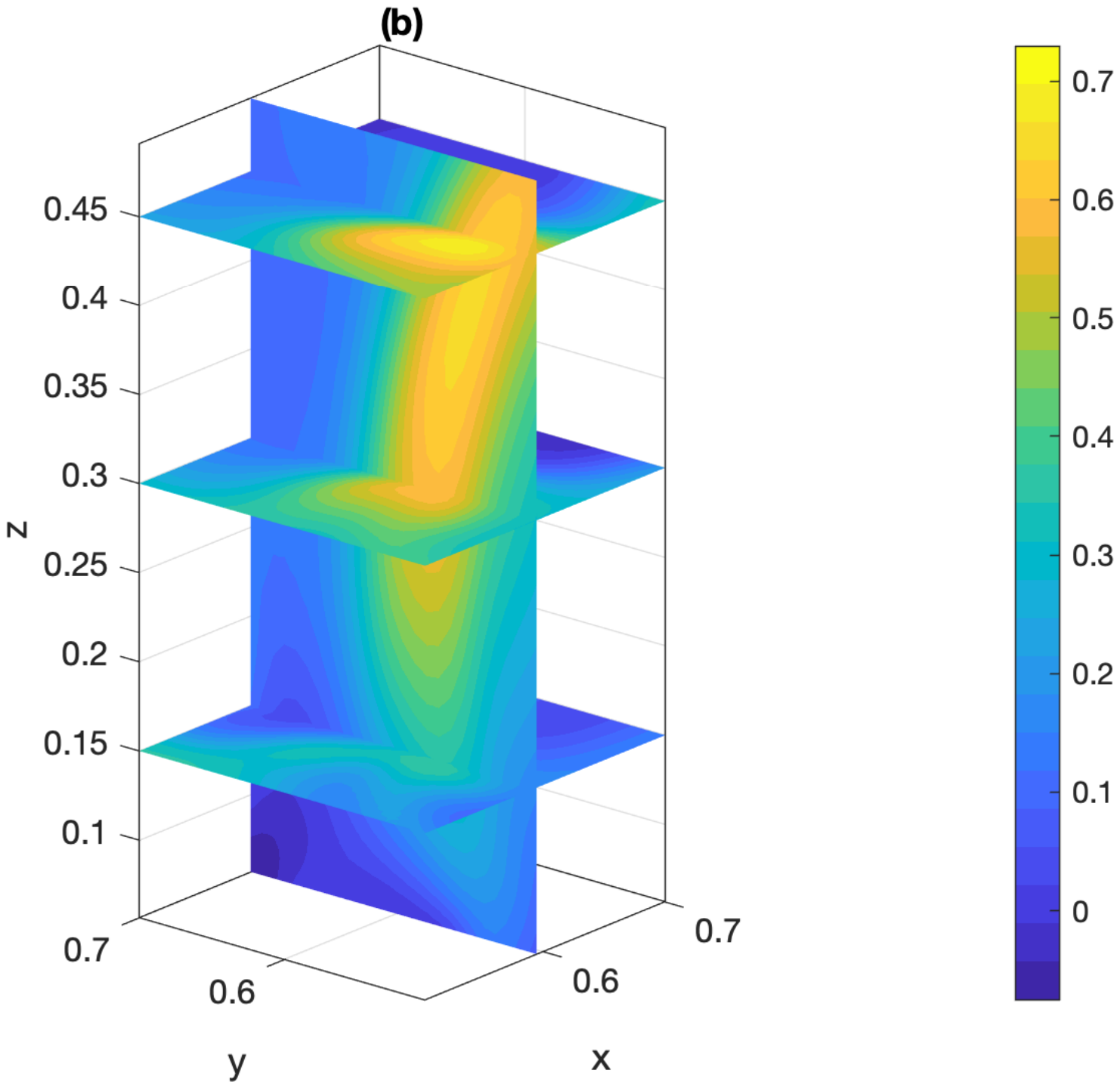}\\
\includegraphics[trim={5.5cm 7cm 5.5cm 7cm},clip,width=3cm]{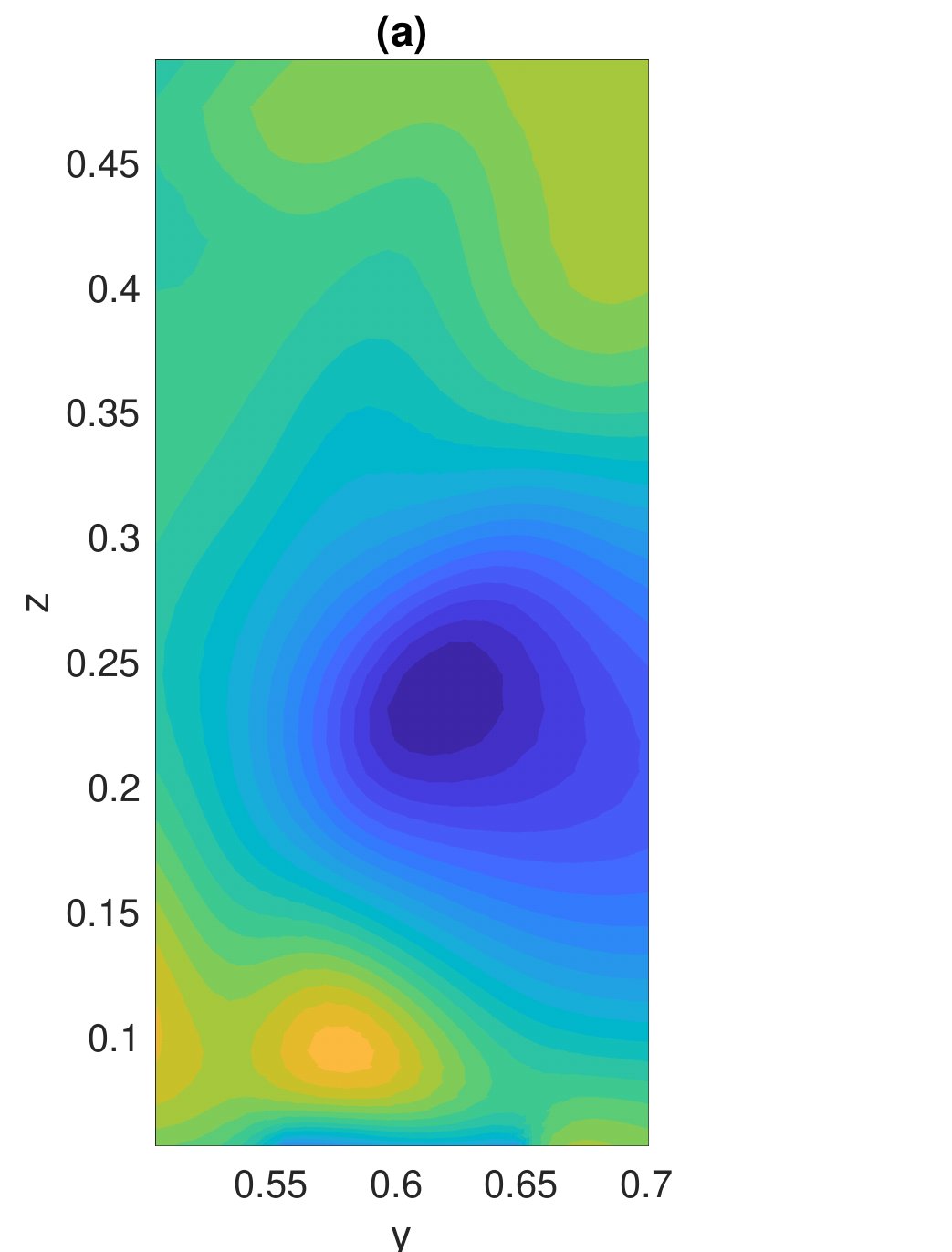}\includegraphics[trim={5.5cm 7cm 2.5cm 7cm},clip,width=3.86cm]{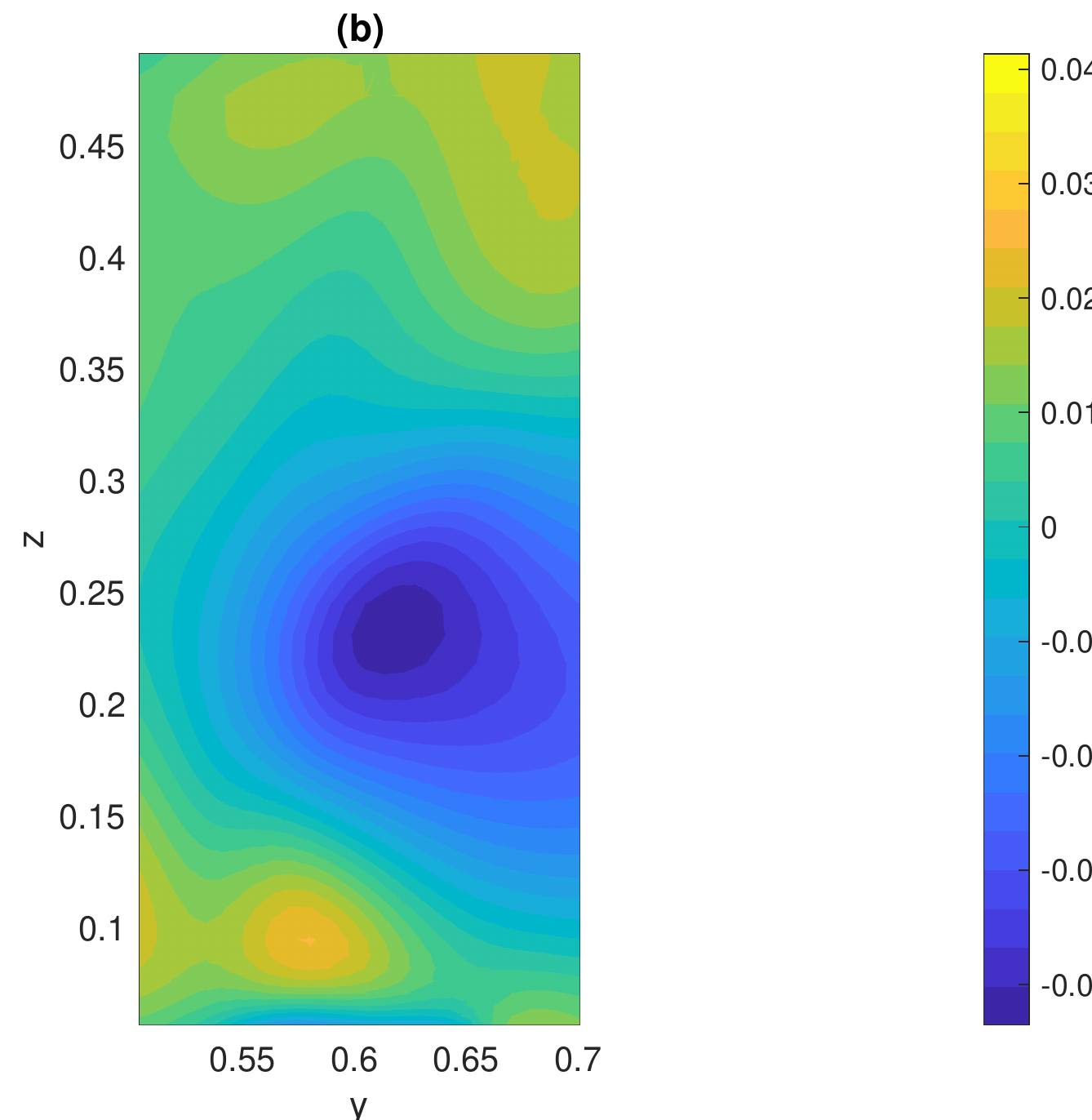}\includegraphics[trim={0cm 7cm 4cm 7cm},clip,width=4.6cm]{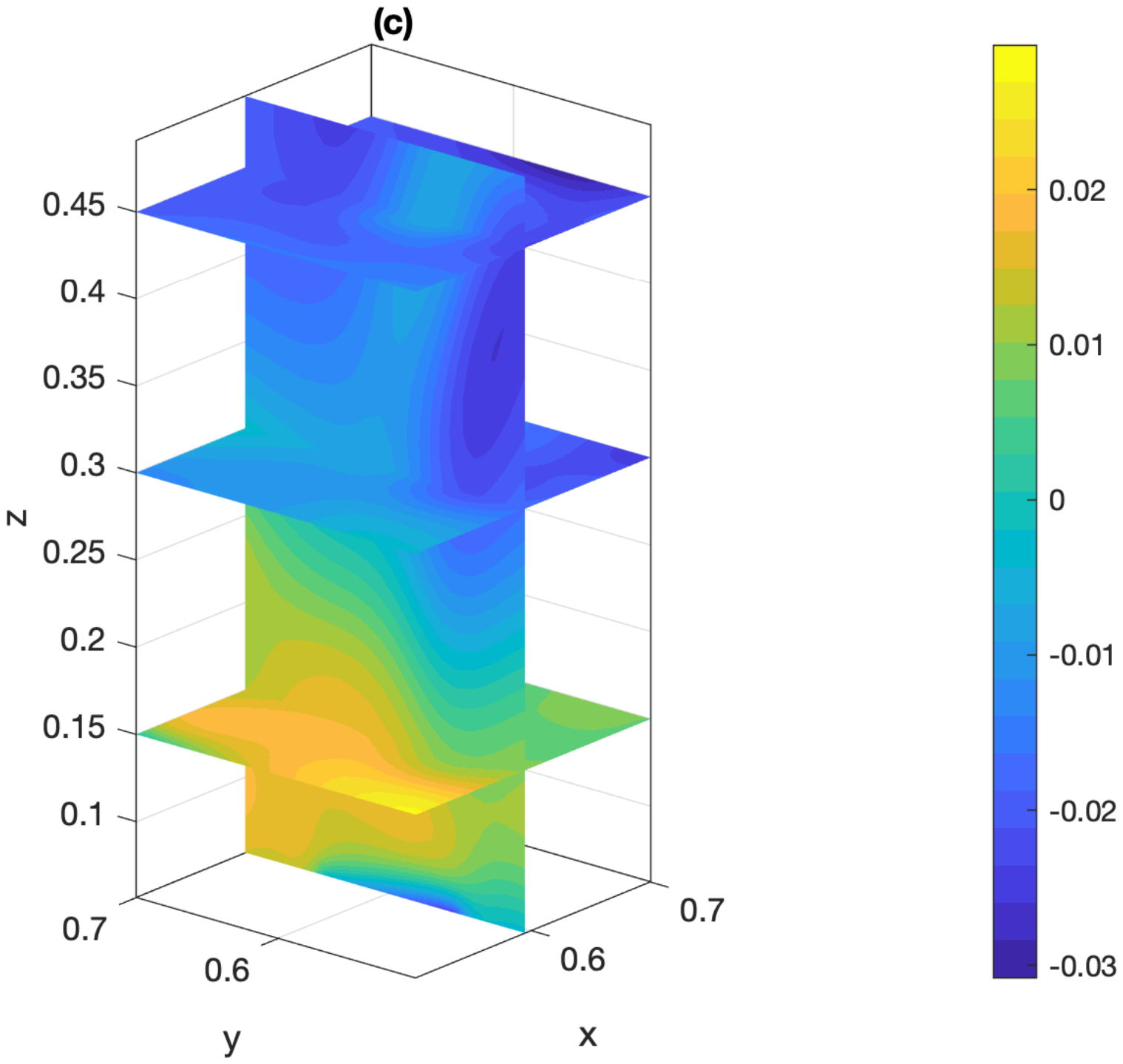}\includegraphics[trim={4cm 7cm 0cm 7cm},clip,width=4.6cm]{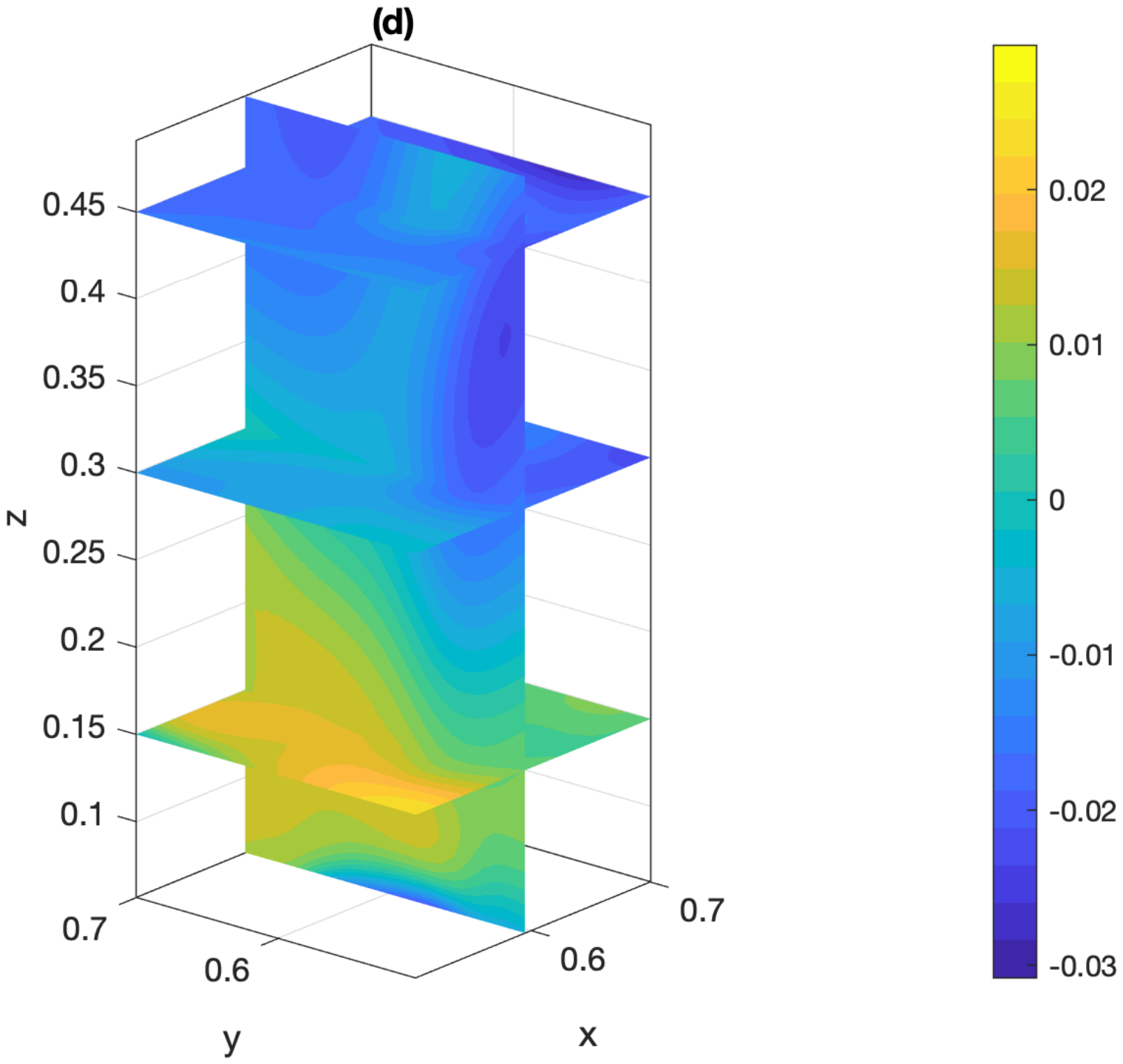}
\caption{Comparison of some ground truth (DNS) (a,c) and 2Ref PINN-predicted (b,d) instantaneous  fields (a,b: vertical slices; c,d: vertical and horizontal slices) at two different time instants. Those time instants are chosen so as to correspond to DNS snapshots that are {\em not included} in the training database. Top row: temperature (a,b): $T(x_0=0.58,\cdot,\cdot,t=74.7)$ and (c,d): $T(x_0=0.595,\cdot,z_0=[0.15,0.3,0.4],t=70.7)$, middle row: similar representation for the vertical flow velocity $\mathrm{v}_z$ and bottom row: similar representation for the pressure field $p$. Number of isocontour levels have been kept low in order to visually emphasize the subtle differences.}
\label{fig:slabs}
\end{figure}
Fig. \ref{fig:slabs} shows some spatial comparisons  of some instantaneous flow quantities in the training domain at two different instants, representative of some different flow topologies. Two different types of view are proposed: - for the first time instant (2 left columns) the reader faces a vertical map corresponding to a vertical $(y,z)$-slice taken through the data, while for the second time instant (2 right columns), this slice is seen from the side and three additional horizontal-$(x,y)$ slices are proposed to give an idea of the three-dimensionality of the flow. It is important to emphasize that the training database used for this model (cf. Tables (\ref{tab:testcases}-\ref{tab:training})) uses only half (i.e. one every two) of the available DNS snapshots. It is therefore possible to generate an {accurate} prediction from the PINN surrogate for some specific time instants at which neither DNS data nor PDE residual points where used for the training, {as shown on the figure}. 
{The two instants show two types of flow organization in the figure: a simple plume or a fork-like plume. In both cases, we see a plume foot carrying heat from the local roughness toward the top of the domain, with a local vertical acceleration a little above the heat source. But whereas the single plume undergoes a loss of pressure in its ascent, the pressure distribution is more complex for the fork-like plume: the pressure loss is followed by a local increase along the two  new emerging  conduits of plume.}
Despite strongly evolving three-dimensional spatial structures, PINN predictions accuracy is remarkable and only subtle differences are noticed. 

\subsection{Improved PINNs capability with temporal- or spatial-penalty padding}

\begin{figure}[!h]
\centering
\includegraphics[trim={10cm 3cm 7cm 7cm},clip,width=10cm]{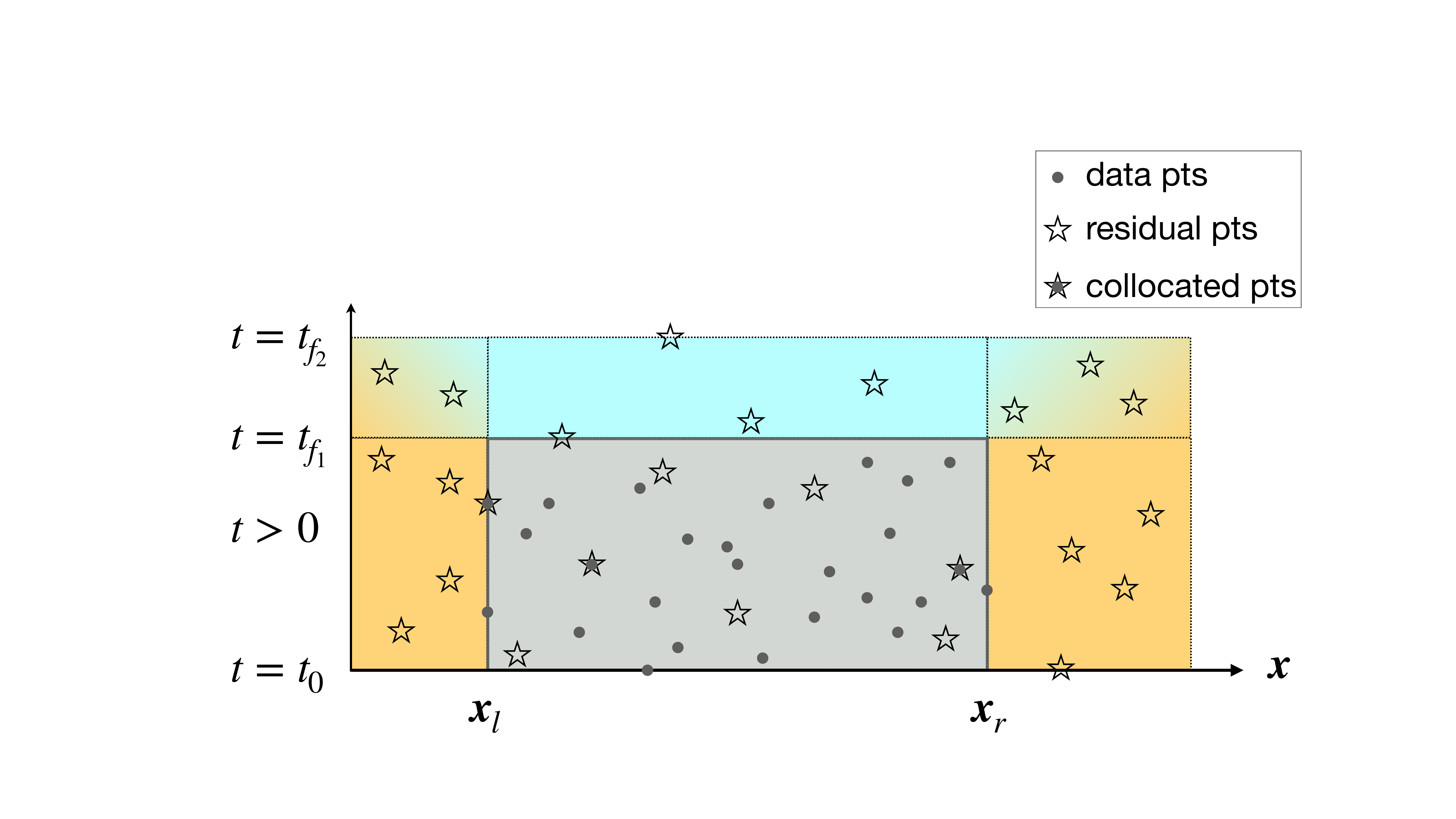}
\caption{Different spatial/temporal sampling strategies. Only one dimension is represented in space for clarity. Points at which data are available (iron color circles) and points at which PDEs residual are minimized (black hollow stars) are not necessarily collocated due to the random selection of the minibatch approach. Residual points may be distributed in more distant/later space/time regions where data points are lacking in order to regularize the approximation in close neighborhoods of the training domain (gray region) populated with data points.  }
\label{fig:sampling}
\end{figure}

Here the idea is to check if it is possible, given a computational budget, to increase the predictive accuracy of the PINN by adopting a different sampling strategy for the choice of the domain points at which the residual penalties are imposed. Let us assume that we dispose of a training data set of size $(N_L,N_R=N_L)$, so we can afford to reduce the residuals of the PDEs over a number of points equal to the number of data labels. Moreover, let us assume that the labeled data points are localized in a certain temporal/spatial domain of interest, cf. black dots in grey area in Figure (\ref{fig:sampling}) where space is reduced to a single dimension to simplify the sketch. A standard approach is to allocate all of the residual points within the same domain. Another possibility is to use part of the residual samples to pad, either in space or in time, the surrounding regions of this domain in order to check if the accuracy is improved within the domain of interest. Indeed, it is known that predictive accuracy of the PINNS is lower close to spatial and temporal boundaries, so we hope to improve the accuracy closer to the boundaries by extending the domain of regularization. \\
Table (\ref{tab:training2}) shows the details of some numerical experiments which have been setup in order to investigate this idea. Three new cases, 2C4, 2C5 and 2C6, adopt the same database and data sampling as case 1C7, i.e. database 1Db2, but this time the residual points span a longer time period for case 2C4, a wider spatial vertical (horizontal) range for case 2C5 (2C6), respectively. Numerical testing over the 1Db2 database shows that the accuracy is much improved for those cases compared to case 1C7 for which all residual points are located within the domain of interest. In particular, improvement of case 2C6 for which the {spatial domain has been horizontally padded} is spectacular (i.e. spatial density of residual points is halved compared to case 1C7), cf. subplots (e-f) from the Figure (\ref{fig:scatters}). Another numerical experiment has been ran (i.e. 1C7$^{*}$), in order to check that the improvement was not simply due to the lower density of residual point within the training domain, instead of the padding effect. This was not the case as 1C7$^{*}$ produced very poor results, cf. fourth line of Table (\ref{tab:training2}).



%

\begin{table}[!ht]\footnotesize
	\centering
	\begin{adjustbox}{center}
	\begin{tabular}{l c c c c c c c c c c }
    \toprule
\multicolumn{1}{c}{Model} & \multicolumn{2}{c}{Training}&
\multicolumn{2}{c}{Testing} & \multicolumn{5}{c}{Accuracy}\\
\midrule
\multicolumn{1}{c}{} & \multicolumn{1}{c}{size $(N_L,N_R)$} &\multicolumn{1}{c}{database} &  \multicolumn{1}{c}{size $(N_T)$}& \multicolumn{1}{c}{database} &\multicolumn{1}{c}{aRMSE} & \multicolumn{1}{c}{aMAE}& \multicolumn{1}{c}{$\mu$ error } &\multicolumn{1}{c}{$\sigma$ error}& \multicolumn{1}{c}{$aR_{\mathrm{corr}}$ } & \multicolumn{1}{c}{$aR^2$}\\
{1C7} & {$(1\mathrm{e}6,1\mathrm{e}6)$} &
{1Db2} &
{$2.844\mathrm{e}5$}& {1Db2} & 4.952e-03 & 2.841e-03 & 0.5 & 2.5 & 9.968e-01 & 9.928e-01 \\
{1C7$^{*}$} & {$(1\mathrm{e}6,5\mathrm{e}5)$} &
{1Db2} &
{$2.844\mathrm{e}5$}& {1Db2} & 9.858e-03 & 6.034e-03 & 0.9 & 7.6 & 9.783e-01 & 9.526e-01\\

{2C4} & {$(1\mathrm{e}6,1\mathrm{e}6)$} &
{(1Db2,\,$\text{grid}_{\text{2Db2}}$)} &
{$2.844\mathrm{e}5$} & {1Db2} & 2.626e-03 & 1.736e-03 & 0.4 & 1.3 & 9.992e-01 & 9.982e-01 \\

{2C5} & {$(1\mathrm{e}6,1\mathrm{e}6)$} &
{(1Db2,\,$\text{grid}_{\text{1Db2}}^{*}$)} &
{$2.844\mathrm{e}5$} & {1Db2} & 3.762e-03 & 2.354e-03 & 0.4 & 1.8 & 9.981e-01 & 9.958e-01 \\

{2C6} & {$(1\mathrm{e}6,1\mathrm{e}6)$} &
{(1Db2,\,$\text{grid}_{\text{1Db2}}^{\ddag}$)} &
{$2.844\mathrm{e}5$} & {1Db2} & 2.927e-03 & 1.982e-03 & 0.4 & 1.2 & 9.991e-01 & 9.980e-01 \\
\bottomrule
    \end{tabular}
    \end{adjustbox}
    \caption{The caption is similar to the one of Tables (\ref{tab:training},\ref{tab:accuracy}), but this time pay attention to the fact that label and residual data points can be sampled from different databases, e.g. {(1Db2,\,$\text{grid}_{\text{2Db2}}$)} means that the labeled data are chosen from the 1Db2 database while the PDEs residuals are evaluated at space/time grid points from the 2Db2 database. $\text{grid}_{\text{1Db2}}^{*}$ refers to the 1Db2 grid which has been vertically extended to cover the domain $\Omega_{\text{PINN}}^*=[0.5,0.7]\times [0.5,0.7] \times [0.05,0.9]$ with resolution $(26\times 26 \times 60)$ and $\text{grid}_{\text{1Db2}}^{\ddag}$ refers to the 1Db2 grid which has been horizontally extended to cover the domain $\Omega_{\text{PINN}}^{\ddag}=[0.5,0.78]\times [0.5,0.78] \times [0.05,0.5]$ with resolution $(37\times 37 \times 38)$.
    }
    \label{tab:training2}
\end{table}

\begin{figure}[!h]
\centering
\includegraphics[width=10cm]{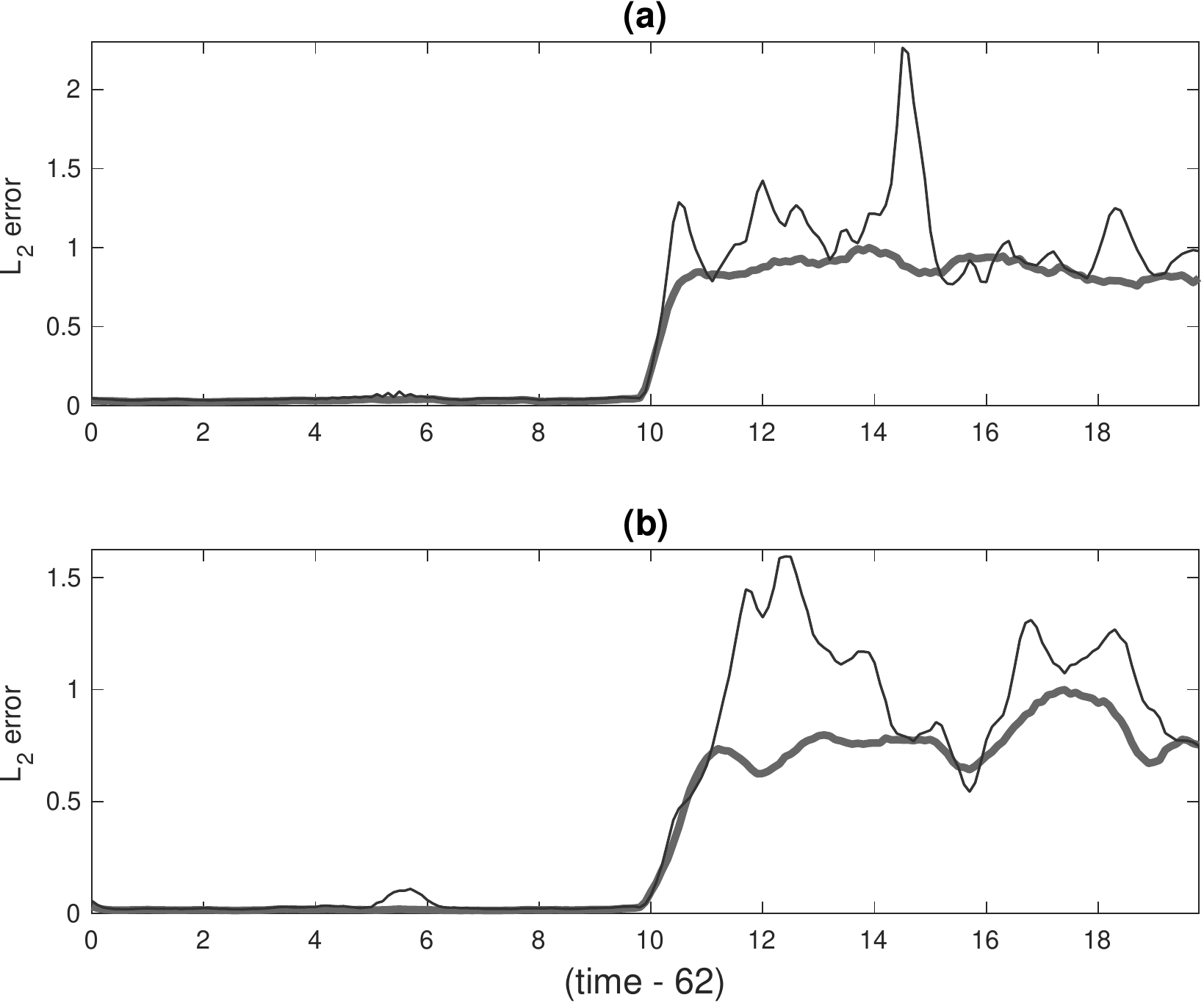}
\caption{Comparison of the temporal distribution of $L_2$ spatial errors for case 1C7 (thin line) and 2C4 (thick line) for temperature (a) and $\mathrm{v}_z$ vertical flow velocity (b). The curves in each subplot are normalized by the maximum value of the error for case 2C4 (i.e. 20\% for temperature and 83\% for velocity). The padded approach 2C4 controls errors much better, especially in the second half of the time window.}
\label{fig:prolongation}
\end{figure}

{But the other padded cases are also interesting. For instance, }Figure (\ref{fig:prolongation}) compares the temporal distribution of the $L_2$ spatial errors integrated for case 1C7 (thin black line) and 2C4 (thick gray line) for temperature (a) and $\mathrm{v}_z$ vertical flow velocity (b). The curves in each subplot are normalized by the maximum value of the error for case 2C4 (i.e. 20\% for temperature and 83\% for velocity. The PINN models are asked to predict the solution on the long time range and for a resolution that is doubled (i.e. 200 snapshots for $\Delta T_l$) compared to the one of their training data.  
We clearly identify the first half of the time domain in which errors are low. The 1C7 PINN solution is then predicted in the later time domain (no information from this time range was used during training of this model), while the 2C4 prediction benefits from  a training with residual points (but no data) in this time range, as explained previously. Not only, the 2C4 solution is improved in the first half but it is also much more robust in the second half, controlling better error spikes all along the time window.\\
In the following section, we will discuss the results in light of the PINNs salient numerical points. Then we will address the issue of modeling  higher turbulence levels by proposing a new simple idea that considerably improves the prediction results.

\section{Discussion, preliminary results and perspectives}
\label{sec:discussion}

\subsection{Accuracy vs. sampling}
An important matter to this study was the one of the PINNs robustness to lower resolution in the observation data.
While it was shown that we were able to ``replay" DNS simulations with the PINNs models, the PINNs accuracy depends on many factors including the amount of data and regularization used and the choice of the physical quantity under scrutiny. For this particular application, DNS temperature data inside the domain was preferentially provided, justifying the very good predictive accuracy obtained for the temperature and the vertical component of the flow velocity. Indeed, $\mathrm{v}_z$ is strongly correlated to the temperature gradient due to the vertical buoyant forces induced by the {consideration of the gravity}. In plane flow velocities which magnitude is lower were in general a bit less accurate. Finally, the pressure field was the least accurate predicted quantity. We
emphasize that no data was provided for the pressure as boundary or initial conditions, which was a hidden state and was obtained indirectly via the incompressibility constraint without splitting the Navier-Stokes equations. \\
As an example of this fine expressivity, for the training of case 2C3, a half-a-million time-space scattered temperature data points over 50 snapshots and available boundary velocity data points (on ${\partial \Omega^{\dag}}$) were used and provided a very good average accuracy of $aR^2=0.988$ over a testing sample encompassing 100 snapshots. This training corresponds to a moderate sampling of the DNS data: i.e. temperature  data were randomly collected with a temporal sampling of $\Delta t =0.2$, that is every 80 DNS time steps and a spatial sampling of about 20\% of the $(26\times 26 \times 38)$ available DNS temperature data points at each time step.  \\
It remains that figuring {\em a priori} the amount of information that is necessary for the physics-informed network training to converge well and fast, is a complex matter because of the interplay of many different terms involved in the composite loss function. Indeed the lost function, that is just an evolving scalar, encapsulates various error terms related to the data (in some chosen norm) and to a soft penalty represented by some appropriate functional. This functional is designed to constraint the outputs of the neural network to satisfy a set of specified conditions.  This apparently simple formulation hides its underlying complexity because $\mathcal{L}_{\text{Label}}$ gathers several data-fit terms, {possibly} including initial, boundary and internal data, while $\mathcal{L}_{\text{PDE}}$ also corresponds to various contributions coming for instance from conservation of mass and momentum. \\
{It was shown that} this approach affects the loss gradient and might lead to an unstable imbalance in the magnitude of the back-propagated gradients during model training using gradient descent, as explained in \cite{wang2020understanding}.
Some formulations have proposed a regularization parameter that acts as a weight in front of the penalty term. Indeed we have noticed that the magnitude of the penalty contribution to the total loss varies relatively to the data error term, cf. Figure (\ref{fig:loss}).\\
{It is straightforward} to monitor the distribution of the back-propagated gradients of the loss with respect to the neuraal network parameters during training.
A finer analysis would consist in monitoring the distribution of the back-propagated gradients of each individual loss terms with respect to the weights in each hidden layer. This  tedious work may reveal that some gradients are too small to condition the optimization search. This is often the case for the gradients corresponding to the boundary (or initial conditions) loss terms \cite{wang2020understanding}. {This lack of this information affects the training} and restrains networks accuracy as it is known that a PDE system may have infinitely many solutions in this case. 
Despite {these} proposed adaptive dynamic weights strategy, it seems that turbulent flows {nonetheless} require the use of {additional} user-tuned scaling parameter that affects the respective weight of the adaptation, {cf. section 4 in \cite{jin2020nsfnets}}. This shows that the perfect balance of the residuals gradients is not easily reachable and asks the question of the proper normalization with careful tuning of {those} anisotropic weights.\\
The loss function is typically minimized using SGD algorithm and a large number of training points can be randomized within each SGD iteration. Therefore, it is also the relative amount of points density which are sampled from those data and PDE residuals penalization sources, that are weighting {either explicitly or implicitly (depending on the formulation)} the {importance of the} different terms. In our case, we have decided to first keep our approach simple and to fine tune the error balance by adjusting sampling density of data and residual points. Interestingly, we have shown that it is more beneficial to distribute less densely the amount of points at which the PDEs residuals are minimized. More specifically, we have found that the accuracy is improved if those points are also scattered across spatial or temporal domains encompassing the domain within which the labeled data points are available.\\

\subsection{Importance of boundary {information: data vs. penalty padding}}
{The previous discussion based on our results confronted to a literature review, clearly points to the importance of the boundary data information in the PINN formulation.}
Unlike two-dimensional laminar flow problems, it was noticed in \cite{raissi2020hidden,jin2020nsfnets} that for more complex convective three-dimensional flows, temperature data was not sufficient (problem of well-posedness) to satisfactorily train a PINN model, and information relative to flow velocity boundary conditions were also necessary. This is indeed something that we have confirmed in previous studies and also the importance of the positioning of the training domain relative to the flow features \cite{AgrawalFrontUQ19}. For numerical experiments in our paper, DNS fluid velocity from the domain boundaries (except the top one) is used to complement the temperature data. 
The dimensionality of this information being lower, for instance for case 1Ref: $ |  \mathbf{v}_{\partial \Omega^{\dag}}^{\mathrm{DNS}} | =4628$ at each given time, the small mini-batch size that we use at each training iteration ($\MB=2000$) collects (on average due to the random sampling) about $(\MB/100_{\text{{snapshots}}})/5_{\text{{faces}}}=4$ flow velocity data points per face which is a small number. Nevertheless, for each epoch based on the temperature data, the algorithm cycles more than once (here about 4 times) through the fluid velocity boundary values, therefore using this information many times.\\
It will be interesting to further quantify the impact of the boundaries information on the method efficiency. This could be achieved by playing with the ratio of training data points chosen at the boundaries vs. the inside of the domain.\\
{The penalty padding that we have proposed in the previous section can also be handy in this case. It comes to play as a regularization over  the spatial and/or temporal zones surrounding the training boundaries which are often regions of poor accuracy of the PINN surrogate. It seems to complement  the local  boundary data and blends the solution nicely across the chosen boundaries, resulting in a noticeable global accuracy improvement.  Future works need to be pursued in order to determine how to improve this technique, e.g. choice of the padding domains extent and shapes, distribution of the residual points density, choice of PDEs to enforce, etc.\\
An interesting approach would be to see if the padding regularization may substitute (at least in part) the amount of boundary data. That is we wish to reduce the training data at the boundaries relative to the padding penalty.}
To this end, a test was carried out in order to infer on the importance of missing boundaries information: the best padded case (2C6) was rerun without including velocity data on certain domain faces covered by the padding. To make things clear, only velocity data at the bottom, back and left faces   were provided during training, while velocity at front (respectively right) face located at $x=0.7H$ (respectively  $y=0.7H$) were not used, cf. Figure (\ref{fig:domain}-(a))). The idea was to check if missing local boundary information could be supplemented thanks to the padded neighbor region filled with low PDE residuals enforcement points. The results (not reported here) were deceptive with an averaged error close to 20\%. This shows that the boundary information is very important for this type of flows, especially when the amount of bulk labeled data is on the lower side. This finding is consistent with recent works applied to incompressible internal flows where a structured DNN architecture was devised in order to automatically enforce (in a ``hard" way) initial/boundary conditions. In this particular case, it was not necessary to include any bulk simulation data, the DNN being trained by solely minimizing the residuals of the Navier-Stokes equations.

\subsection{Improving modeling capability for more turbulent scenario by relaxing surrogate constraints}

The promising results obtained in the previous sections motivate an investigation of more challenging natural convection metamodeling at higher turbulence levels. The attempt might fail as it is known that conventional PINN models are not very successful at approximating 
 complex dynamics leading to solutions with non-trivial behavior, such as directional anisotropy, multi-scale features or very sharp gradients. 
 The specificity of the PINN approach is that the constraints alter the loss landscape of that type of deep neural networks. As seen previously, different terms in the PINN composite loss function  have different nature and magnitudes, sometimes leading to imbalanced gradients during back-propagation.
 Recent works have pointed to the problem of the stiffness of the PINN gradient flow dynamics. They have proposed a learning rate annealing algorithm that utilizes gradient statistics during training to adaptively weight the different terms in the {\em label part} of the loss function. They have also proposed a new fully-connected neural network architecture that is less sensitive to the stiffness of gradient flow.\\
Our approach is to propose a minimal modification of the PINN computational framework in order to make it more efficient for our type of application.  In particular, we do not want to modify the PINN architecture, the  training computational budget or to upgrade too drastically the training dataset size. As we will describe in more details below, our idea is to relax some of the PDE residual losses in order to enhance the accuracy and robustness of our PINNs.

\begin{figure}[!h]
\centering
\includegraphics[trim={1.0cm 0.1cm 0.5cm 0.0cm},clip,height=70mm]{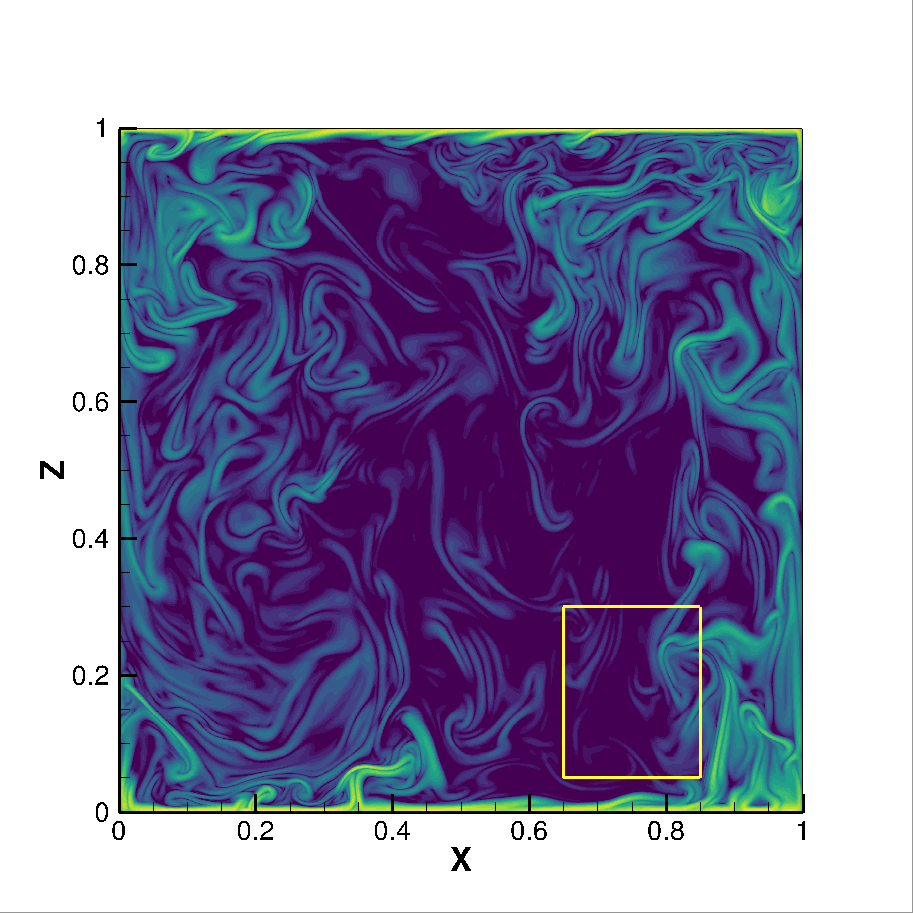}(a)\includegraphics[trim={1.0cm 0.1cm 0.5cm 0.0cm},clip,height=68mm]{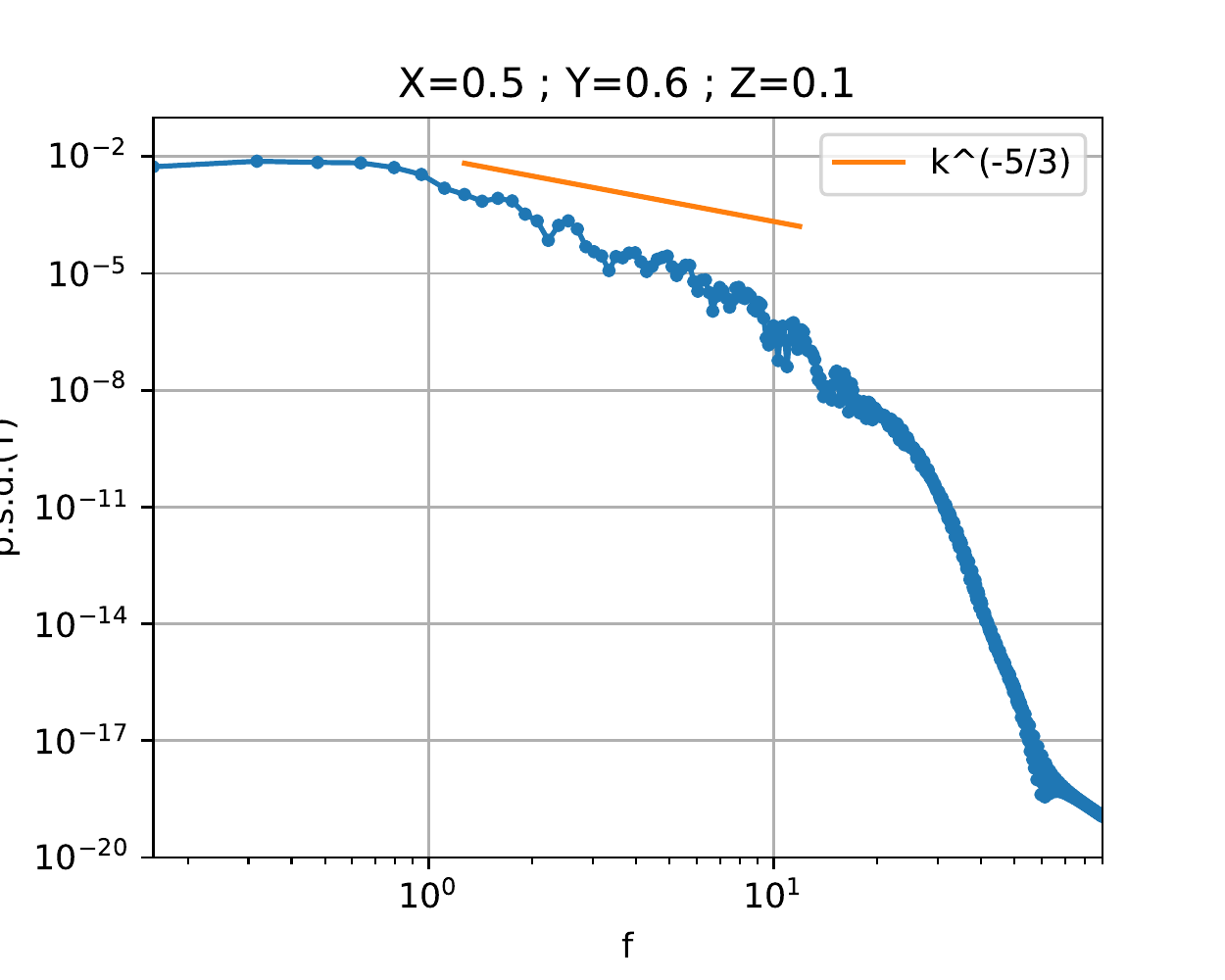}(b)
\caption{Example of instantaneous {heat dissipation} isocontours (2D $(x,z)$ slice at $(y=0.25H,t=290.83)$) from DNS of the RB cavity flow at $Ra=2 \cdot 10^9$ (a). The domain $\Omega_{\text{PINN}}=[0.65H,0.85H]\times [0.2H,0.3H] \times [0.05H,0.3H]  $ over which the PINN model is trained, is depicted as a transparent box. {Temporal power spectrum of temperature at location $(x=0.65H,y=0.25H,z=0.1H)$} (b)}
\label{fig:domain2}
\end{figure}

\begin{figure}[!h]
\centering
\includegraphics[width=11cm]{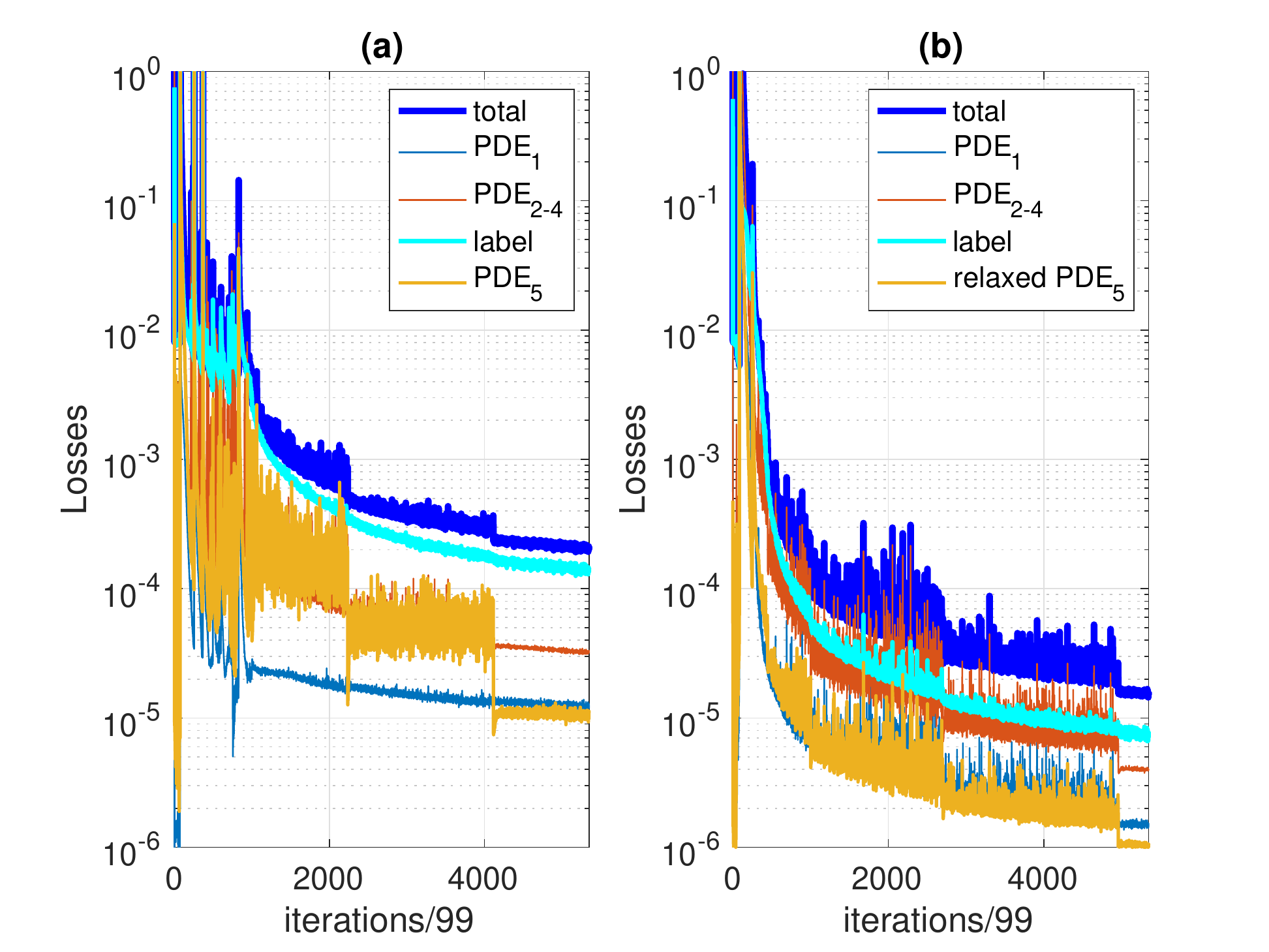}
\caption{$Ra=2 \cdot 10^9$ study: details of the convergence of loss functions for   standard PINN (a) and relaxed PINN$^{\text{r}}$ (b) for which the mass conservation PDE constraint (i.e. PDE$_5$) is relaxed during training. PDE$_1$: loss term associated with the temperature equation and PDE$_{2-4}$: sum of loss terms associated with the momentum equations, cf. system (\ref{eq:PDEs_NS}).}
\label{fig:loss_details}
\end{figure}

In the following, we consider a much more challenging case of RB flow in a smooth cavity filled with water ($Pr=4.4$) at higher $Ra$ number, cf. Fig. \ref{fig:domain2}. The geometry does not bear any roughness anymore and the flow turbulence is now much more developed than previously, with a larger value by two order of magnitude, i.e. $Ra=2 \cdot 10^9$. Our goal is to investigate the level of accuracy we can achieve with a computational budget equivalent to the previous simulations. We keep the same simple fully-connected PINN architecture with $\ell=10$ layers and the same total number of $\text{epochs}=1500$ (which is low compared to other works). A DNS is performed in a computational domain of size $\Omega = [0,H]\times[0,H/2]\times[0,H]$, cf. Fig. \ref{fig:domain2}. The spatial flow scales are obviously harder to capture with more dispersed and less organized small and thin turbulent plumes. The quite large domain $\Omega_{\text{PINN}}=[0.65H,0.85H]\times [0.2H,0.3H] \times [0.05H,0.3H]  $ over which the PINN model is trained, is depicted as a transparent box. Another difficulty relates to the travelling direction and orientation of the plumes relative to $\Omega_{\text{PINN}}$. They do not necessarily travel across the domain from bottom to top because the flow velocity components are less dominated by their vertical component. \\
The amount of data from the ``true" DNS is colossal with a spatial resolution for the {\em PINN domain alone} of $(129\times 65 \times 219)$ points updated every $\Delta t_{\text{DNS}}=4.5\cdot 10^{-4}$. The database being too large to be stored, we have saved the solution with a coarser time resolution, i.e. 249 snapshots collected every $40\times\Delta t_{\text{DNS}}$, totalizing almost half a billion ($4.57242435\mathrm{e}8$) points at which the flow data are saved. It is this latter downgraded version that we will refer as our ``full" DNS. A more tractable database 3Db1, referenced in Table (\ref{tab:testcases}), is constructed from the aforementioned full DNS, with a lower resolution in space and time.
The PINN training dataset retained from the 3Db1 database corresponds to 90\% of the (velocity-temperature) data points of the initial condition, 100\% of the (velocity) boundary conditions and only 25\% of the (temperature) bulk totalizing $7.334712\mathrm{e}6$ data points, which is only 1.6\% of the full (i.e. 0.04\% of the true) DNS spatial/temporal coordinates points. The training dataset being 3.7 times the size of the largest dataset used for the $Ra=2 \cdot 10^7$  studies, the mini-batch size is now increased to $\MB=18522$. \\
The convergence of the various terms of the loss function are depicted in Figures (\ref{fig:loss_details}). In subplot (a), standard PINN results show that the optimization is very sensitive to the incompressibility constraint (PDE$_5$: yellow line), in agreement with the findings of other researchers \cite{jin2020nsfnets}. Oscillations are very strong compared to the other components. Moreover PDE$_5$ convergence exhibits a piecewise behavior with error magnitude and fluctuations getting lower across changes in learning rate cycles, but a convergence that does not really progresses within each of the learning cycles. Despite being quite low, the convergence of the temperature loss (PDE$_1$) looks very flat and seems to stagnate.\\
With the PINN velocity-pressure formulation, the pressure equation is not obtained through  an additional Poisson pressure equation as it is usually done with splitting methods. In fact, the pressure is a hidden state and is obtained via the incompressibility constraint.
Nevertheless, the incompressibility condition is hard to impose through our SGD algorithm. Recent works have investigated other Navier-Stokes (NS) formulations \cite{jin2020nsfnets} including a  streamfunction-pressure formulation for which the incompressibility constraint is exactly satisfied \cite{wang2020understanding}. They concluded that these alternative formulations were more efficient, especially for laminar flows.\\ Keeping our original velocity-pressure formulation, our novel idea is simply to relax the incompressibility constraint. We refer to this relaxed version of the physics-informed neural network as PINN$^{\text{r}}$. Results obtained with this new approach are spectacular, and the total loss reaches a much lower value at the end of the training, cf. Figure (\ref{fig:loss_details}-(b)). The loss corresponding to the relaxed version of the divergence-free equation now converges much faster and more regularly. Its magnitude is now comparable to the temperature loss that  follows a much more flattering convergence slope as well. Very interestingly, there is a retroaction of these lower PDE residual losses onto the convergence of the label losses (cyan line). This coupling is extremely interesting as it demonstrates how an improved learning on the PDE part of the losses directly benefits the learning of the scarce temperature data. 

\begin{table}[!ht]\footnotesize
	\centering
	\begin{adjustbox}{center}
	\begin{tabular}{l c c c c c c c c c c }
    \toprule
\multicolumn{1}{c}{Model} & \multicolumn{2}{c}{Training}&
\multicolumn{2}{c}{Testing} & \multicolumn{5}{c}{Accuracy}\\
\midrule
\multicolumn{1}{c}{} & \multicolumn{1}{c}{size $(N_L,N_R)$} &\multicolumn{1}{c}{database} &  \multicolumn{1}{c}{size $(N_T)$}& \multicolumn{1}{c}{database} &\multicolumn{1}{c}{aRMSE} & \multicolumn{1}{c}{aMAE}& \multicolumn{1}{c}{$\mu$ error } &\multicolumn{1}{c}{$\sigma$ error}& \multicolumn{1}{c}{$aR_{\mathrm{corr}}$ } & \multicolumn{1}{c}{$aR^2$}\\
{3C1} & {$(7.334712\mathrm{e}6)$} &
{3Db1} &
{$6\mathrm{e}6$}& {3Db1} & 5.354e-03 & 3.030e-03 & 8.1 & 6.8 & 9.704e-01 & 9.413e-01 \\
{3C1$^{\text{r}}$} & {$(7.334712\mathrm{e}6)$} &
{3Db1} &
{$6\mathrm{e}6$}& {3Db1} & 1.262e-03 & 6.522e-04 & 1.4 & 1.0 & 9.983e-01 & 9.964e-01\\
\bottomrule
    \end{tabular}
    \end{adjustbox}
    \caption{$Ra=2 \cdot 10^9$ study: the caption is similar to the ones of Tables (\ref{tab:training},\ref{tab:accuracy}). Only $N_L$ is written in the table, as $N_R=N_L$ for those cases.}
    \label{tab:training3}
\end{table}

\begin{figure}[!h]
\centering
\includegraphics[trim={1cm 7cm 1cm 7cm},clip,width=7cm]{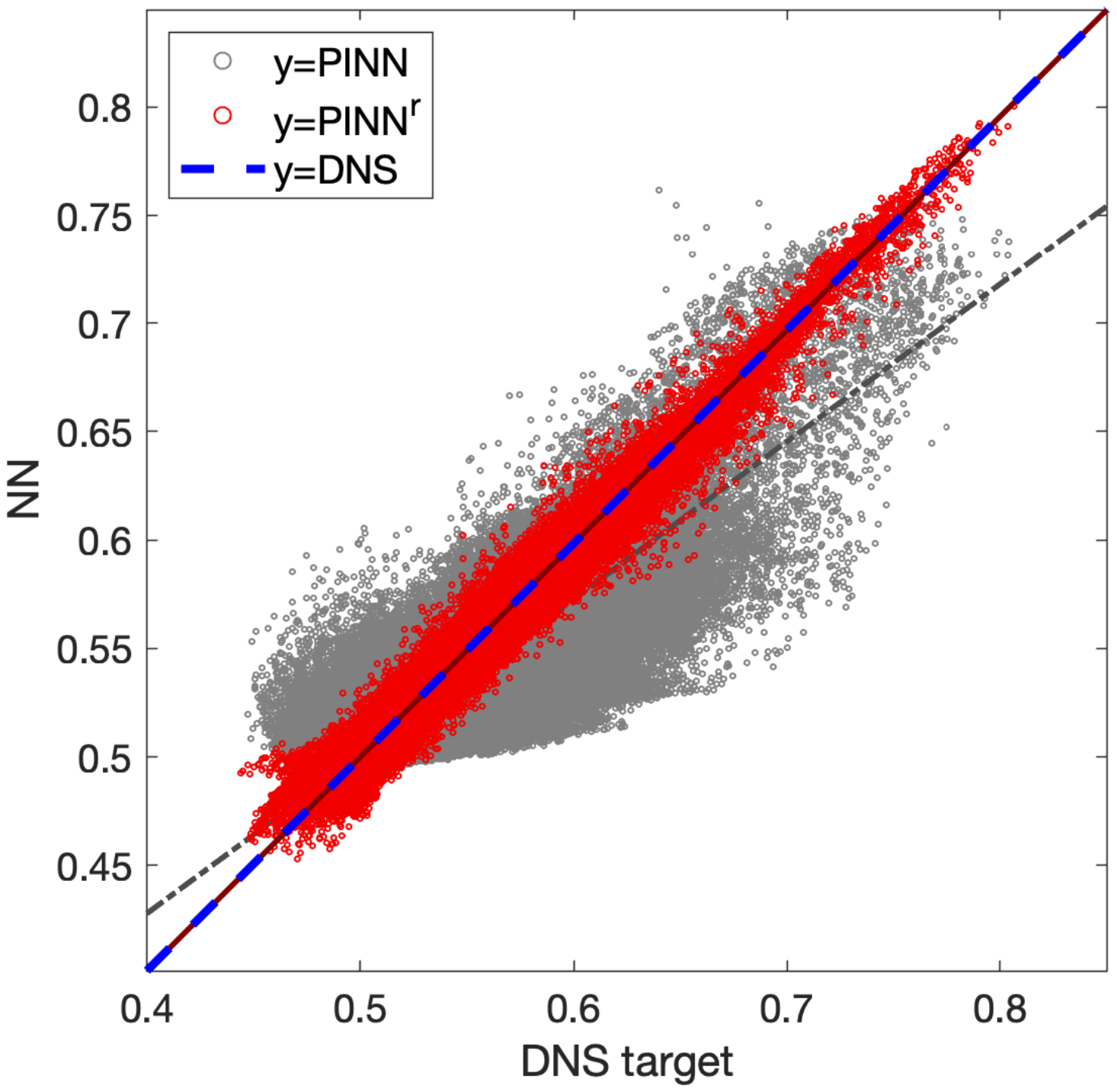}(a)\includegraphics[width=6.9cm]{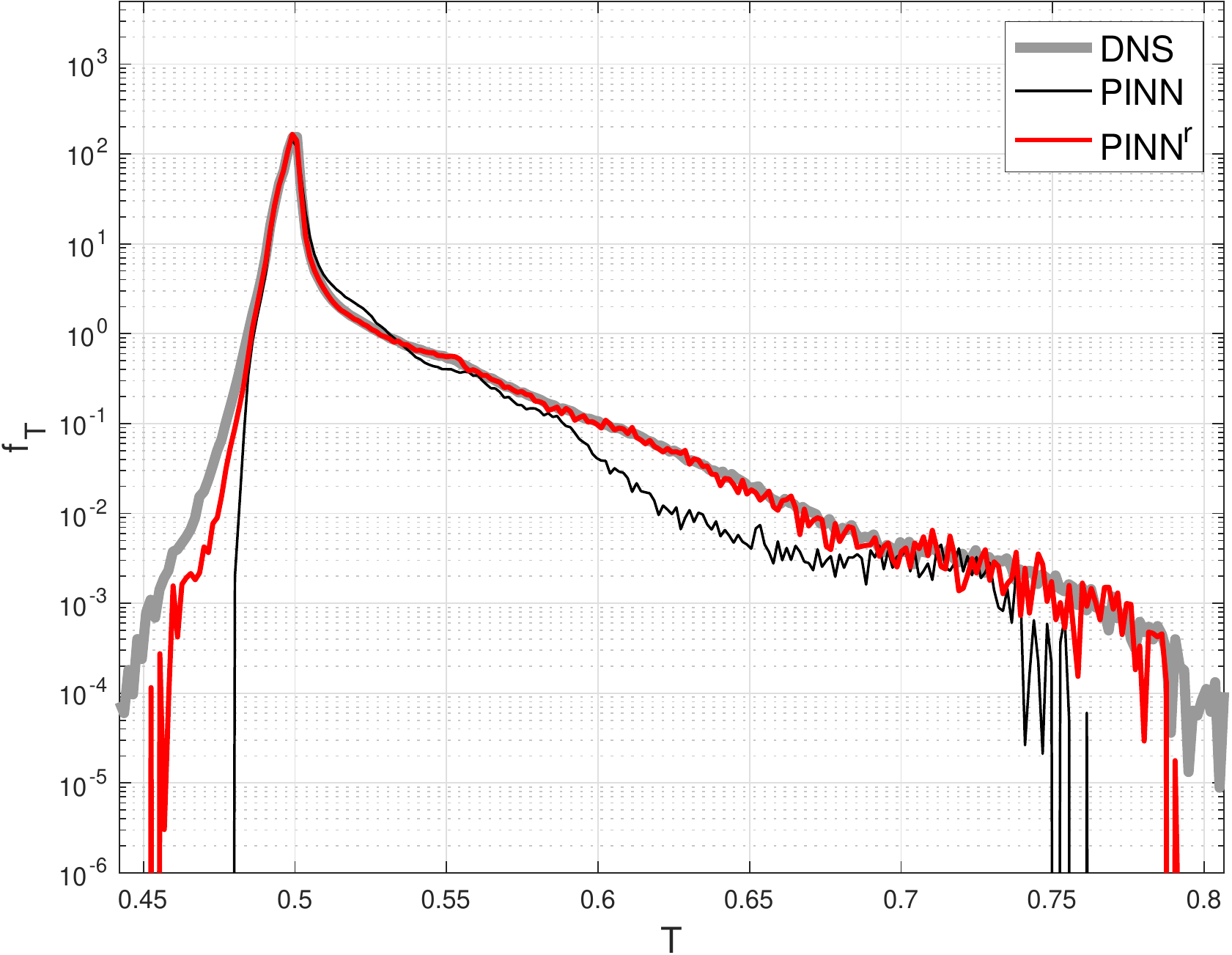}(b)
\caption{PINN predictive capabilities for RB flow at $Ra=2 \cdot 10^9$: comparisons between standard and relaxed (PINN$^\text{r}$) approaches. Temperature scatter plots (a) and probability density function (b) compared to the reference DNS. {The reference pdf is computed from the full DNS while the PINNs pdf are evaluated on the predicted test sample.} }
\label{fig:scatter_pdf_turbulent}
\end{figure}
\begin{figure}[!h]
\centering
\includegraphics[trim={4cm 7cm 4cm 7cm},clip,width=4.0cm]{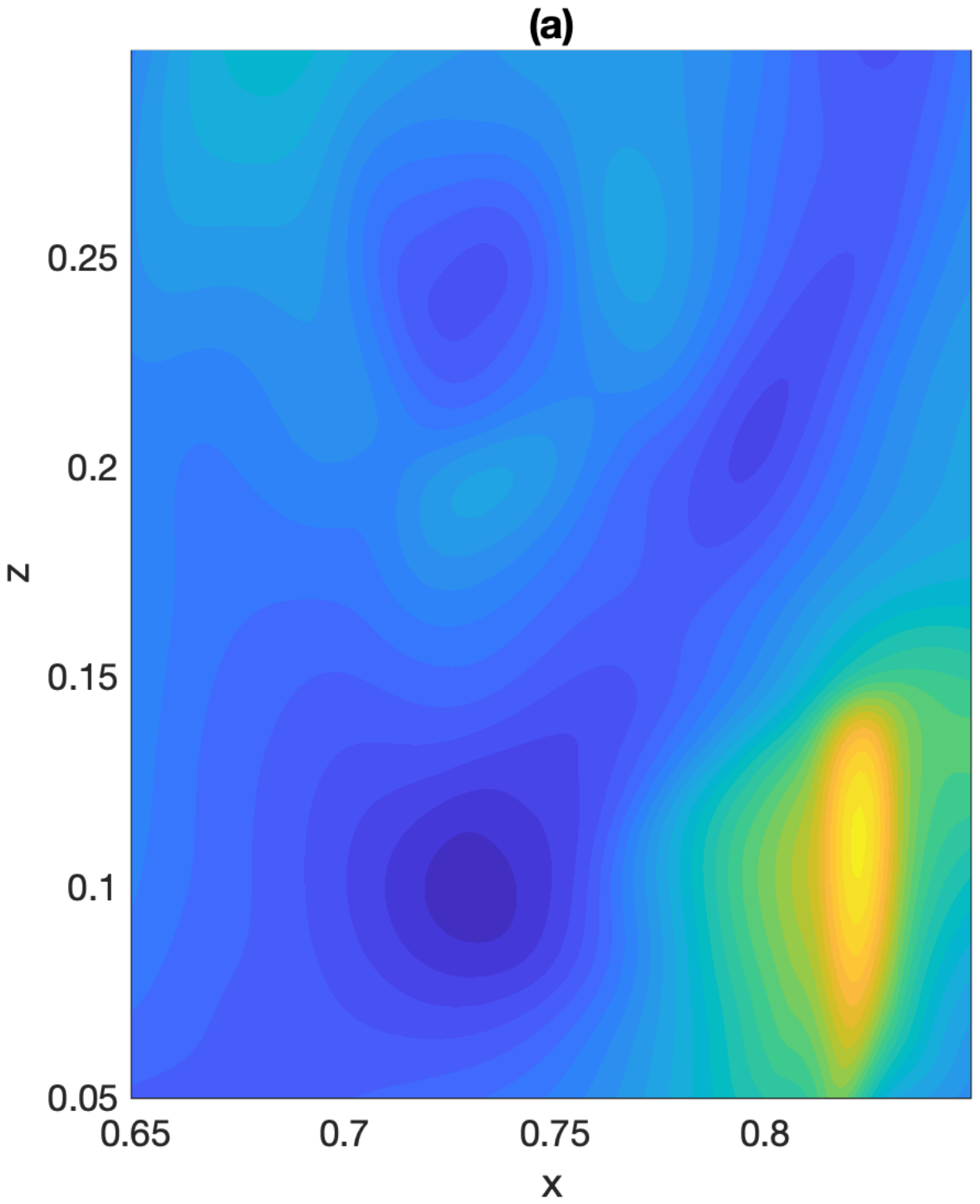}\hspace{0.1cm}
\includegraphics[trim={2.5cm 7.0cm 4.5cm 7cm},clip,width=4.3cm]{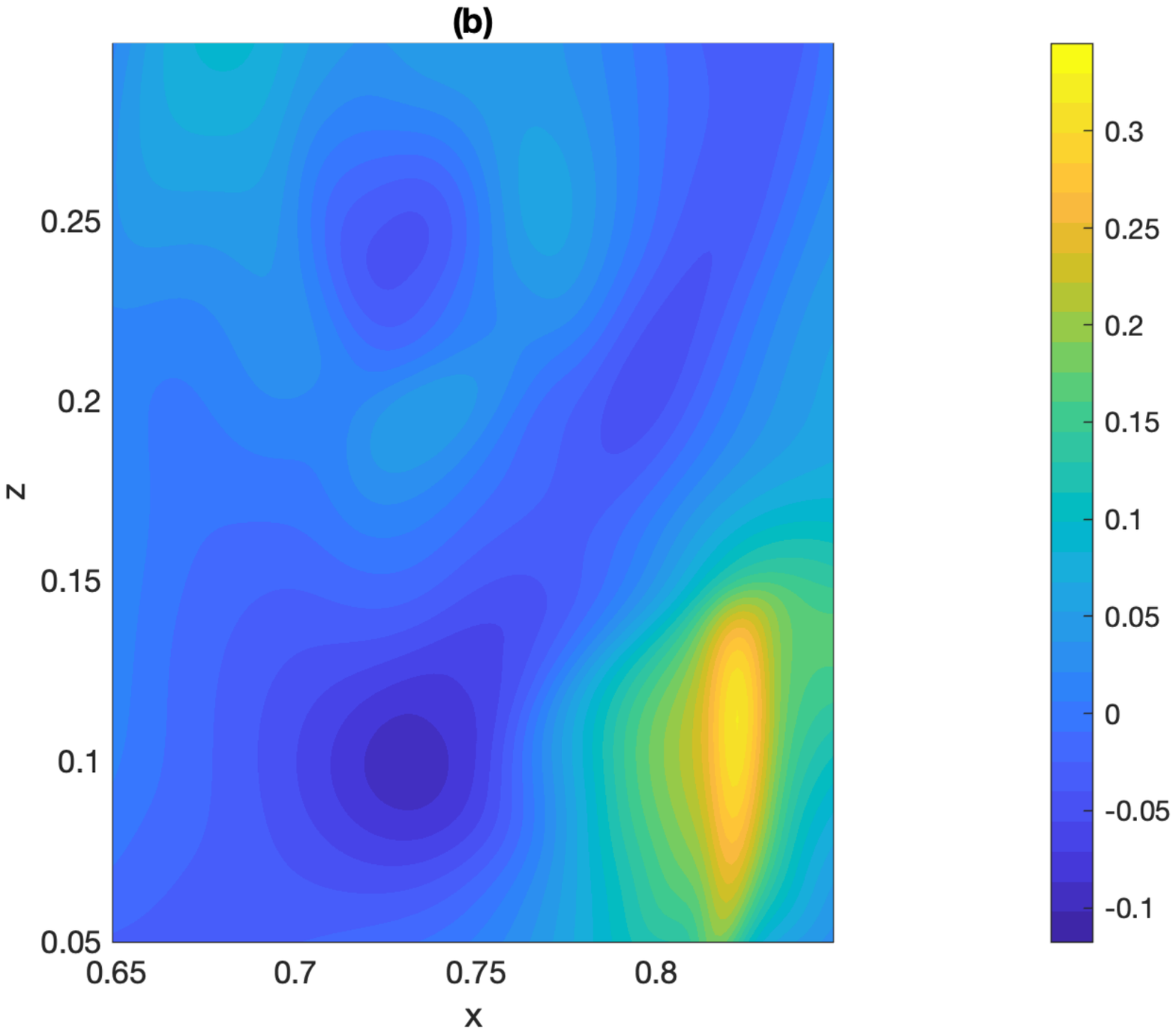}
\includegraphics[trim={4.0cm 7.0cm 4.0cm 7.0cm},clip,width=4.0cm]{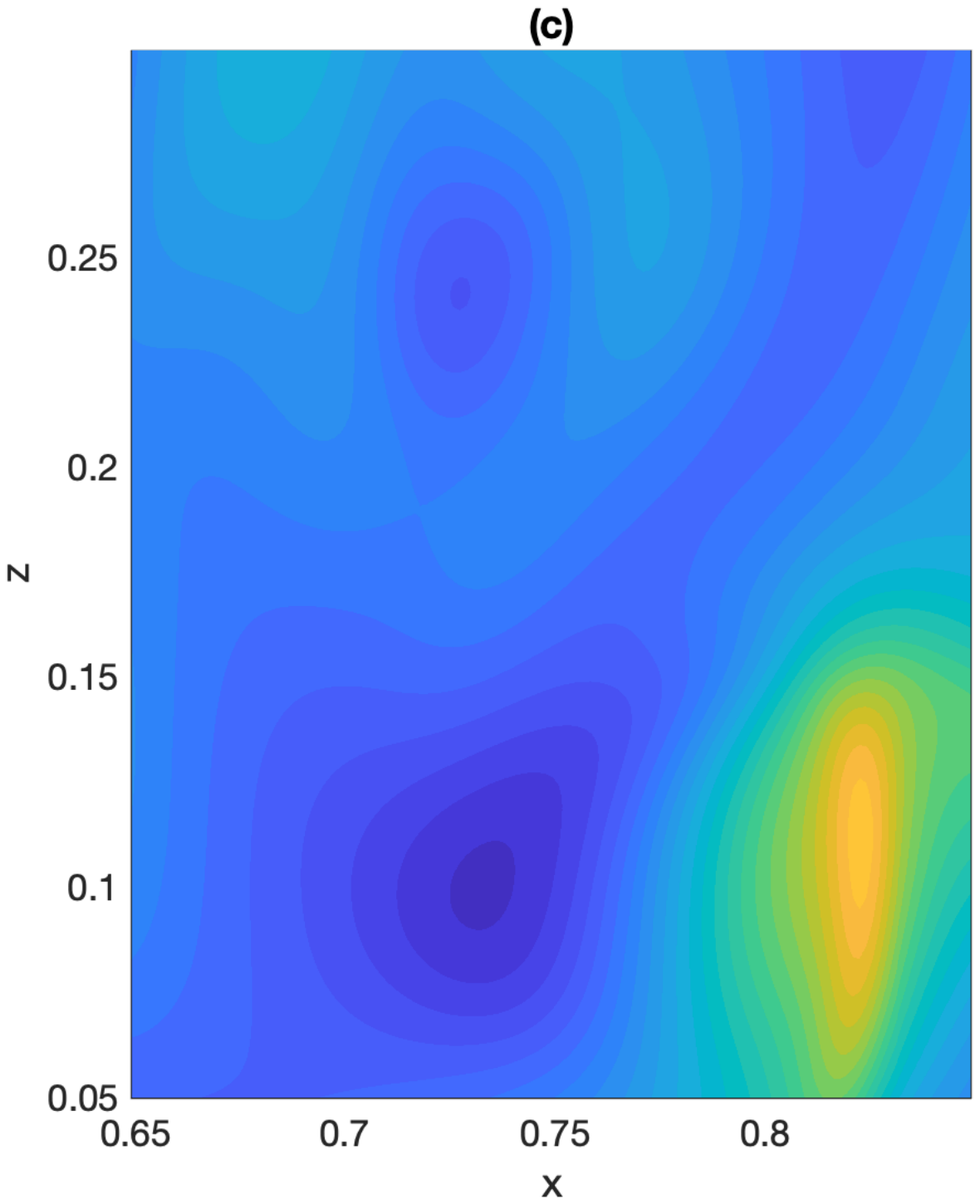}
\includegraphics[trim={16.5cm 7.0cm 0.0cm 6.3cm},clip,width=1.375cm]{././wz_pred_x0p82yz.pdf}\\
\hspace{-0.3cm}\includegraphics[trim={5cm 7cm 6cm 7cm},clip,width=3.0cm]{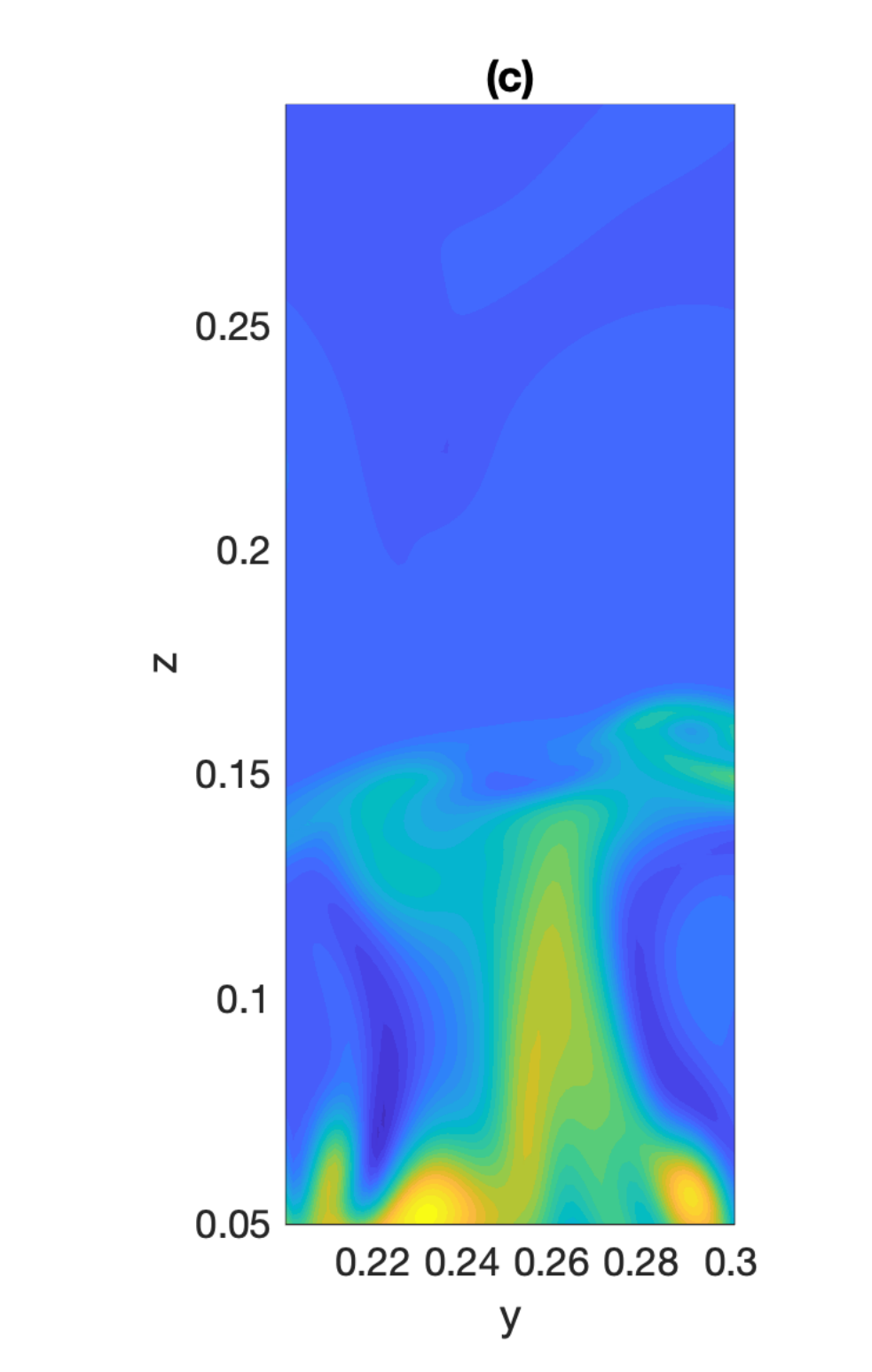}
\hspace{1.6cm}\includegraphics[trim={5.0cm 7.0cm 6.0cm 7.0cm},clip,width=3cm]{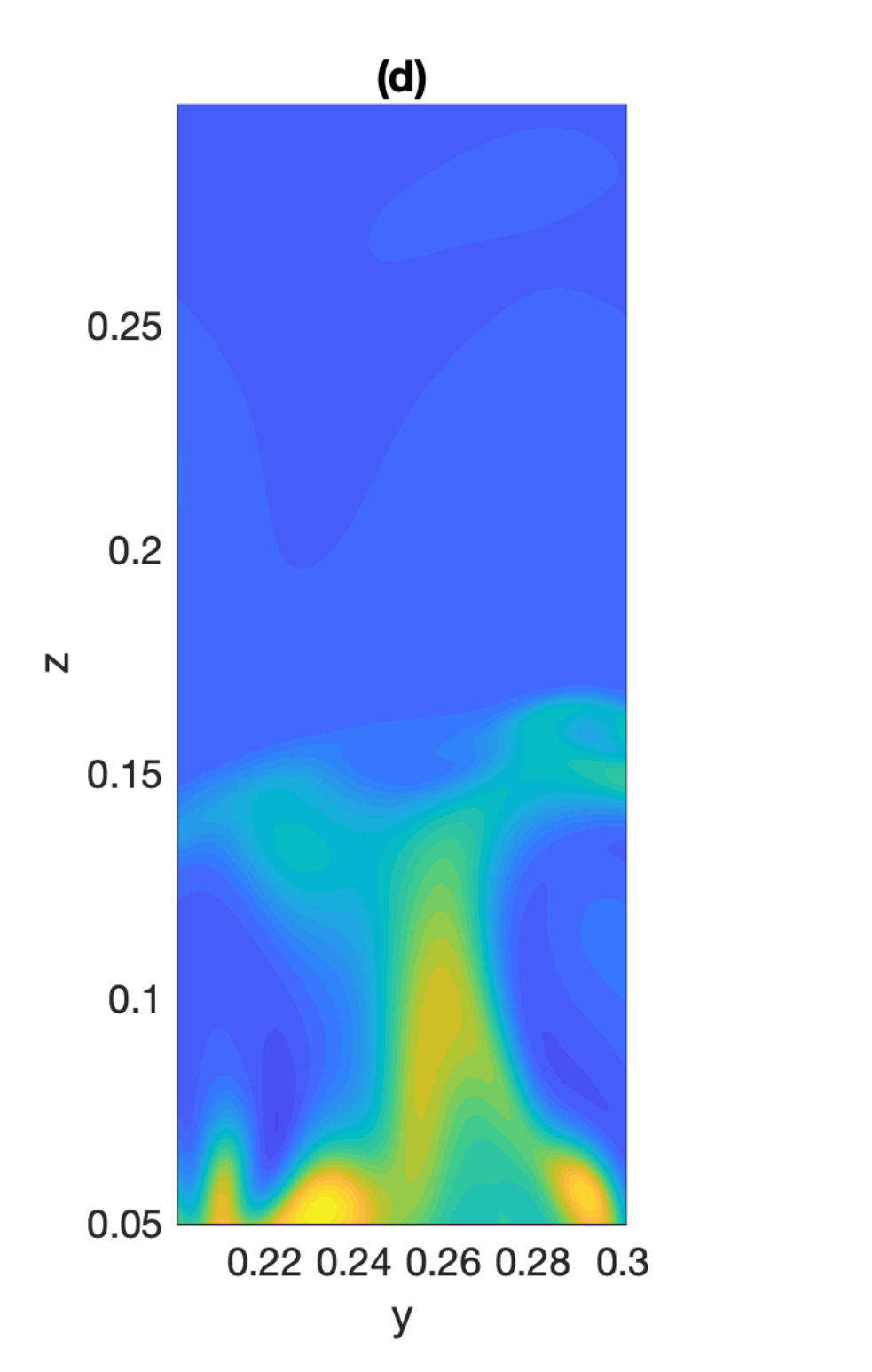}
\hspace{0.9cm}\includegraphics[trim={5.0cm 7.0cm 6.0cm 7.0cm},clip,width=3cm]{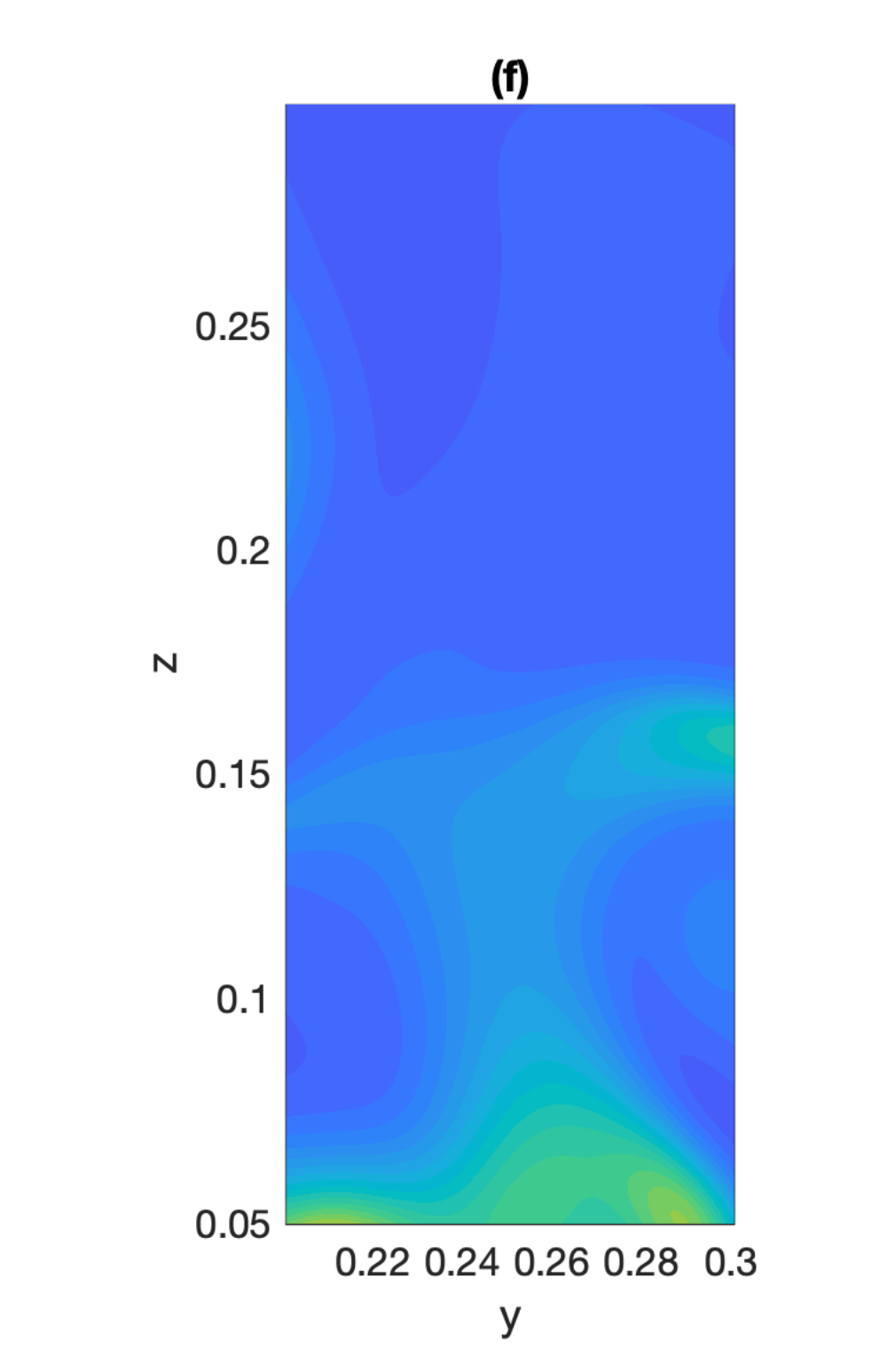}
\includegraphics[trim={16.5cm 7.0cm 0.0cm 6.3cm},clip,width=1.34cm]{././Temp_pred_x0p82yz.pdf}
\caption{Comparison of some ground truth (DNS) (a,d), PINN$^{\text{r}}$-predicted (b,e) and standard PINN-predicted (c,f) instantaneous sliced fields with a thin sheet-like plume located at the bottom of the domain. Top row: vertical flow velocity $\mathrm{v}_z(\cdot,y_0=0.25,\cdot,t=290.83)$ and bottom row: temperature $T(x_0=0.82,\cdot,\cdot,t=290.83)$ fields. The time instant is chosen so as to correspond to a DNS snapshot that is {\em not} included in the training database. Predictions are requested on the fine DNS spatial grid.}
\label{fig:slabs2}
\end{figure}

Table (\ref{tab:training3}) summarizes the specifics and accuracy of the two tested models. The results are very clear and show the undeniable superiority of the PINN$^{\text{r}}$ approach with very good accuracy. 
{Moreover, an ambitious validation of this approach was also carried out on the full DNS. That is to say that the model was used to predict the solution at the coordinates of the full DNS referenced hereinbefore (i.e. on a sample of size $N_T=129\times 65 \times 219 \times 249$, corresponding to a doubling of the spatial resolution in each direction and a tripling of the temporal resolution compared to the training sample). Very impressively, the errors computed in the relative $L_2$ norm were only: $0.3\%$ for the temperature, $1.79\%$ for $\mathrm{v}_x$, $2.708\%$ for $\mathrm{v}_y$, $3.416\%$ for $\mathrm{v}_z$ and $4.038\%$ for the pressure, respectively.}\\
Figures (\ref{fig:scatter_pdf_turbulent}) present the scatter plot of the PINN temperature results of case 3C1 and {3C1$^{\text{r}}$} (a), and the corresponding pdf compared to the reference DNS. 
The difference is for instance very striking when looking at the way the approximation is now capable of sharply capturing the long  tail of the skewed temperature probability density function (PDF).
{This asymmetric PDF shape is typical of the mixing layer \cite{wang_jfm_2019}, in which the training domain is placed. The long quasi-exponential tail for large temperature fluctuations is the signature of the travelling hot plumes, intermittently passing through the domain.
Even if  temperature is only scrutinized here, quantitative improvements of the PINN$^{\text{r}}$ approach do occur for all flow variables. On the contrary, it can be seen that the original PINN approach miss a large part of the warm plumes, but especially the cold (descending) ones. }\\
Figures (\ref{fig:slabs2}) were chosen to show an example on how the surrogate is capable of accurately predicting a strong small plume with very anisotropic structure, despite being located close to the domain boundaries and occurring at a time instant never visited during training. {It is remarkable how well the intricate temperature distribution within the plume is approached by PINN$^\textrm{r}$.}

\subsection{Some perspectives}

These promising results open the way for more involved {\em parametric} surrogate modeling (useful in the context of design optimization, model calibration, sensitivity analysis, etc), i.e. nonlinear problem parametrized by some (potentially not well-known) physical quantities, playing the role of additional inputs to our PINNs models. 
Despite the large body of literature on uncertainty quantification, including aleatoric and epistemic uncertainties in fluid mechanics \cite{UQinCFD2013}, few works have attempted to propose DNN-based scalable algorithms for parametric surrogate CFD modeling, due to the lack of  {\em a posteriori} error estimation and convergence theory. Moreover, training data is a severe bottleneck in most parametric fluid dynamics problems since each data point in the parameter space requires an expensive numerical simulation based on first principles.\\
Nevertheless, some numerical perspectives may be drafted by examining some of the limiting computational aspects of the PINN approach.  The PINN algorithm infuses the system governing equations into the network by modifying the loss function with a contribution acting as a penalizing term to constrain the space of admissible solutions. The high-dimensional non-convex optimization problem of this composite loss function involves a large training cost related to the time-integration of the nonlinear PDEs and the depth of the neural network architectures. We have seen that the approach may be efficient despite a training based on a very sparse data sample, while other approaches investigate variant sampling strategy, e.g. \cite{MAO2020112789,mishra2020enhancing}.\\
But in the more demanding case of parametric surrogate construction, an effort should be pursued on the front of efficient data sampling strategy. Indeed,  optimal sampling would ensure a right balance and therefore good complementarity between the information provided by the PDEs and by the data. A potential breakthrough  would be to propose a dynamic selection of relevant data for the PINN learning, operating synchronously with the physical simulation, and allowing sparser spatial-temporal sampling, responding in part to the storage problem of DNS simulations. Moreover, it could be beneficial to simultaneously build a data index structure allowing to benefit from importance sampling, e.g. importance sampling tree technique \cite{pmlr-v48-canevet16}. \\
In the case of parametric surrogate modeling, another interesting approach would be the one of transfer learning (TL). The TL domain seeks precisely to transfer the knowledge acquired to a training dataset to better process a new so-called ``target dataset". The transfer can therefore take the form of a parallel relearning of the neural network taking into account the evolving parameters (geometric or physical for instance).\\
More importantly, we have experienced the high sensitivity of the learning process to the way we enforce (some of) the PDEs
in the PINNs framework and the impact it had on the global accuracy of the scheme. Inspired by the work of Perdikaris and co-authors \cite{wang2020understanding}, we believe it would be worthwhile tracking the gradients of each individual terms in the PDEs constraints with respect to the weights in each hidden layer of the neural network, rather than tracking the gradients of the aggregated loss. This will help monitoring the distribution of those back-propagated gradients during training and propose a learning rate annealing algorithm that utilizes gradient statistics to balance the interplay between the different terms in the regularization components of the composite loss function. More specifically, due to the stochastic nature of the gradient descent updates, updated learning rates should be computed as running averages of previous values and do not need to be updated at each iteration of the optimization solver.

Other perspectives and current works involve -- the decomposition of the computational domain into several training sub-domains in order to better scale locally-adapted PINN models, -- handling of aleatoric uncertainty associated with noisy training data by means of physics-informed Bayesian neural networks \cite{yang2021b},
-- the mixing of various labeled data sources -- hybrid regularization techniques combining physics-informed regularization with more classical $L_2$, $L_1$ and/or dropout regularizations. 

\subsection{Acknowledgments}

{The DNS database has been built using granted access to the HPC resources of IDRIS under allocation 2a0326 made by GENCI. We thank Dr. Yann Fraigneau for his help and great expertise in the development of the DNS SUNFLUIDH solver. }

\bibliographystyle{elsarticle-num}
\bibliography{mybibfile}

\end{document}